%% file: LLaMAX.tex
\pdfoutput=1

\documentclass[11pt]{article}

\usepackage{ACL2023}

\usepackage{times}
\usepackage{latexsym}
\usepackage{pifont}
\newcommand{\cmark}{\ding{51}}%
\newcommand{\xmark}{\ding{55}}%

\usepackage[T1]{fontenc}

\usepackage[utf8]{inputenc}

\usepackage{microtype}

\usepackage{inconsolata}
\usepackage{makecell}
\usepackage{enumitem}
\usepackage{xspace}
\usepackage{graphicx}
\usepackage{multirow}
\usepackage{multicol}
\usepackage{booktabs}
\usepackage[ruled]{algorithm2e}
\usepackage{amssymb}
\usepackage{xcolor}
\usepackage{colortbl}

\usepackage{CJKutf8}
\usepackage[encapsulated]{CJK}
\usepackage{amsmath}
\newcommand{\vect}[1]{\boldsymbol{#1}}
\newcommand{\x}[1][]{\ifx\empty#1\empty \vect{x} \else \vect{x}^{\scriptscriptstyle (#1)} \fi}
\newcommand{\xp}[2][]{\x[#1]_{<#2}}
\newcommand{\xt}[2][]{\ifx\empty#1\empty x_{\scriptscriptstyle #2} \else x^{\scriptscriptstyle (#1)}_{\scriptscriptstyle #2} \fi}

\definecolor{mycolor}{HTML}{42a5f5}

\newcommand{\method}{LLaMAX\xspace}
\newcommand{\name}{LLaMAX2\xspace}
\newcommand{\latest}{LLaMAX3\xspace}

\title{\method: Scaling Linguistic Horizons of LLM by Enhancing Translation Capabilities Beyond 100 Languages}

\author{
Yinquan Lu\textsuperscript{\rm1}, Wenhao Zhu\textsuperscript{\rm1,\rm2}, Lei Li\textsuperscript{\rm3}, Yu Qiao\textsuperscript{\rm1}, Fei Yuan\textsuperscript{\rm1}\thanks{Corresponding author.} \\
\textsuperscript{\rm 1} Shanghai AI Laboratory, \textsuperscript{\rm 2} Nanjing University, \textsuperscript{\rm 3} Carnegie Mellon University
\\
\texttt{\{luyinquan,yuanfei\}@pjlab.org.cn}, \texttt{zhuwh@smail.nju.edu.cn}, \texttt{leili@cs.cmu.edu}
}

\begin{document}
\maketitle

\input{EMNLP-2024/sections/0.abstract}

\input{EMNLP-2024/sections/1.introduction}

\input{EMNLP-2024/sections/4.technique_design}

\input{EMNLP-2024/sections/3.approach}

\input{EMNLP-2024/sections/5.experiments}

\input{EMNLP-2024/sections/6.analysis}

\input{EMNLP-2024/sections/2.related_work}

\input{EMNLP-2024/sections/7.conclusion}


\bibliography{custom}
\bibliographystyle{acl_natbib}

\clearpage
\newpage
\input{EMNLP-2024/sections/8.appendix}

\end{document}

%% file: EMNLP-2024/sections/0.abstract.tex
\begin{abstract}
    Large Language Models~(LLMs) demonstrate remarkable translation capabilities in high-resource language tasks, yet their performance in low-resource languages is hindered by insufficient multilingual data during pre-training. To address this, we conduct extensive multilingual continual pre-training on the LLaMA series models, enabling translation support across more than 100 languages. Through a comprehensive analysis of training strategies, such as vocabulary expansion and data augmentation, we develop \method. Remarkably, without sacrificing its generalization ability, \method achieves significantly higher translation performance compared to existing open-source LLMs~(by more than 10 spBLEU points) and performs on-par with specialized translation model~(M2M-100-12B) on the Flores-101 benchmark. Extensive experiments indicate that \method can serve as a robust multilingual foundation model. The code~\footnote{\url{https://github.com/CONE-MT/LLaMAX/.}} and the models~\footnote{\url{https://huggingface.co/LLaMAX/.}} are publicly available.
\end{abstract}


%% file: EMNLP-2024/sections/1.introduction.tex
\section{Introduction}

\begin{figure}[!ht]
    \centering
    \includegraphics[width=1\linewidth]{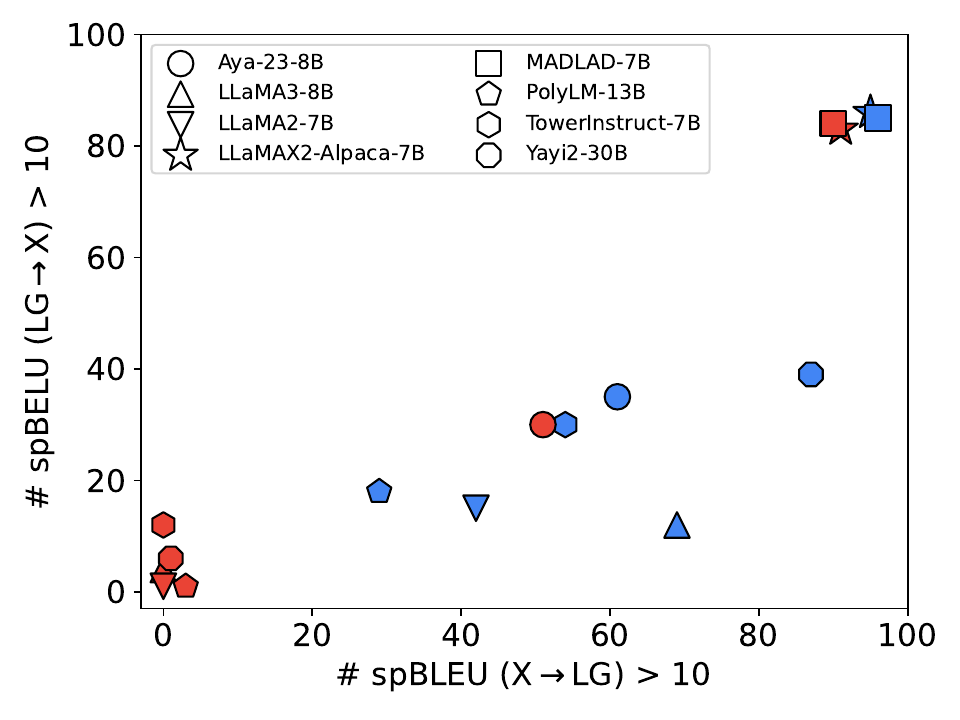}
    \caption{We assess translations in both directions, X$\rightarrow$LG and LG$\rightarrow$X, across various models using Flores-101 test, with X representing all 101 languages included in Flores-101. The results are visualized in a figure where different markers represent various models, a \textcolor{red}{red} marker indicates that the language (LG) is \textcolor{red}{Arabic}, while a \textcolor{blue}{blue} marker indicates \textcolor{blue}{English}. We count the number of translation directions that achieve a spBLEU score higher than 10. The findings indicate that modest LLMs demonstrate strong support for English-centric translation, but underperform in Arabic-centric translation. }
    \label{fig:introduction}
\end{figure}

Large Language Models~~(LLMs;~\citealp{gpt,zhang2022opt,palm,openai2023gpt4,llama1,LLaMA2}) exhibit excellence performance in translation tasks involving high-resource languages~\citep{vilar-etal-2023-prompting,Zhu2023MultilingualMT}, yet their effectiveness in low-resource translation is suboptimal~\citep{hendy2023good,bang2023multitask,Zhu2023MultilingualMT}. 
Figure~\ref{fig:introduction} illustrates the number of translation directions with performance exceeding 10 spBLEU~\citep{flores101} score on Flores-101~\citep{flores101}.
It is evident the majority of models are clustered around the origin point for Arabic-centric translations, demonstrating a significant disparity when compared to their English-centric performance.

This discrepancy is primarily due to the lack of pre-training data for these languages~\citep{wei2023polylm,yuan2023multilingual,alves2024tower}. Many researchers are actively working to address this issue. \citet{guo-etal-2024-teaching-large} enhance the LLMs' ability by translating low-resource languages after learning textbooks. ~\citet{Zhu2023MultilingualMT} find cross-lingual examples that can provide better task guidance for low-resource translation. In addition to the efforts focus on the fine-tuning stage, some studies have attempted to train a multilingual LLM from scratch~\citep{wei2023polylm}, or to train a language-specific LLM~\citep{faysse2024croissantllm, alves2024tower, cui2024efficient}. 
However, the languages covered by these works are not extensive~\citep{wei2023polylm, alves2024tower,luo2023yayi}, and the translation performance is still unsatisfactory~\citep{wei2023polylm,alves2024tower,luo2023yayi}. 


To tackle this discrepancy, we conduct a massive multilingual continual pre-training for non-English languages. Firstly, we present a comprehensive analysis of critical technical designs, including vocabulary extension~(Section~\ref{sec:vocab_analysis}) and data augmentation~(Section~\ref{sec:aug_analysis}). These analyses establish the groundwork for the training procedure, directly influencing the efficacy and, ultimately, the performance of the LLMs. Subsequently, we apply those strategies in continual pre-training using both parallel and monolingual data to enhance the translation performance of LLMs across the 102 languages covered by Flores-101,  particularly for low-resource languages.

A primary challenge in expanding language support lies in determining the appropriate vocabulary~\citep{cui2024efficient,fujii2024continual}. After assessing the impact of adding language-specific tokens from various angles: tokenization granularity, embedding quality, and the model's inner distribution, we find that introducing a small number of new tokens significantly degrades existing LLM performance, while a larger new token set increases training complexity and data requirements. Surprisingly, adhering to the original vocabulary of LLMs emerges as the most cost-effective strategy for extending LLMs to 102 languages.



Another great challenge in extending language support is the scarcity of data for low-resource languages~\citep{chang2023multilingualitycurselanguagemodeling, guo-etal-2024-teaching-large}. 
To alleviate the scarcity of training data, we delve into dictionary-based data augmentation~\citep{pan-etal-2021-contrastive,2023arXiv230506575L} and conduct a comprehensive analysis of various augmentation strategies. 
This analysis takes into consideration different dictionaries and data sources~(monolingual or parallel data). 
We find that the optimal approach for data augmentation involves using parallel data, with the choice of dictionary correlated to the number of target language entities it covers.

Finally, we leverage the above discussed techniques to perform large-scale, multilingual continual pre-training on LLaMA series models~\citep{LLaMA2, llama3modelcard}, resulting in \method series models~(\name and \latest). The \name, trained over 60 days using 24 A100 GPUs, significantly enhances translation capabilities and achieves comparable performance~(evaluated on Flores-101) to the specialized translation model M2M-100-12B~\citep{fan2021beyond}. 
Specifically, our method demonstrates an average improvement of more than 10 spBLEU compared to baseline models in low-resource-centric translation, as shown in Table~\ref{tab:flores}. 
Furthermore, when extending our evaluation to Flores-200~\citep{nllb2022}, it shows significant performance enhancements even for languages not included in the training set. All these translation performance improvements do not compromise general task performance. 
Interestingly, enhancing translation capabilities also establishes a robust multilingual base model foundation. When comparing results of supervised fine-tuning using task-specific English data on the X-CSQA~\citep{xcsqa}, XNLI~\citep{xnli}, and MGSM~\citep{mgsm} tasks, we observe an average improvement of 5 points over LLaMA2. Our main contributions can be summarized as follows:
\begin{itemize}[nosep,leftmargin=0.3cm]
    \item A series of open-sourced \method models enhance the translation performance across more than 100 languages.
    \item Comprehensive analysis of the key techniques in multilingual continual pre-training, including vocabulary extension and data augmentation.
    \item Extensive experiments on key technique design, comprehensive translation benchmark evaluation across various models, general task testing, and supervised fine-tuning on task-specific data demonstrate the superiority of \method.
\end{itemize}

%% file: EMNLP-2024/sections/4.technique_design.tex
\section{Key Technique Design}
\label{sec:technique_design}

\begin{table*}[!htb]
    \centering
    \small
    \resizebox{\linewidth}{!}{
    \begin{tabular}{r|cccccc|cccccc}
        \toprule
       \multirow{2}{*}{\textbf{\makecell[c]{\# New \\ Token}}}& \multicolumn{6}{c|}{\textbf{Romanian~(ro)}} & \multicolumn{6}{c}{\textbf{Bengali~(bn)}} \\
        & \textbf{fertility} & \textbf{cosine} & \textbf{R@1} & \textbf{shift distance}  & \textbf{\# shift token }  & \textbf{spBLEU} &  \textbf{fertility} & \textbf{cosine} & \textbf{R@1}  & \textbf{shift distance}  & \textbf{\# shift token } & \textbf{spBLEU} \\

        \midrule
         0 & 2.25 & 0.39 & 0.37 & 0.4708 & 112 & \textbf{32.50} & 8.62 & 0.17 & 0.01 & 0.4689 & 112 & \textbf{20.12} \\ 
         \midrule
        100 & 2.19 & 0.36 & 0.34 & 0.4720 & 112 & 28.75 &  4.96 & 0.14 & 0.02 & 0.4680 &  113 & 14.02  \\ 
        800 & 2.02 & 0.35 & 0.36 & 0.4682 & 113 & 27.78 & 3.21 & 0.13 & 0.02 & 0.4706 & 113 & 10.18\\ 
        1600 & 1.93 & 0.34 & 0.34 & 0.4690 & 113 & 26.40 &  2.78 & 0.13 & 0.02 & 0.4695 & 113 & 1.82\\ 
        6,400 & 1.74 & 0.31 & 0.31 & 0.4694 & 113 & 22.66 &2.15 & 0.12 & 0.02 & 0.4712 & 113 & 1.96 \\ 
        12,800 & 1.63 & 0.29 & 0.29 & 0.0205 & 1 & 21.95 & 1.95 & 0.12 & 0.02 & - & 0 & 1.84 \\ 
        25,600 & 1.53 & 0.27 & 0.28 & - & 0 & 19.72 & 1.80 & 0.12 & 0.02 & - & 0 & 2.58 \\ 
        51,200 & 1.45 & 0.26 & 0.25 & 0.0203 & 1 & 17.79 & 1.70 & 0.12 & 0.03 & - & 0 & 1.14\\ 
        \bottomrule
    \end{tabular}}
    \caption{Building upon LLaMA2, we add varying numbers of languages-specific new tokens, fully fine-tune LLaMA2, and test the translation performance of en$\rightarrow$ro~(bn) using Flores-101 test. Furthermore, we assess the effect of new tokens using several metrics: fertility, the cosine similarity with English sentence embeddings, the performance in the English language retrieval translation task (R@1), and the distribution shift of the original embedding vector. Our experiments demonstrate that the inclusion of new words significantly complicates the learning process, underscoring that the integration of new words is a complex task. }
    \label{tab:extend_vocab}
\end{table*}

\paragraph{Existing Pipeline.}  Exploring adapting pre-trained LLMs to new languages without starting from scratch seems to have a concise pipeline, resulting in ChineseLLaMA2~\citep{cui2024efficient}, Swallow~\citep{fujii2024continual}, and so on. 
This pipeline comprises three crucial steps: 1) vocabulary expansion: extending the vocabulary of LLMs by adding new tokens specific to that language and initializing these new tokens as the average of embeddings from the existing tokens~\citep{dobler-de-melo-2023-focus}. 2) continual pre-training: continual pre-training LLM on a large corpus of text data from the target language. 3) instruction tuning: aligning the model with specific tasks or instructions, enhancing its performance. Instead of simply following the pipeline, we analyze primarily two key challenges related to the extension of language support: determining an appropriate vocabulary~(in Section~\ref{sec:vocab_analysis}) and improving the effectiveness of data augmentation~(in Section~\ref{sec:aug_analysis}). For a more detailed analysis, refer to the discussions on the selection of multi-hop translation in the lexicon (see Appendix~\ref{sec:select_hop_translation}) and the format of parallel data during continual pre-training (see Appendix~\ref{sec:parallel_format}).

\begin{table*}[!htb]
    \centering
    \footnotesize
    \resizebox{0.9\textwidth}{!}{
    \begin{tabular}{lrrrrrrrrrr}
    \toprule
        \multicolumn{1}{c}{\multirow{2}{*}{\textbf{Setting}}} & \multicolumn{3}{|c}{\textbf{spBLEU}} &  \multicolumn{4}{|c}{\textbf{\# entity}}  & \multicolumn{3}{|c}{\textbf{similarity}}  \\
       
        & \multicolumn{1}{|c}{\textbf{MUSE}} & \textbf{PanLex} & \textbf{$\Delta$} &\multicolumn{1}{|c}{\textbf{MUSE}} & \textbf{PanLex} & \textbf{$\Delta$} & \textbf{ratio}  & \multicolumn{1}{|c}{\textbf{MUSE}} & \textbf{PanLex} & \textbf{$\Delta$} \\ 
        \midrule
        en$\rightarrow$ta & \multicolumn{1}{|c}{3.74} & 3.45 & -0.29 & \multicolumn{1}{|c}{139,134} & 91,652 & -47,482 & 0.66 & \multicolumn{1}{|c}{0.08} & 0.04 & -0.04  \\ 
        en$\rightarrow$th & \multicolumn{1}{|c}{5.45} & 6.14 & 0.69 & \multicolumn{1}{|c}{21,567} & 297,573 & 276,006 & 13.80 & \multicolumn{1}{|c}{0.20} & 0.06 & -0.14  \\ 
        en$\rightarrow$fr & \multicolumn{1}{|c}{44.03} & 43.85 & -0.18 & \multicolumn{1}{|c}{139,134} & 568,428 & 429,294 & 4.09 & \multicolumn{1}{|c}{0.31} & 0.35 & 0.04  \\ 
        en$\rightarrow$zh & \multicolumn{1}{|c}{14.65} & 16.64 & 1.99 & \multicolumn{1}{|c}{139134} & 1,333,762 & 1,194,628 & 9.59 & \multicolumn{1}{|c}{0.14} & 0.09 & -0.05  \\ 
        en$\rightarrow$es & \multicolumn{1}{|c}{26.98} & 27.36 & 0.38 & \multicolumn{1}{|c}{142,780} & 433,468 & 290,688 & 3.04 & \multicolumn{1}{|c}{0.28} & 0.32 & 0.04  \\
    \bottomrule
    \end{tabular}}
    \caption{Evaluate a specific data augmentation technique with different dictionaries. We measure translation performance (spBLEU), the number of target language entities in the dictionary (\# entity), and average cosine similarity of entities (similarity), revealing a strong correlation between performance and ``\# entity''.}
    \label{tab:dictionary_compare}
\end{table*}

\subsection{Existing Vocabulary is Adequate.}
\label{sec:vocab_analysis}

\paragraph{Setting.} We conduct a series of analytical experiments on the LLaMA2 vocabulary. Our initial focus is on examining the correlation between fertility and the quality of token representation. Here, fertility refers to the ratio of the length of the token sequence produced by the LLaMA2 tokenizer to the length of the input sentence when split by spaces~(Chinese and Japanese is split by character). Furthermore, we carry out experiments using 10,000 en$\rightarrow$ro and en$\rightarrow$bn bilingual sentence pairs from Lego-MT dataset. For new tokens, the BBPE algorithm is executed on language-specific data from MADLAD-400 to produce a vocabulary of 100,000 tokens. Within this vocabulary, language-specific tokens are arranged based on their frequency in the corpus. Subsequently, we identify the top-k tokens (where k is determined by the corresponding ``\#New Token'' in Table~\ref{tab:extend_vocab}) that are absent in the original LLaMA vocabulary and incorporate them as new tokens into the LLaMA vocabulary.
In each experiment, we introduce a varying number of language-specific new tokens and evaluate each model on the Flores-101.

\paragraph{Research Question 1: Why is adding new tokens considered a straightforward method for extending language support?} We assess the quality of representation by en$\rightarrow$X translation task. This task identifies the translated result that best aligns with the corresponding English sentence within an extensive target dataset, and evaluates with Recall at top 1, denoted as R@1~\citep{kabir-carpuat-2021-umd}. A higher R@1 value signifies a more robust quality of the representation. Concurrently, we present the cosine similarity of representations generated by LLaMA2 for identical sentences in English and other languages. On experiments across 102 languages, more details in Appendix~\ref{sec:corr_fert_and_quality}, there exists a strong correlation between fertility and the quality of representation, evidenced by a Spearman correlation coefficient of approximately \textbf{-0.88} for each assessed quality metric.


\paragraph{Research Question 2: Does adding new tokens to reduce fertility yield prompt performance improvements?} Extending vocabulary is a common method to reduce fertility. However, while adding new tokens indeed reduces fertility, it does not necessarily enhance its ability to capture and generalize linguistic patterns across multiple languages. As shown in Table~\ref{tab:extend_vocab}, the more new tokens added, the worse the translation performance. 

\paragraph{Research Question 3: What is the impact of adding new tokens on model performance?} As demonstrated in Table~\ref{tab:extend_vocab}, even the addition of a small number~(100) of new language-specific tokens can have a significant impact on the multilingual performance of LLMs. In addition, we conduct a further analysis on the original tokens~(32k) embedding distribution and the token number before and after adding new tokens by KS-Lottery~\citep{yuan2024ks}. For more details on KS-Lottery, refer to Appendix~\ref{sec:ks_lottery}. As the experimental result of ``shift distance'' and ``\# shift token'' in Tabel~\ref{tab:extend_vocab}, fine-tuning the entire model with limited new tokens follows a similar pattern to that with the original vocabulary. However, an excessive number of new tokens can shift the model's training focus. This holds true regardless of whether the language~(ro) is well-supported by the model or not~(bn). The influence of these additional tokens is substantial, indicating that the process of enhancing the multilingual capabilities of LLMs is not as straightforward as simply expanding the vocabulary and training with more multilingual data.

\paragraph{Finding: The original vocabulary suffices to present the multilingualism of LLMs.} The LLaMA tokenizer, which utilizes the Byte-level Byte Pair Encoding~(BBPE;~\citealp{bbpe}) algorithm, is the foundation for multilingual language processing tasks. Its universal compatibility across all languages, in conjunction with the absence of the requirement for an ``unknown'' token, optimizes vocabulary sharing~\citep{yuan2023multilingual} and improves its robustness. It allows the model to understand/generate responses in various languages using the same vocabulary. Meanwhile, studies have shown that LLMs trained on unbalanced English-centric datasets, often use English as an internal pivot language. This helps LLMs to map the inputs closer to English in internal space before generating the output~\citep{zhu2024question,huang2024mindmerger,yoon2024langbridge}. Maintaining the original vocabulary helps to preserve this behavior, which also benefits for improving the multilingual capability.

\subsection{Data Augmentation}
\label{sec:aug_analysis}


\paragraph{Setting.} Given a parallel dataset subset~($\mathcal{D}_\mathrm{P}$) from~$\mathcal{D}_\mathrm{para}^A$ that contains translations in all directions for 6 languages~(en,fr,es,zh,ta,th) and a monolingual subset~($\mathcal{D}_\mathrm{M}$) from ~$\mathcal{D}_\mathrm{mono}^A$ for the same 6 languages. We then perform non-repetitive sampling 12,500 sentence pairs from $\mathcal{D}_\mathrm{P}$ in each direction to generate two subsets of parallel corpus data $\mathcal{D}_\mathrm{P_1}$ and $\mathcal{D}_\mathrm{P_2}$, respectively. Consequently, we preserve $\mathcal{D}_\mathrm{P_1}$ and evaluate the effect of augmentation on parallel data $\mathcal{D}_\mathrm{P_2}$ or monolingual data $\mathcal{D}_\mathrm{M}$, resulting in two new dataset, $\mathcal{D}_\mathrm{P_2}'$ and $\mathcal{D}_\mathrm{M}'$, post-augmentation. To assess both the in-domain and out-of-domain capabilities of the model, we perform inference on it using 10 languages (en, fr, es, pt, de, zh, ta,  th, is, zu), utilizing the Flores-101.

\paragraph{Finding: The choice of dictionary is related to the number of entities in the dictionary.} As shown in Table~\ref{tab:dictionary_compare}, there is no clear dictionary preference is observed for en/ta/th/zh-centric translation, with optimal performance randomly distributed across the two dictionaries. Furthermore, we conduct an in-depth analysis of the MUSE and PanLex dictionary for translation from en to another 5 languages. We compare the end-to-end translation performance~(spBLEU), the number of target language entities in the dictionary~(\# entity), and the similarity of entities embedding~(simple average with entity token embeddings) extracted from the trained model. And find a clear correlation between the translation performance and \#entity.


%% file: EMNLP-2024/sections/3.approach.tex
\section{Training Data Construction}

To build powerful LLMs that support translation across a hundred languages, it is crucial to collect and construct a sufficient amount of data.

\subsection{Components of Training Data}

\begin{algorithm*}[!t]
\footnotesize
\KwIn{$A$: all language list. $\mathcal{D}_\mathrm{mono}^A$: monolingual data for all languages. $\mathcal{D}_\mathrm{En}$: an English monolingual data. $\mathcal{D}_\mathrm{para}^A$: a parallel data for all translation directions. Notably, $\mathcal{D}_\mathrm{mono}^A \bigcap \mathcal{D}_\mathrm{En} = \varnothing$. $\vect{x}$: a single data point. $g(\vect{x};\vect{\varphi)}$: A translation model with parameter $\vect{\varphi}$. In a parallel sentence pair, $s$ represents the language of the source sentence, and $t$ represents the language of the target sentence. $f(\vect{x};\vect{\theta})$: a large language model with parameter $\vect{\theta}$. $h(\vect{x}, z)$: augmentation function $h$ enhances input sentence $\vect{x}$ using the dictionary $z$. }

\KwOut {$\mathcal{D}_\mathrm{train}$: a training dataset for current training epoch.}

$\mathcal{D}_\mathrm{train} = \{\}$ \\
\For{$s \in A$}{
    $\mathcal{D}_\mathrm{mono}^{s} \subset \mathcal{D}_\mathrm{mono}^A $ \textcolor{blue}{// Extract a $s$-specific monolingual subset}  \\
    \For{$t \in A$}{
        $\mathcal{D}_\mathrm{para} \leftarrow \mathcal{D}_\mathrm{para}^{s \rightarrow t} \cup \mathcal{D}_\mathrm{para}^{ t \rightarrow s}$   \\
        $\mathcal{D}_\mathrm{para}^{s} \subset \mathcal{D}_\mathrm{para} $ \textcolor{blue}{// Extract the $s$-centric parallel subset} \\
        \uIf{$|\mathcal{D}_\mathrm{para}^s| < 25,000$ }{  
        \textcolor{blue}{// The quantity of 25,000 determined by the machine's memory capacity} \\
            $\mathcal{D}_\mathrm{En}^{s} \subset \mathcal{D}_\mathrm{En}$, s.t. $|\mathcal{D}_\mathrm{En}^{s}| = 25,000 - |\mathcal{D}_\mathrm{para}^s|$ \textcolor{blue}{// Extract an English subset for $s$ language} \\
            $\mathcal{D}_\mathrm{En}^{s \rightarrow t} \leftarrow g(\vect{x};\vect{\varphi})$ or $\mathcal{D}_\mathrm{En}^{t \rightarrow s} \leftarrow g(\vect{x};\vect{\varphi})$, where $\vect{x} \in \mathcal{D}_\mathrm{En}^{s}$ \\
            $\mathcal{D}_\mathrm{aug}^{s \rightarrow t} \leftarrow h(\vect{x}, z) $, where $\vect{x} \in \mathcal{D}_\mathrm{En}^{s \rightarrow t} $, or 
            $\mathcal{D}_\mathrm{aug}^{t \rightarrow s} \leftarrow h(\vect{x}, z) $, where $\vect{x} \in \mathcal{D}_\mathrm{En}^{t \rightarrow s}  $ \\
            $\mathcal{D}_\mathrm{aug}^s \leftarrow \mathcal{D}_\mathrm{aug}^{s \rightarrow t} \cup \mathcal{D}_\mathrm{aug}^{t \rightarrow s}$ \\
        }
    }
    $\mathcal{D}_\mathrm{train} \leftarrow \mathcal{D}_\mathrm{train} \cup \mathcal{D}_\mathrm{mono}^{s} \cup \mathcal{D}_\mathrm{para}^s \cup  \mathcal{D}_\mathrm{aug}^s $
}
\caption{\footnotesize{Illustration of the Training Data Construction Process During a Single Training Epoch}}
\label{alg: training_algo}
\end{algorithm*}

During the continual pertaining stage, the collected training data covering 102 languages~(refer to $A$, which are all languages supported by Flores-101), mainly consists of two parts: monolingual~($\mathcal{D}_\mathrm{mono}^A$) and parallel~($\mathcal{D}_\mathrm{para}^A$) data. For languages with limited data availability, we generated a pseudo-parallel dataset~($\mathcal{D}_\mathrm{aug}$) with multilingual dictionaries: MUSE~\citep{lample2018unsupervised} and PanLex~\citep{wang-etal-2022-expanding}. 
The whole continual pre-training utilizes over 64 billion tokens. More details on supported languages, dataset description, and data statistics can be found in the Appendix~\ref{sec:data_info}.

\paragraph{Monolingual Data~($\mathcal{D}_\mathrm{mono}^A$).} Our monolingual training data includes 94 languages supported by Flores-101 from MC4~\citep{xue-etal-2021-mt5} and MADLAD~\citep{kudugunta2024madlad}, totaling 40,000,000 sentences. To ensure efficient handling and processing of the data, we use a strategy in which each piece of monolingual data is split into multiple entries, with a block size of 512.

\paragraph{Parallel Data~($\mathcal{D}_\mathrm{para}^A$).} Our parallel data from Lego-MT~\cite{yuan-etal-2023-lego} encompasses 102 languages, forming a total $4,737$ language pairs and $9,474$ translation directions. For each translation direction, denoted as source language~($s$) to target language~($t$), we concatenate each translation pair, merely using a space as a delimiter, to form a single entry for training data. For each language pair, the probability of occurrence for each translation direction, for example, $s \rightarrow t$ and $t \rightarrow s$ is set as 50\%. During the training stage, the gradient is computed for the entire data entry, rather than only for the target sentence. For language pairs that have fewer than 25,000~(bound by machine resources) sentence pairs, we replicate the original data thrice~\citep{muennighoff2023scaling}.

\begin{figure}[!t]
    \centering
    \includegraphics[trim={0cm 3.5cm 0cm 3.3cm},clip,scale=0.25]
    {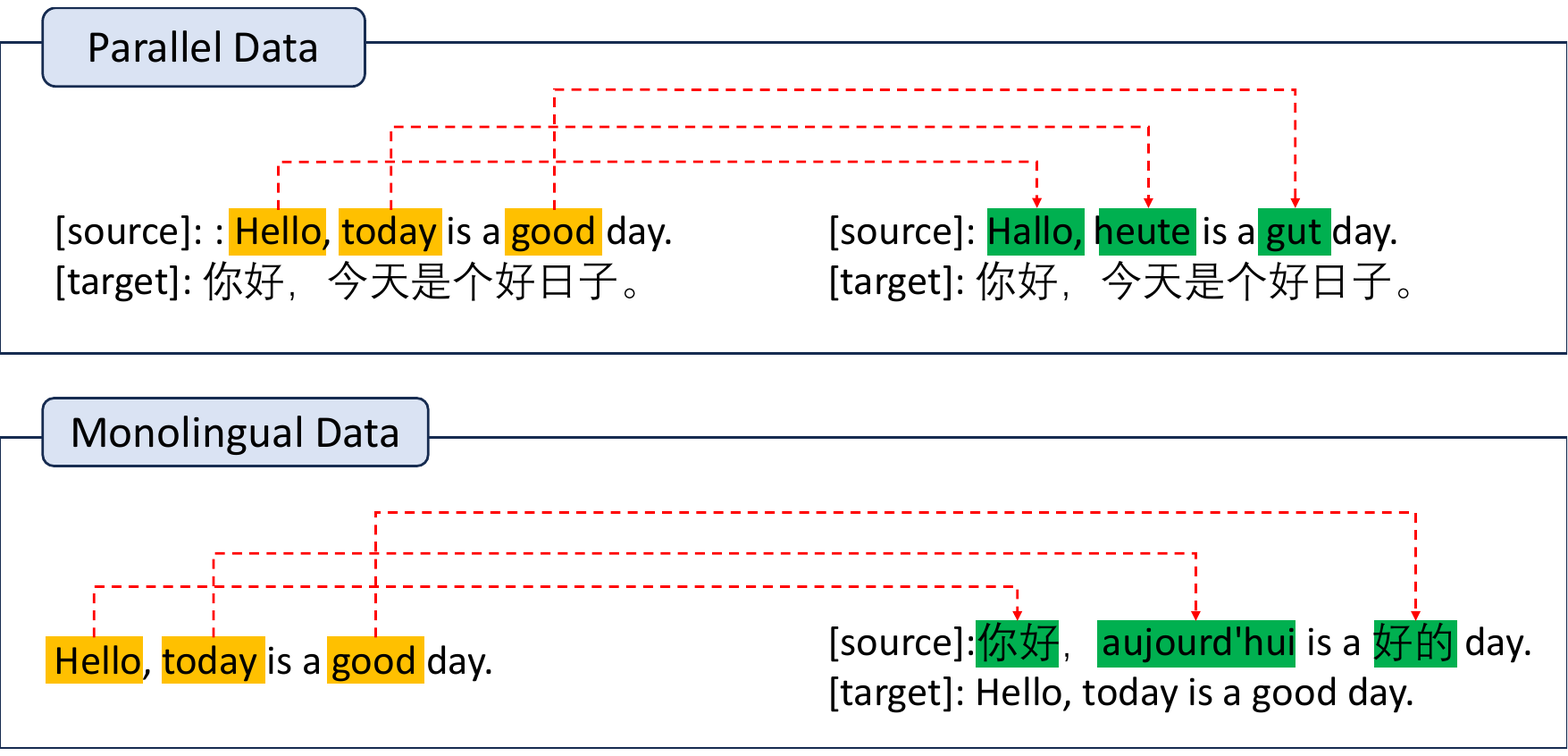}
    \caption{A case illustrating the detailed process of constructing pseudo-parallel data using multilingual dictionary from monolingual or parallel data sources.}
    \label{fig:augumentation}
\end{figure}

\paragraph{Data Generated Through Augmentation~($\mathcal{D}_\mathrm{aug}$).} The way which is followed by~\citet{pan-etal-2021-contrastive}, to obtain code-switch data consists of two steps: 1) build multilingual lexicons; 2) construct pseudo-parallel data. We show the data augmentation process in Figure~\ref{fig:augumentation}.

\input{EMNLP-2024/sections/main_table}

\paragraph{Step 1: Building multilingual lexicons.} The existing multilingual dictionaries, MUSE and PanLex, encompass multiple bilingual dictionaries, such as en-fr, en-de, en-zh bilingual dictionaries.  A dictionary comprises numerous entries, each being a word or a term defined, usage, and provided with other relevant information. We iterate through each entry in the bilingual dictionary, reformat all entries, and create entries in the format of \textit{\{entity\}\_\{language\}}. For instance, the English word ``hello'' as translation in three bilingual dictionaries~(en-fr, en-de, en-zh), leading us to construct a multilingual lexicons entry as \textit{hello\_en, Bonjour\_fr, Hallo\_de, \begin{CJK}{UTF8}{gbsn}{你好}\end{CJK}\_zh}.

\paragraph{Step 2: Constructing pseudo-parallel data.} The foundational data for construction can be based on either parallel or monolingual data, as shown in Figure~\ref{fig:augumentation}. 
For each sentence, we convert it to lowercase and subsequently divide it into multiple words using spaces (for Chinese sentences, the Jieba tokenizer is utilized).  
In parallel data processing, words in a source sentence are randomly replaced with translation from a different language using the multilingual dictionary created in Step~1. 
During the training, the loss is computed solely on the target sentence. In monolingual data processing, each word is individually replaced with a randomly chosen word from the multilingual dictionary. If no suitable replacement word in another language is found, the original word remains unchanged. Consequently, the modified sentence and the original sentence can form pseudo-parallel data. During the training, the loss is computed solely on both the source and the target sentence.


\subsection{Training Algorithm.}

Given an LLM $f(\vect{x};\vect{\vect{\theta}})$ on a collected training data~$\{\x[i]\}_{i=1}^n$, where $\vect{\theta}$ is the pre-trained parameters, our objective is to obtain an LLM through continual pre-training, denoted as $f(\vect{x};\vect{\theta}')$. Here, $\vect{\theta}'$ indicates the updated parameters. The target of $f(\vect{x};\vect{\theta}')$ is to preserve the general capabilities of the model in high-resource languages while simultaneously enhancing the translation performance across all translation directions among 102 languages. 
The process of constructing training data is outlined in Algorithm~\ref{alg: training_algo}. 
We gather monolingual data for each of the languages and parallel data for every translation direction. 
In particular, there is no augmentation for translations involving high-resource languages. 
Instead, we solely augment the translation data that is insufficient by utilizing a trained translation model, Lego-MT model. Then we train the $f(\vect{x}; \vect{\theta})$, the loss function is:
\begin{equation}
      \arg\max_{\vect{\theta}}~ \sum_{i=1}^n \sum_{t=1}^{T_i} \log f(\xt[i]{t}|\xp[i]{t}; \vect{\theta})
\end{equation}
where $T$ is the total decoding time step.

After continual pre-training, we perform instruction tuning on \textbf{\method} using Alpaca~\citep{alpaca}, a dataset comprising 52,000 English instruction examples. This process enhances the model's capability to comprehend and follow instructions without introducing additional multilingual information, resulting in \textbf{\method-Alpaca}. We are currently using Alpaca to enhance the model's capacity for instruction following. In the future, we will release a more robust instruction model fine-tuned with a multilingual instruction dataset.

%% file: EMNLP-2024/sections/main_table.tex
\begingroup
\renewcommand{\arraystretch}{1.2} 
\begin{table*}[!ht]
\centering
\footnotesize
\resizebox{\linewidth}{!}{
\begin{tabular}{l|r|cccccccccccccc}
\toprule
\rowcolor{gray!20}
&  & \multicolumn{2}{c}{\textbf{en-X}}                    & \multicolumn{2}{c}{\textbf{zh-X}}                    & \multicolumn{2}{c}{\textbf{de-X}}                    & \multicolumn{2}{c}{\textbf{ne-X}}                    & \multicolumn{2}{c}{\textbf{ar-X}}                    & \multicolumn{2}{c}{\textbf{az-X}}                    & \multicolumn{2}{c}{\textbf{ceb-X}}                   \\

\rowcolor{gray!20}
\multirow{-2}{*}{\textbf{System}}                        &               \multirow{-2}{*}{\textbf{Size}}                    & \textbf{COMET}                & \textbf{BLEU}               & \textbf{COMET}                & \textbf{BLEU}                 & \textbf{COMET}                & \textbf{BLEU}                 & \textbf{COMET}                & \textbf{BLEU}                 & \textbf{COMET}                & \textbf{BLEU}                 & \textbf{COMET}                & \textbf{BLEU}                 & \textbf{COMET}    & \textbf{BLEU}              \\
\midrule
\multicolumn{16}{l}{\textbf{Encoder-Decoder Models}} \\
\midrule
M2M-100$^*$~\cite{fan2021beyond}                & 418M   & 63.76 & 17.26 & 61.41 & 10.13 & 61.62 & 14.10 & 46.98 & 4.03 & 59.97 & 11.52 & 45.75 & 4.17 & 44.23 & 6.13 \\               
M2M-100$^*$~\cite{fan2021beyond}                 & 1.2B  & 70.00 & 21.54 & 67.29 & 13.13 & 67.62 & 17.73 & 56.04 & 7.14 & 62.62 & 12.57 & 52.39 & 6.06 & 52.79 & 9.46 \\              
M2M-100$^*$~\cite{fan2021beyond}                 & 12B  & 74.19 & 24.74 & 71.56 & 14.91 & 72.07 & 20.34 & 62.19 & 9.68 & 68.91 & 16.36 & 54.78 & 6.24 & 60.09 & 12.48 \\                 
Lego-MT$^*$~\cite{yuan-etal-2023-lego}           & 1.2B & 69.49 & 24.96 & 68.23 & 16.28 & 69.20 & 21.42 & 68.37 & 16.98 & 65.57 & 18.38 & 65.69 & 13.51 & 58.21 & 16.83 \\   
NLLB-200~\cite{nllb2022} & 1.3B              & \underline{81.69} & \underline{31.77} & \underline{78.05} & \underline{19.61} & \underline{79.49} & \underline{25.99} & \underline{81.63} & \underline{23.65} & \underline{78.66} & \underline{24.32} & \underline{78.46} & \underline{19.18} & \underline{76.50} & \underline{23.71} \\
MADLAD-400~\cite{kudugunta2024madlad}       & 7B  & 77.79 & 29.19 & 74.07 & 18.23 & 74.73 & 23.15 & 72.74 & 17.74 & 74.53 & 22.14 & 61.29 & 9.92 & 64.44 & 15.29 \\             
Aya-101~\cite{ustun2024aya}                 & 13B  & 77.26 & 24.30 & 75.29 & 15.50 & 76.17 & 20.86 & 77.78 & 18.65 & 74.82 & 18.44 & 75.36 & 15.46 & 71.90 & 18.76                \\

\midrule
\multicolumn{16}{l}{\textbf{LLM based Decoder-Only Models}} \\
\midrule

LLaMA2~\cite{LLaMA2}                 & 7B  & 43.95 & 4.21 & 44.62 & 0.91 & 45.26 & 2.14 & 38.22 & 0.39 & 39.43 & 0.54 & 47.43 & 0.68 & 33.50 & 1.49     \\
LLaMA2~\cite{LLaMA2} & 13B  &  31.37 & 0.24 & 34.91 & 0.25 & 31.22 & 0.10 & 35.32 & 0.21 & 32.34 & 0.11 & 36.03 & 0.17 & 30.84 & 0.17\\
LLaMA3~\cite{llama3modelcard} & 8B & 45.04 & 3.84 & 45.14 & 3.50 & 42.11 & 3.27 & 44.15 & 2.65 & 39.36 & 2.36 & 43.00 & 1.86 & 36.06 & 2.43 \\
LLaMA2-Alpaca~\cite{alpaca}            & 7B     & 52.83 & 9.44 & 51.29 & 3.80 & 51.47 & 6.82 & 46.59 & 1.31 & 46.76 & 2.84 & 48.63 & 1.36 & 41.02 & 2.69\\ 
LLaMA2-Alpaca~\cite{alpaca} & 13B  & 57.16 & 11.85 & 53.93 & 6.25 & 54.70 & 9.42 & 51.47 & 3.11 & 50.73 & 5.23 & 50.68 & 2.74 & 47.86 & 4.96 \\
LLaMA3-Alpaca~\cite{alpaca} & 8B  & 67.97 & 17.23 & 64.65 & 10.14 & 64.67 & 13.62 & 62.95 & 7.96 & 63.45 & 11.27 & 60.61 & 6.98 & 55.26 & 8.52  \\

PolyLM~\cite{wei2023polylm}         & 13B  &  45.16 & 5.72 & 52.41 & 1.42 & 47.89 & 3.59 & 38.00 & 0.45 & 45.82 & 1.04 & 38.65 & 0.57 & 29.74 & 0.77  \\
Yayi2~\cite{luo2023yayi}            & 30B  & 54.13 & 7.80 & 55.23 & 4.38 & 56.48 & 4.72 & 47.88 & 0.92 & 49.45 & 1.73 & 53.06 & 1.23 & 36.75 & 1.87  \\
TowerInstruct~\cite{alves2024tower} & 7B  & 58.69 & 9.41 & 57.75 & 4.15 & 58.31 & 6.79 & 51.42 & 2.07 & 50.76 & 3.35 & 48.01 & 1.79 & 41.69 & 3.36 \\    
Aya-23~\cite{aryabumi2024aya}       & 8B & 57.91 & 11.18 & 56.65 & 7.20 & 55.69 & 9.30 & 51.78 & 3.50 & 55.49 & 8.00 & 51.45 & 3.27 & 44.14 & 4.24 \\ 
Qwen2-Instruct~\citep{qwen2} & 7B & 59.64 & 9.61 & 59.70 & 6.84 & 57.44 & 7.69 & 58.62 & 4.40 & 57.22 & 6.35 & 54.49 & 3.83 & 49.61 & 3.76 \\ 
ChineseLLaMA2-Alpaca~\cite{cui2024efficient}                   & 7B      & -           & -               &  49.72   & 2.31          & -           & -          & -           & -          & -           & -          & -           & -          & -           & -                         \\
\midrule
\name-Alpaca                   & 7B  &  \textbf{76.66} & \textbf{23.17} & \textbf{73.54} & 14.17 & \textbf{73.82} & \textbf{18.96} & 74.64 & 14.49 & 72.00 & 15.82 & 70.91 & 11.34 & 68.67 & 15.53    \\
\latest-Alpaca & 8B & 75.52 & 22.77 & 73.16 & \textbf{14.43} & 73.47 & 18.95 & \textbf{75.13} & \textbf{15.32} & \textbf{72.29} & \textbf{16.42} & \textbf{72.06} & \textbf{12.41} & \textbf{68.88} & \textbf{15.85} \\

\toprule
\rowcolor{gray!20}
&    & \multicolumn{2}{c}{\textbf{X-en}}                    & \multicolumn{2}{c}{\textbf{X-zh}}                    & \multicolumn{2}{c}{\textbf{X-de}}                    & \multicolumn{2}{c}{\textbf{X-ne}}                    & \multicolumn{2}{c}{\textbf{X-ar}}                    & \multicolumn{2}{c}{\textbf{X-az}}                 & \multicolumn{2}{c}{\textbf{X-ceb}}                   \\
 
\rowcolor{gray!20}
\multirow{-2}{*}{\textbf{System}}  &    \multirow{-2}{*}{\textbf{Size}}                       & \textbf{COMET}                & \textbf{BLEU}                 & \textbf{COMET}                & \textbf{BLEU}                 & \textbf{COMET}                & \textbf{BLEU}                 & \textbf{COMET}                & \textbf{BLEU}                 & \textbf{COMET}                & \textbf{BLEU}                 & \textbf{COMET}                & \textbf{BLEU}                 & \textbf{COMET}    &  \textbf{BLEU}              \\
\midrule
  \multicolumn{16}{l}{\textbf{Encoder-Decoder Models}} \\
\midrule
M2M-100$^*$~\cite{fan2021beyond}            & 418M   & 68.47 & 21.19 & 62.15 & 10.34 & 60.19 & 14.25 & 40.43 & 1.30 & 63.33 & 11.53 & 49.74 & 2.44 & 47.80 & 4.85 \\  
M2M-100$^*$~\cite{fan2021beyond}            & 1.2B & 73.06 & 26.26 & 67.91 & 12.94 & 67.78 & 19.33 & 42.60 & 1.40 & 60.28 & 8.57 & 55.86 & 4.58 & 55.87 & 6.83 \\
M2M-100$^*$~\cite{fan2021beyond}            & 12B & 74.45 & 28.01 & 69.27 & 13.35 & 70.17 & 21.31 & 45.50 & 2.85 & 69.94 & 15.15 & 61.36 & 6.44 & 57.07 & 8.77 \\  
Lego-MT$^*$~\cite{yuan-etal-2023-lego}      & 1.2B & 75.44 & 30.71 & 71.41 & 16.42 & 70.75 & 23.75 & 59.66 & 15.02 & 70.73 & 18.21 & 66.73 & 11.88 & 59.28 & 15.06 \\ 
NLLB-200~\cite{nllb2022} & 1.3B & \underline{84.22} & \underline{38.60} & 76.75 & 15.27 & \underline{79.50} & 25.71 & \underline{73.70} & \underline{21.84} & \underline{79.85} & 21.80 & \underline{80.02} & \underline{15.55} & \underline{69.05} & \underline{24.72} \\
MADLAD-400~\cite{kudugunta2024madlad}    & 7B & 83.05 & 38.14 & 78.49 & 20.48 & 77.50 & \underline{26.79} & 61.94 & 13.93 & 77.84 & \underline{22.25} & 75.41 & 13.85 & 51.33 & 4.24 \\           
Aya-101~\cite{ustun2024aya}               & 13B & 80.72 & 31.92 & \underline{78.51} & \underline{22.49} & 77.37 & 15.43 & 69.69 & 17.13 & 77.90 & 16.54 & 78.70 & 13.51 & 67.76 & 21.58                    \\
\midrule
\multicolumn{16}{l}{\textbf{LLM Based Decoder-Only Models}} \\
\midrule
LLaMA2~\cite{LLaMA2}                & 7B & 55.46 & 11.80 & 43.50 & 0.55 & 43.10 & 3.22 & 34.41 & 0.42 & 39.13 & 0.25 & 43.98 & 0.59 & 41.64 & 1.16 \\
LLaMA2~\cite{LLaMA2} & 13B & 38.25 & 0.75 & 37.06 & 0.22 & 31.73 & 0.25 & 30.13 & 0.15 & 33.68 & 0.06 & 33.47 & 0.08 & 37.49 & 0.20 \\
LLaMA3~\cite{llama3modelcard} & 8B & 67.66 & 19.81 & 42.52 & 1.37 & 49.42 & 6.61 & 33.38 & 0.52 & 34.12 & 0.49 & 37.27 & 0.79 & 37.97 & 1.41 \\
LLaMA2-Alpaca~\cite{alpaca}            & 7B & 65.85 & 16.44 & 56.53 & 4.46 & 56.76 & 9.01 & 34.96 & 1.03 & 44.10 & 2.18 & 40.67 & 0.63 & 45.69 & 1.73 \\
LLaMA2-Aplaca~\cite{alpaca} & 13B  & 68.72 & 19.69 & 64.46 & 8.80 & 62.86 & 12.57 & 38.88 & 2.16 & 52.08 & 4.48 & 41.18 & 0.87 & 48.47 & 2.51 \\ 
LLaMA3-Alpaca~\cite{alpaca} & 8B  & 77.43 & 26.55 & 73.56 & 13.17 & 71.59 & 16.82 & 46.56 & 3.83 & 66.49 & 10.20 & 58.30 & 4.81 & 52.68 & 4.18 \\
PolyLM~\cite{wei2023polylm}         & 13B & 50.98 & 7.75 & 42.60 & 1.20 & 43.95 & 3.69 & 33.69 & 0.36 & 42.27 & 1.67 & 40.24 & 0.44 & 39.29 & 0.96 \\
Yayi2~\cite{luo2023yayi}            & 30B & 68.06 & 19.37 & 57.81 & 6.07 & 53.82 & 5.62 & 40.95 & 0.48 & 46.61 & 0.52 & 49.29 & 0.71 & 45.50 & 1.71 \\
TowerInstruct~\cite{alves2024tower} & 7B & 65.37 & 18.87 & 64.26 & 10.37 & 60.73 & 12.81 & 38.80 & 0.62 & 44.72 & 0.39 & 47.17 & 0.71 & 47.15 & 2.24 \\ 
Aya-23~\cite{aryabumi2024aya}        & 8B & 67.53 & 20.57 & 66.11 & 11.20 & 63.09 & 14.09 & 44.33 & 2.69 & 63.59 & 11.84 & 46.97 & 1.19 & 45.17 & 2.29 \\
Qwen2-Instruct~\citep{qwen2} & 7B  & 73.25 & 19.04 & 72.52 & 13.52 & 64.61 & 11.33 & 41.41 & 2.27 & 64.94 & 8.50 & 47.96 & 1.66 & 55.45 & 3.00 \\
ChineseLLaMA2-Alpaca~\cite{cui2024efficient}       & 7B                    & -           & -               &  55.06   & 6.15          & -           & -          & -           & -          & -           & -          & -           & -          & -           & -           \\

\midrule
\name-Alpaca    & 7B & 80.55 & 30.63 & 75.52 & 13.53 & 74.47 & 19.26 & \textbf{67.36} & \textbf{15.47} & 75.40 & 15.32 & \textbf{72.03} & \textbf{10.27} & \textbf{65.05} & \textbf{16.11} \\
\latest-Alpaca & 8B & \textbf{81.28} & \textbf{31.85} & \textbf{78.34} & \textbf{16.46} & \textbf{76.23} & \textbf{20.64}& 65.83 & 14.16 & \textbf{75.84} & \textbf{15.45} & 70.61 & 9.32 & 63.35 & 12.66\\

\bottomrule
\end{tabular}}
\caption{Comparison with different architecture, including \textbf{encoder-decoder} and  \textbf{decoder-only} models, on Flores-101 dataset, where X refers to any language in 101 languages. $^*$ refers to that model comparisons are restricted to 85 languages, denoted as |X| = 85. We make this choice because the M2M-100 baselines cover only 86 languages, as reported in the work by Flores-101~\citep{flores101,yuan-etal-2023-lego}. This table compares our instruction-aligned \name model (\name-Alpaca) with the instruction-aligned LLaMA2 model (LLaMA2-Alpaca) to demonstrate the benefits of our multilingual continual pre-training. Additionally, we compare \method with other open-source multilingual-focus LLMs to highlight the impressive multilingual capabilities.}
\label{tab:flores}
\end{table*}
\endgroup

\begingroup
\renewcommand{\arraystretch}{1.2} 
\begin{table*}[!ht]
\centering
\footnotesize
\resizebox{\linewidth}{!}{
\begin{tabular}{l|r|cccc|cc|cccc}
\toprule
\multirow{2}{*}{\textbf{System}} & \multirow{2}{*}{\textbf{Size}} & \multicolumn{2}{c}{\textbf{TED (en-X)}} & \multicolumn{2}{c|}{\textbf{TED (X-en)}} & \multicolumn{2}{c|}{\textbf{TICO (en-X)}}  & \multicolumn{2}{c}{\textbf{WMT23 (en-X)}} & \multicolumn{2}{c}{\textbf{WMT23 (X-en)}}                     \\
                       &                  & \textbf{COMET}                & \textbf{BLEU}                 & \textbf{COMET}                & \textbf{BLEU}                 & \textbf{COMET}                & \textbf{BLEU}                 & \textbf{COMET}                & \textbf{BLEU}                 & \textbf{COMET}  & \textbf{BLEU}                \\
\toprule

LLaMA2~\cite{LLaMA2}               & 7B    & 52.15	&	3.34	&	61.54	&	8.66	&	39.63	&	3.45	&	51.55	&	2.96	&	65.68	&	14.87   \\
LLaMA2~\cite{LLaMA2} & 13B & 34.66	&	0.17	&	40.87	&	0.49	&	31.65	&	0.42	&	33.74	&	0.43	&	41.18	&	0.85\\ 
LLaMA3~\cite{llama3modelcard} & 8B & 44.72	&	2.09	&	53.56	&	6.04	&	40.02	&	4.82	&	47.44	&	2.61	&	55.18	&	7.84\\
LLaMA2-Alpaca~\cite{alpaca}        & 7B    & 62.04	&	9.15	&	68.62	&	12.67	&	44.73	&	8.60	&	73.17	&	17.23	&	75.82	&	24.97   \\
LLaMA2-Alpaca~\cite{alpaca} & 13B  & 65.62	&	11.40	&	70.74	&	14.54	&	48.64	&	10.79	&	77.93	&	21.60	&	77.90	&	28.67 \\
LLaMA3-Alpaca~\cite{alpaca} & 8B & 73.20	&	14.13	&	75.03	&	16.83	&	56.73	&	14.49	&	80.05	&	24.11	&	79.22	&	29.76 \\

PolyLM~\cite{wei2023polylm}        & 13B   & 50.18	&	5.53	&	55.16	&	7.28	&	40.36	&	7.17	&	62.67	&	10.62	&	69.15	&	19.09    \\
Yayi2~\cite{luo2023yayi}           & 30B   & 61.53	&	8.54	&	70.92	&	14.09	&	47.02	&	7.91	&	65.69	&	10.76	&	75.60	&	20.47  \\
TowerInstruct~\cite{alves2024tower}& 7B    & 64.83	&	8.22	&	70.91	&	15.29	&	50.48	&	10.14	&	74.03	&	18.42	&	80.08	&	30.03   \\
Qwen2-Instruct~\citep{qwen2} & 7B & 66.68	&	8.84	&	71.83	&	13.37	&	55.16	&	11.47	&	75.11	&	18.86	&	77.48	&	25.61 \\
Aya-23~\cite{aryabumi2024aya}      & 8B    & 68.06	&	10.69	&	72.87	&	16.44	&	52.44	&	12.98	&	\textbf{83.29}	&	\textbf{27.15}	&	\textbf{82.00}	&	\textbf{31.21}   \\
\midrule
\name-Alpaca               & 7B    & \textbf{75.58}	&	\textbf{16.12}	&	76.18	&	17.81	&	\textbf{68.33}	&	19.79	&	80.17	&	23.91	&	79.55	&	30.30  \\      
\latest-Alpaca  & 8B & 74.95	&	15.15	&	\textbf{76.99}	&	\textbf{18.47}	&	67.71	&	\textbf{20.06}	&	79.96	&	24.49	&	79.88	&	30.34\\
\bottomrule
\end{tabular}}
\caption{Benchmarking results on \textsc{WMT23}, \textsc{TED} and \textsc{TICO} dataset. X denotes various languages across different translation benchmarks; detailed information is available in Appendix~\ref{sec:data_info}. Evaluation results across these benchmarks further validate the strong multilingual translation capabilities of \method.}
\label{tab:mix}
\end{table*}
\endgroup

\begin{table*}[!t]
\centering
\footnotesize
\resizebox{0.9\linewidth}{!}{
\begin{tabular}{l|ccc|cc|cc|cc|c}
\toprule
\multicolumn{1}{l}{}  & \multicolumn{3}{|c|}{\textbf{Knowledge}} & \multicolumn{2}{c|}{\textbf{Commonsense Reasoning}} & \multicolumn{2}{c|}{\textbf{Math Reasoning}} & \multicolumn{2}{c|}{\textbf{Code}} &  \multirow{2}{*}{\textbf{Avg.}}     \\
                      & \textbf{MMLU}     & \textbf{BBH}      & \textbf{NQ}      & \textbf{HellaSwag}             & \textbf{Winogrande}             & \textbf{GSM8K}            & \textbf{Math}            & \textbf{HumanEval}     & \textbf{MBPP}     &   \\
\midrule
LLaMA2-Alpaca         & 44.22    & 37.95    & 24.32   & 31.12                 & 61.09             & 14.03            & 3.82            & 14.63         & 27.63    & 28.76 \\
\name-Alpaca & 44.60    & 38.25    & 23.21   & 33.75                 & 61.48             & 12.21            & 3.74            & 12.20          & 25.29    & 28.30 \\
\bottomrule
\end{tabular}}
\caption{Evaluation results, assessed by OpenCompass~\citep{2023opencompass}, on monolingual general benchmarks.}
\label{tab:general}
\end{table*}

%% file: EMNLP-2024/sections/5.experiments.tex
\section{Benchmarking Results}
In this section, we present multilingual benchmarking results to comprehensively demonstrate the potential of \name. We evaluate translation quality with spBLEU~\citep{flores101} and COMET-22~\citep{rei-etal-2020-comet} for both LLMs and translation models. See Appendix~\ref{sec:model_info} for training details on \name and description of baseline models.

\paragraph{We significantly enhances the multilingual translation capabilities of the base LLaMA2 model through massive multilingual continual pre-training.}
The benefits of our continual pre-training is enhancing the base LLM's multilingual translation capabilities.
Evaluation results on Flores-101 benchmark are shown in Table~\ref{tab:flores}. 
By comparing our multilingual-enhanced model with the base LLaMA2 model in instruction-tuned versions (\name-Alpaca vs. LLaMA2-Alpaca), we consistently observe a significant performance improvement on both English-centric and non-English-centric translation.
In addition to Flores-101, we also make evaluation on a range of diverse translation benchmarks (Table~\ref{tab:mix}).
The performance enhancement brought by our multilingual continual pre-training is consistent across these benchmarks.

\paragraph{\method outperforms other open-source decoder-only LLMs on multilingual translation by a large margin.}
Next, we compare \name-Alpaca model with other open-source decoder-only LLMs built for multilingual purposes (Table~\ref{tab:flores}, Table~\ref{tab:mix}).
Compared to other from-scratch trained LLMs, such as PolyLM, Yayi2, \name consistently shows better performance across various multilingual translation benchmarks, indicating that the LLaMA2 base model provides a strong foundation for language extension.
Furthermore, when compared to other LLaMA-based continual pre-trained models, such as TowerInstruct, \name also achieves superior performance, demonstrating the effectiveness of our optimized continual pre-training pipeline.

\begin{figure}[!t]
    \centering
    \includegraphics[width=0.45\textwidth]{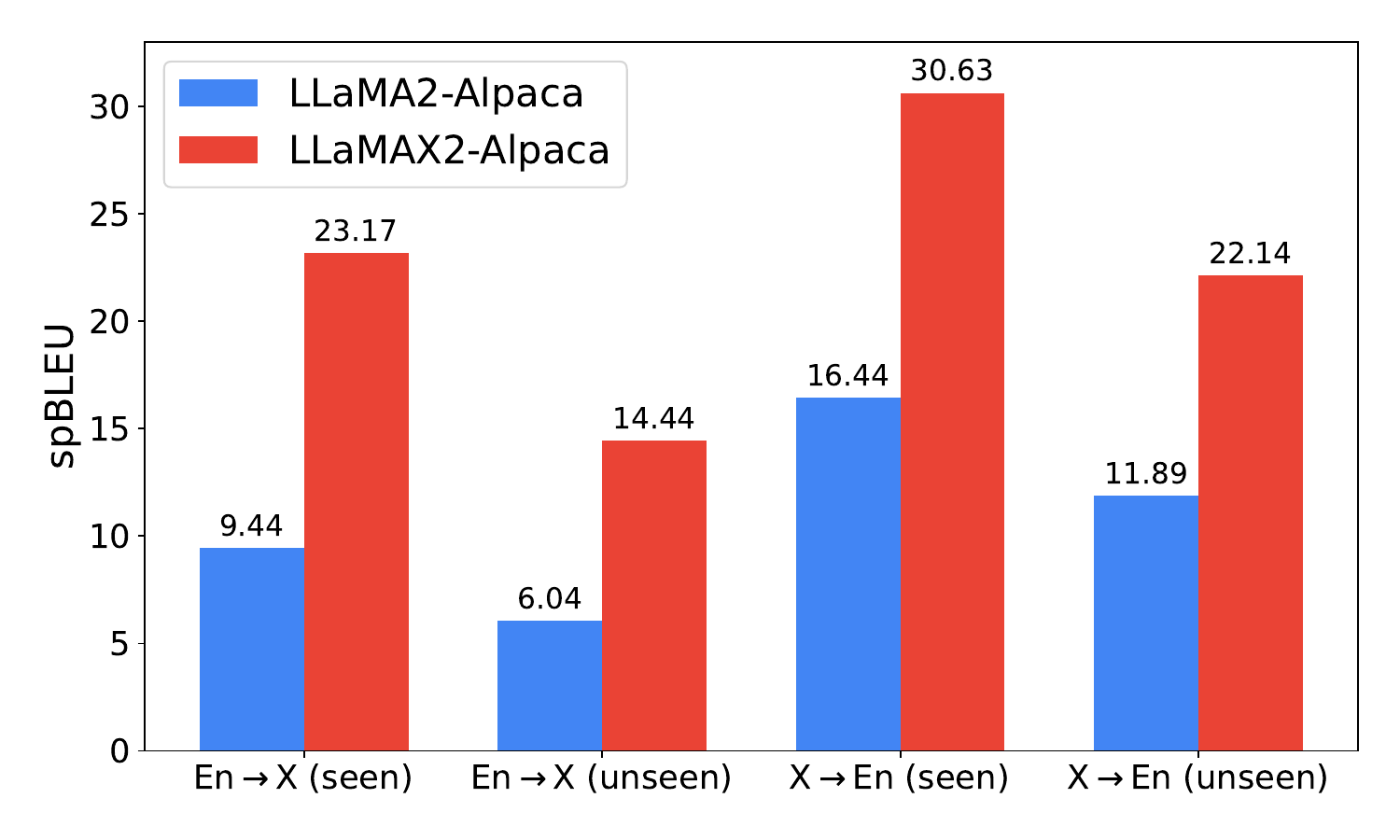}
    \caption{Comparison results between \name-Alpaca and LLaMA2-Alpaca on Flores-200. Some non-English languages are not covered in Flores-200, but \name also boosts its translation performance.}
    \label{fig:unseen}
\end{figure}

\paragraph{\method benefits unseen long-tail low-resource languages as well.}
A significant challenge in multilingual enhancement is that the substantial cost of collecting scarce multilingual resources makes it prohibitive to cover massive languages.
While our multilingual pre-training corpus already covers 102 languages, we acknowledge that there remains a large group of long-tail, low-resource languages that are not well covered.
To assess the generalization capability of \name, we evaluate it on Flores-200 dataset and observe its performance on these unseen languages (Figure~\ref{fig:unseen}). 
We find that for languages not encountered during training, \name still achieves significant improvements, demonstrating the generalization capability of our massive continual pre-training.

\begin{figure}[!t]
    \centering
    \includegraphics[width=0.45\textwidth]{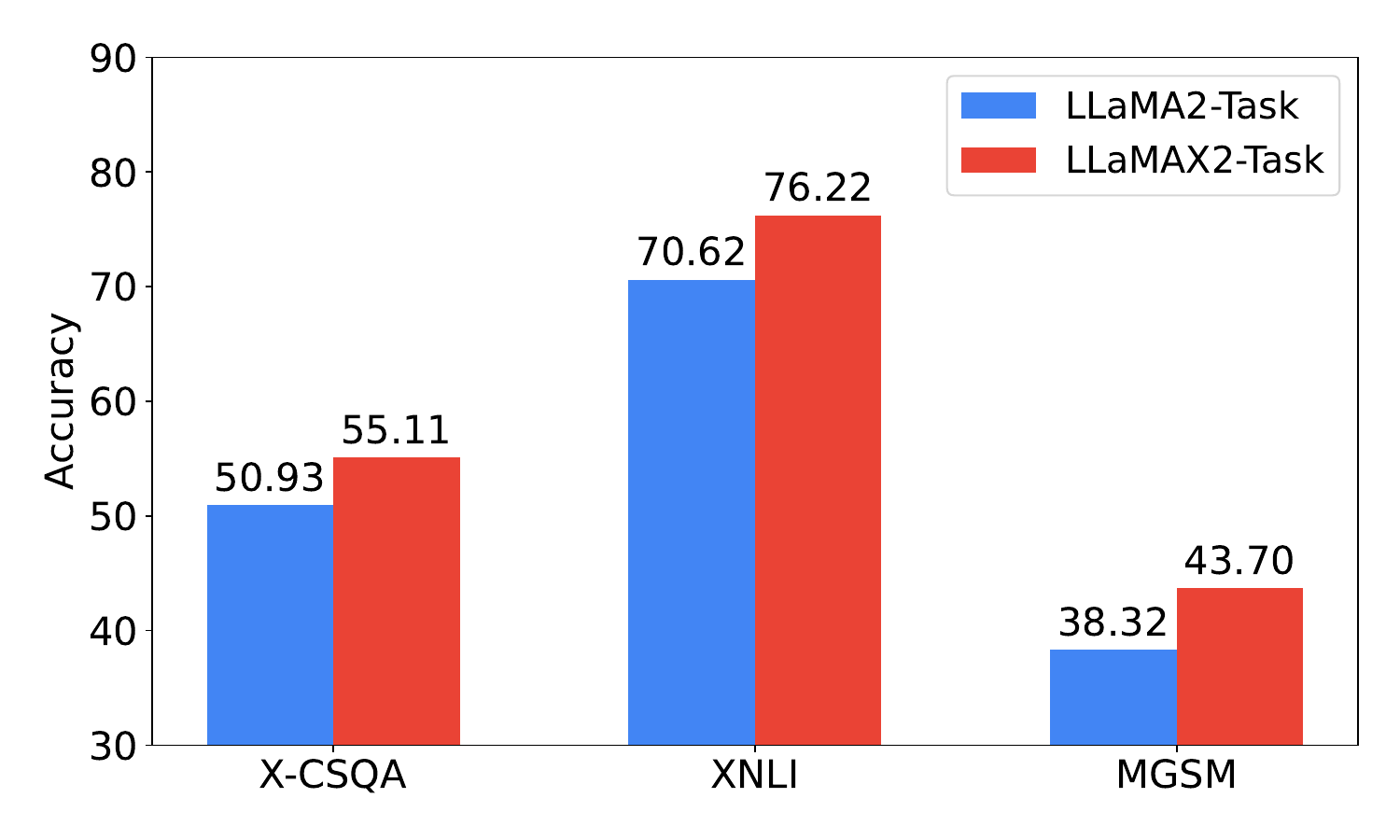}
    \caption{Comparison results between instruction-tuning our multilingual enhanced model and the base model with specialized instruction data. We take \textsc{X-CSQA}, \textsc{XNLI}, \textsc{MGSM} as three examples tasks.}
    \label{fig:specialized}
\end{figure}

\paragraph{\method is closing the performance gap between open-source LLM translator and specialized encoder-decoder translation systems.}
While \name has achieved the state-of-the-art translation performance among open-source decoder-only LLMs, the next critical question is whether we can close the gap between LLMs and specialized encoder-decoder translation systems.
Table~\ref{tab:flores} provides a comprehensive comparison, reveals \name has reached the level of the M2M-100-12B model.
Future work will be needed to optimize the language extension framework to match the performance of advanced translation systems.

\paragraph{\method provides a better starting point for specialized instruction-tuning on English task data.}
In the end, we demonstrate the usage of our continual pre-trained model (\name) on tasks beyond translation.
While in previous experiments we use basic Alpaca instruction data to teach LLM to follow translation instructions, we now show that our released checkpoint can be enpowered to handle more multilingual tasks beyond translation. 
Figure~\ref{fig:specialized} presents three example tasks where we use specialized instruction data to unlock \name's abilities on specific tasks, such as math reasoning and common sense reasoning.
We find that the instruction-tuned \name model outperforms its LLaMA2 model counterpart in non-English performance across all three tasks, demonstrating that provides a better starting point for instruction-tuning with task-specific data.

\begin{table*}[]
    \centering
    \footnotesize
    \resizebox{0.9\linewidth}{!}{
    \begin{tabular}{c|cc|cc|cc|cc}
    \toprule
         \multirow{3}{*}{\textbf{Direct}} &  \multicolumn{4}{c|}{\textbf{spBLEU}} & \multicolumn{4}{c}{\textbf{COMET}} \\
         & \multicolumn{2}{c|}{\textbf{LLaMA3-Alpaca}} & \multicolumn{2}{c|}{\textbf{\name-Alpaca}} & \multicolumn{2}{c|}{\textbf{LLaMA3-Alpaca}} & \multicolumn{2}{c}{\textbf{\name-Alpaca}} \\
         & \textbf{src$\rightarrow$trg} & \textbf{src$\rightarrow$en$\rightarrow$trg} & \textbf{src$\rightarrow$trg} & \textbf{src$\rightarrow$en$\rightarrow$trg} & \textbf{src$\rightarrow$trg} & \textbf{src$\rightarrow$en$\rightarrow$trg} & \textbf{src$\rightarrow$trg} & \textbf{src$\rightarrow$en$\rightarrow$trg}  \\
        \midrule
zh$\rightarrow$X	&	10.14	&	11.34	&	14.17	&	\textbf{15.54}	&	64.65	&	66.61	&	73.54	&	\textbf{74.74} \\ 
X$\rightarrow$zh	&	13.17	&	\textbf{15.37}	&	13.53	&	15.11	&	73.56	&	75.66	&	75.52	&	\textbf{77.21} \\
de$\rightarrow$X	&	13.62	&	14.24	&	18.96	&	\textbf{19.38}	&	64.67	&	65.79	&	73.82	&	\textbf{74.36} \\
X$\rightarrow$de	&	16.82	&	18.08	&	19.26	&	\textbf{20.71}	&	71.59	&	73.11	&	74.47	&	\textbf{76.04} \\
ar$\rightarrow$X	&	11.27	&	12.60	&	15.82	&	\textbf{17.10}	&	63.45	&	65.33	&	72.00	&	\textbf{73.17} \\
X$\rightarrow$ar	&	10.20	&	10.88	&	15.32	&	\textbf{16.00}	&	66.49	&	69.54	&	75.40	&	\textbf{76.32} \\
ne$\rightarrow$X	&	7.96	&	10.29	&	14.49	&	\textbf{16.16}	&	62.95	&	67.87	&	74.64	&	\textbf{76.86} \\
X$\rightarrow$ne	&	3.83	&	7.08	&	15.47	&	\textbf{16.86}	&	46.56	&	58.89	&	67.36	&	\textbf{69.47} \\
az$\rightarrow$X	&	6.98	&	9.52	&	11.34	&	\textbf{13.54}	&	60.61	&	65.16	&	70.91	&	\textbf{73.6}0 \\
X$\rightarrow$az	&	4.81	&	6.96	&	10.27	&	\textbf{11.44}	&	58.30	&	67.52	&	72.03	&	\textbf{75.60} \\
ceb$\rightarrow$X	&	8.52	&	10.69	&	15.53	&	\textbf{16.98}	&	55.26	&	60.71	&	68.67	&	\textbf{70.76} \\
X$\rightarrow$ceb	&	4.18	&	7.17	&	16.11	&	\textbf{18.94}	&	52.68	&	59.55	&	65.05	&	\textbf{66.52} \\
\midrule
Avg.	&	9.29	&	11.19	&	15.02	&	\textbf{16.48}	&	61.73	&	66.31	&	71.95	&	\textbf{73.72} \\

        \bottomrule
    \end{tabular}}
    \caption{We can convert a translation task from the source language (src) to the target language (trg), represented as src$\rightarrow$trg, to src$\rightarrow$en$\rightarrow$trg. The experimental results indicate that the performance of English as a powerful pivot falls short compared to \name-Alpaca (LLaMA3 pivot translation vs. \name-Alpaca). Furthermore, conducting similar pivot translation experiments on \name-Alpaca can further improve translation performance.}
    \label{tab:bridge_exp}
\end{table*}

\paragraph{\method circumvents catastrophic forgetting issue.}
A common concern with continual pre-training on additional multilingual corpus is that the process might disturb the parametric knowledge and working pattern of the original model, a phenomenon known as catastrophic forgetting~\cite{goodfellow2013empirical}.
Furthermore, we compare \name with LLaMA2 on popular English benchmarks that measure a diverse set of core capabilities of LLMs. 
Experiment results in Table~\ref{tab:general} show that the two models achieve very similar performance on these benchmarks~(More details about these benchmarks are in Appendix~\ref{sec:data_info}.), demonstrating that our continual pre-training does not compromise the general capability of the base model.




\paragraph{Beyond the English-centric translation is more efficient and effective.} We further investigate the necessity and feasibility of multilingual augmentation for an English-centric LLM. We can effectively transform a translation task~(src$\rightarrow$trg) from the source language~(src) to the target language~(trg) into src$\rightarrow$en and en$\rightarrow$trg, which allows us to leverage the power of English as a central language, facilitating seamless communication and comprehension across various language pairs. We refer to this experimental setup as a pivot translation experiment.  As shown in Table~\ref{tab:bridge_exp}, the experimental results demonstrate that the pivot translation experiments effectively leverage the power of English to enhance translation performance (compared src$\rightarrow$en$\rightarrow$trg to src$\rightarrow$trg on the same model), although it still falls short of the results obtained from large-scale multilingual continual pre-trained models~(LLaMA3-Alpaca src$\rightarrow$en$\rightarrow$trg vs. \name-Alpaca src$\rightarrow$trg). Interestingly, conducting pivot translation experiments based on \name-Alpaca reveals the potential for significant improvements in translation performance~(\name-Alpaca src$\rightarrow$en$\rightarrow$trg vs. \name-Alpaca src$\rightarrow$trg).

%% file: EMNLP-2024/sections/6.analysis.tex


%% file: EMNLP-2024/sections/2.related_work.tex
\section{Related Work}

\noindent\textbf{Multilingual Large Language Models.} Large Language Model~(LLMs;~\citealp{openai2023gpt4,zhang2022opt,gpt,palm,llama1,LLaMA2}) trained with English-centric data can also solve various non-English tasks~\citep{hendrycks2021ethics,hendryckstest2021,BBH,NQ,hendrycksmath2021}, but the performance between non-English and English is significantly large~\cite{yuan2023multilingual}. Efforts to develop more multilingual LLMs in two different ways: retraining LLMs with diverse multilingual data from scratch~\citep{wei2023polylm}; or continuous training of pre-trained models using language-specific data with the option to expand the vocabulary~\citep{zhao2024llama, cui2024efficient, faysse2024croissantllm,alves2024tower}. Instead of training from scratch, continual pre-training aims at updating pre-trained models with new data, making the process more efficient and cost-effective~\citep{gupta2023continual, alves2024tower,xie2023efficient}.

\noindent\textbf{Multilinguality in LLMs.} Recent research has shed light on the multilingual capabilities of LLMs. A comprehensive survey by~\citet{huang2024survey} discusses various aspects of multilingualism in LLMs, including training and inference methods, model security, multi-domain with languages culture, and emphasizes the need for language-fai technology.~\citet{yuan2023multilingual} analysis multilingualism of LLMs from the vocabulary sharing aspect. ~\citet{zhao2024large} delve into the architecture of LLMs to find how LLMs handle multilingualism. Recently, ~\citet{li2024quantifying} quantify the multilingual performance of LLMs. These studies provide valuable insights into the multilingual capabilities of LLMs, and the key technical design of continual pre-training for \method.

%% file: EMNLP-2024/sections/7.conclusion.tex
\section{Conclusion}

In this work, we enhance the series models of LLaMA translation performance for 102 languages through continual pre-training, creating \method. We compare \method’s translation capabilities with other decoder-only LLMs and encoder-decoder models across multiple benchmarks. \method is also assessed on general tasks and fine-tuned with task-specific instructions. Our results indicate that \method improves translation quality while maintaining general capabilities and can serve as a powerful foundation model for downstream multilingual applications.

%% file: EMNLP-2024/sections/8.appendix.tex
\clearpage

\appendix

\section*{Limitations}
\label{sec:limitation} This work focuses on the discussion of some key technologies, such as the use of vocabulary lists and the determination of data augmentation schemes. However, it does not delve into further processing of the quality of open-source data. We acknowledge a gap in the literature regarding the thorough evaluation of open-source data quality, suggesting an opportunity for future research to improve data preprocessing methods for better model training outcomes.

\section*{Acknowledgments}
\label{sec:ack}
Authors of this paper would like to thank Zixian Huang, Qiushi Sun, Fangzhi Xu, Hanxu Hu, Chuanyang Jin, Yichao Du, and Zichen Ding for giving many helpful comments on previous versions of this paper. We deeply express our gratitude to Shanghai AI Laboratory. This work is partially supported by the National Key R\&D Program of China (NO.2022ZD0160100).

\section*{Outline}

\begin{itemize}[leftmargin=0.3cm]
    
    \item Section~\ref{sec:data_info}: The comprehensive details of the training data, including monolingual and parallel data,  and the evaluation benchmark~(Table~\ref{tab:supported_lgs}). 
    
    \item Section~\ref{sec:model_info}: The detailed information of different models, including open-source Large Language Models~(Section~\ref{sec:llms}) and well-trained translation models~(Section~\ref{sec:translation_models}).

    \item Section~\ref{sec:corr_fert_and_quality}: Analysis the correlation between embedding quality of LLaMA2 and fertility using Flores-101 test~(Figure~\ref{fig:correlation_fertility_with_quality}).

    \item Section~\ref{sec:ks_lottery}: A detailed introduction to the KS-Lottery method.

    \item Section~\ref{sec:select_hop_translation}: Selection about multi-hop translation~(Table~\ref{tab:select_hop} and Table~\ref{tab:hop_example}).

    \item Section~\ref{sec:parallel_format}: The selection of the appropriate format for parallel data during training~(Table~\ref{tab:usage_para}).
    
    \item Section~\ref{sec:gpt4}: The comparison of translation performance across all seven languages between Lego-MT and GPT-4~(Figure~\ref{fig:comparison_gpt-4}).

    \item Section~\ref{sec:compare_lg_specific}: Comparison results between \name-Alpaca with language-specific enhanced LLMs~(Table~\ref{tab:compare_ja}).

    \item Section~\ref{sec:instruction_prompt}: We present comprehensive instructions utilized for all LLMs~(Table~\ref{tab:instruction_example}).
\end{itemize}

\section{Data Information}
\label{sec:data_info}

\begin{table*}[!ht]
    \centering
    \resizebox{\linewidth}{!}{
    \begin{tabular}{c|r|r|rrr || c|r|r|rrr}
\toprule
\textbf{Family}  &   \textbf{ISO}   &   \textbf{Language}   &   \textbf{\# Mono.}   &   \textbf{\# Para.}   &   \textbf{\# Direct.}    &   \textbf{Family}  &   \textbf{ISO}   &   \textbf{Language}   &   \textbf{\# Mono.}   &   \textbf{\# Para.}   &   \textbf{\# Direct.}  \\
\midrule
\multirow{7}{*}{Afro-Asiatic}   &   ha   &   Hausa   &   420,964   &   3,147,704   &   96                           &   \multirow{18}{*}{Indo-European}   &   ne   &   Nepali   &   702,334   &   8,907,527   &   97 \\
   &   om   &   Oromo   &   18,895   &   191,319   &   96                             &      &   or   &   Odia   &   100,530   &   812,235   &   97 \\
   &   so   &   Somali   &   697,864   &   3,804,551   &   97                          &      &   pa   &   Punjabi   &   513,987   &   3,737,780   &   97 \\
   &   am   &   Amharic   &   269,171   &   4,031,552   &   97                         &      &   sd   &   Sindhi   &   472,217   &   821,996   &   95 \\
   &   ar   &   Arabic   &   716,063   &   9,940,756   &   97                          &      &   ur   &   Urdu   &   711,354   &   4,137,619   &   97 \\
   &   he   &   Hebrew   &   300,000   &   3,928,938   &   96                          &      &   fa   &   Persian   &   721,307   &   4,111,536   &   97 \\
   &   mt   &   Maltese   &   671,716   &   1,518,533   &   94                         &      &   ku   &   Kurdish   &   517,239   &   3,597,863   &   97 \\
\cline{1-6}
\multirow{2}{*}{Austroasiatic}   &   km   &   Khmer   &   687,690   &   4,044,652   &   97                          &      &   ps   &   Pashto   &   588,340   &   3,717,480   &   97 \\
   &   vi   &   Vietnamese   &   760,472   &   4,112,089   &   97                     &      &   tg   &   Tajik   &   700,237   &   4,131,709   &   97 \\
\cline{1-6}
\multirow{6}{*}{Austronesian}   &   jv   &   Javanese   &   505,619   &   2,799,761   &   97                        &      &   ast   &   Asturian   &   0   &   1,535,714   &   96 \\
   &   id   &   Indonesian   &   707,962   &   4,243,235   &   97                      &      &   ca   &   Catalan   &   724,597   &   4,145,004   &   97 \\
   &   ms   &   Malay   &   711,895   &   4,121,713   &   97                           &      &   es   &   Spanish   &   706,307   &   4,258,477   &   98 \\
   &   mi   &   Maori   &   180,678   &   3,437702   &   97                           &      &   fr   &   French   &   787,316   &   4,290,003   &   99 \\
   &   ceb   &   Cebuano   &   418,058   &   2,217,926   &   91                        &      &   gl   &   Galician   &   726,512   &   3,131,730   &   96 \\
   &   tl   &   Tagalog   &   0   &   3,927,576   &   97                              &      &   it   &   Italian   &   846,107   &   4,233,108   &   96 \\
\cline{1-6}
\multirow{4}{*}{Dravidian}   &   te   &   Telugu   &   708,459   &   4,219,702   &   97            &      &   oc   &   Occitan   &   36,379   &   1,752,951   &   95 \\
   &   kn   &   Kannada   &   712,832   &   3,592,636   &   97                                     &      &   pt   &   Portuguese   &   795,818   &   4,258,604   &   97 \\
   &   ml   &   Malayalam   &   715,387   &   4,516,012   &   97                                   &      &   ro   &   Romanian   &   702,002   &   4,219,414   &   97 \\
\cline{7-12}
   &   ta   &   Tamil   &   711,863   &   4,444,734   &   97                                       &   Japonic   &   ja   &   Japanese   &   726,455   &   4,207,728   &   97 \\
\cline{1-6} \cline{7-12}
\multirow{32}{*}{Indo-European}   &   hy   &   Armenian   &   712,835   &   3,677,780   &   97     &   Kartvelian   &   ka   &   Georgian   &   703,515   &   4,182,651   &   97 \\
\cline{7-12}
   &   lt   &   Lithuanian   &   718,382   &   3,946,735   &   96                                  &   Koreanic   &   ko   &   Korean   &   711,406   &   4,234,653   &   97 \\
\cline{7-12}
   &   lv   &   Latvian   &   700,889   &   4,011,628   &   97                                     &   \multirow{2}{*}{Kra–Dai}   &   lo   &   Lao   &   357,758   &   2,642,799   &   97 \\
   &   be   &   Belarusian   &   708,288   &   4,169,719   &   95                                  &      &   th   &   Thai   &   707,719   &   4,437,476   &   97 \\
\cline{7-12}
   &   bg   &   Bulgarian   &   711,500   &   4,131,053   &   97                                   &   Mongolic   &   mn   &   Mongolian   &   701,304   &   3,894,353   &   97 \\
\cline{7-12}
   &   bs   &   Bosnian   &   300,000   &   2,953,912   &   97                                     &   \multirow{14}{*}{Niger–Congo}   &   wo   &   Wolof   &   871   &   802,521   &   97 \\
   &   cs   &   Czech   &   711,179   &   4,135,944   &   97                                       &      &   ln   &   Lingala   &   3,325   &   159,684   &   96 \\
   &   hr   &   Croatian   &   300,000   &   4,106,335   &   97                                    &      &   ns   &   Northern Sotho   &   0   &   96,288   &   88 \\
   &   mk   &   Macedonian   &   702,035   &   4,009,787   &   97                                  &      &   lg   &   Luganda   &   13,030   &   216,135   &   95 \\
   &   pl   &   Polish   &   792,829   &   4,200,001   &   98                                      &      &   ny   &   Nyanja   &   226,940   &   3,104,349   &   92 \\
   &   ru   &   Russian   &   853,407   &   4,204,365   &   97                                     &      &   sn   &   Shona   &   386,588   &   3,140,063   &   97 \\
   &   sk   &   Slovak   &   715,540   &   4,100,272   &   98                                      &      &   sw   &   Swahili   &   700,422   &   3,775,394   &   97 \\
   &   sl   &   Slovenian   &   731,613   &   4,073,213   &   97                                   &      &   umb   &   Umbundu   &   0   &   54   &   2 \\
   &   sr   &   Serbian   &   711,535   &   4,033,130   &   97                                     &      &   xh   &   Xhosa   &   122,720   &   3,955,426   &   97 \\
   &   uk   &   Ukrainian   &   714,181   &   4,070,250   &   97                                   &      &   yo   &   Yoruba   &   98,281   &   3,364,040   &   96 \\
   &   cy   &   Welsh   &   703,507   &   3,777,953   &   97                                       &      &   zu   &   Zulu   &   470,403   &   2,899,738   &   97 \\
   &   ga   &   Irish   &   693,460   &   2,814,912   &   96                                       &      &   ig   &   Igbo   &   147,319   &   3,314,731   &   96 \\
   &   is   &   Icelandic   &   704,159   &   4,088,886   &   97                                   &      &   kam   &   Kamba   &   0   &   8   &   1 \\
   &   sv   &   Swedish   &   726,893   &   4,213,939   &   97                                     &      &   ff   &   Fulani   &   26   &   313,870   &   97 \\
\cline{7-12}
   &   da   &   Danish   &   721,543   &   4,194,587   &   97                                      &   Nilo-Saharan   &   luo   &   Dholuo   &   0   &   91   &   6 \\
\cline{7-12}
   &   no   &   Norwegian   &   721,715   &   4,045,571   &   97                                   &   Portuguese   &   kea   &   Kabuverdianu   &   0   &   0   &   0 \\
\cline{7-12}
   &   af   &   Afrikaans   &   703,546   &   4,143,358   &   98                                   &   \multirow{3}{*}{Sino-Tibetan}   &   zh   &   Chinese   &   726,112   &   14,215,583   &   96 \\
   &   de   &   German   &   881,553   &   10,273,597   &   97                                     &     &   zhtrad   &   Chinese   &   0   &   3,747,297   &   96 \\
   &   en   &   English   &   846,712   &   19,548,583   &   100                                   &     &   my   &   Burmese   &   579,160   &   3,887,841   &   97 \\
\cline{7-12}
   &   lb   &   Luxembourgish   &   574,166   &   1,035,619   &   94                               &   \multirow{5}{*}{Turkic}   &   uz   &   Uzbek   &   723,096   &   2,344,375   &   95 \\
   &   nl   &   Dutch   &   769,778   &   4,199,773   &   96                                       &      &   kk   &   Kazakh   &   701,849   &   3,836,259   &   97 \\
   &   el   &   Greek   &   707,751   &   4,081,607   &   97                                       &      &   ky   &   Kyrgyz   &   704,438   &   3,725,583   &   97 \\
   &   bn   &   Bengali   &   707,099   &   4,560,978   &   97                                     &      &   az   &   Azerbaijani   &   712,947   &   8,080,151   &   97 \\
   &   as   &   Assamese   &   33,825   &   1,656,861   &   97                                     &      &   tr   &   Turkish   &   727,711   &   4,169,259   &   97 \\
\cline{7-12}
   &   gu   &   Gujarati   &   704,619   &   3,761,401   &   97                                    &   \multirow{3}{*}{Uralic}   &   et   &   Estonian   &   706,720   &   4,056,200   &   97 \\
   &   hi   &   Hindi   &   715,691   &   4,186,127   &   97                                       &      &   fi   &   Finnish   &   719,416   &   40,76,885   &   97 \\
   &   mr   &   Marathi   &   702,382   &   4,295,708   &   97                                     &      &   hu   &   Hungarian   &   731,479   &   4,154,132   &   97 \\
\bottomrule
    \end{tabular}}
    \caption{The detailed information of the collected monolingual and parallel datasets includes the translation directions for each supported language. Specifically, the ``\# Para.'' represents the count of language-centric sentence pairs, while ``\# Mono.'' denotes the number of individual monolingual sentences.}
    \label{tab:supported_lgs}
\end{table*}

\begin{table*}
    \centering
    \resizebox{\linewidth}{!}{
    \begin{tabular}{crr||crr||crr||crr}
    \toprule
        \textbf{ISO} & \textbf{\# Para.} & \textbf{\# Mono.} & \textbf{ISO} & \textbf{\# Para.} & \textbf{\# Mono.} & \textbf{ISO} & \textbf{\# Para.} & \textbf{\# Mono.} & \textbf{ISO} & \textbf{\# Para.} & \textbf{\# Mono.} \\ 
    \midrule
        af & 201,367,199 & 360,215,552 & hi & 593,592,809 & 366,433,792 & mn & 332,967,182 & 359,067,648 & tg & 347,063,556 & 358,521,344 \\ 
        am & 903,470,803 & 137,815,552 & hr & 212,708,153 & 153,600,000 & mr & 609,131,394 & 359,619,584 & th & 597,728,017 & 362,352,128 \\ 
        ar & 1,054,714,212 & 366,624,256 & hu & 232,631,392 & 374,517,248 & ms & 234,113,298 & 364,490,240 & tl & 244,687,985 & 0 \\ 
        as & 313,146,729 & 17,318,400 & hy & 579,250,350 & 364,971,520 & mt & 102,804,684 & 343,918,592 & tr & 272,252,613 & 372,588,032 \\ 
        ast & 70,949,987 & 0 & id & 232,953,602 & 362,476,544 & my & 1,002,285,410 & 296,529,920 & uk & 218,572,425 & 365,660,672 \\ 
        az & 654,492,231 & 365,028,864 & ig & 242,306,836 & 75,427,328 & ne & 1,237,255,500 & 359,595,008 & umb & 3,170 & 0 \\ 
        be & 306,891,318 & 362,643,456 & is & 251,875,378 & 360,529,408 & nl & 193,189,558 & 394,126,336 & ur & 557,337,279 & 364,213,248 \\ 
        bg & 229,686,547 & 364,288,000 & it & 195,146,279 & 433,206,784 & no & 190,141,837 & 369,518,080 & uz & 148,867,813 & 370,225,152 \\ 
        bn & 755,297,957 & 362,034,688 & ja & 292,857,869 & 371,944,960 & ns & 6,056,515 & 0 & vi & 372,555,263 & 389,361,664 \\ 
        bs & 155,671,162 & 153,600,000 & jv & 150,347,166 & 258,876,928 & ny & 194,642,682 & 116,193,280 & wo & 45,422,689 & 445,952 \\ 
        ca & 196,058,689 & 370,993,664 & ka & 627,397,650 & 360,199,680 & oc & 91,504,042 & 18,626,048 & xh & 242,467,614 & 62,832,640 \\ 
        ceb & 135,958,864 & 214,045,696 & kam & 477 & 0 & om & 13,239,275 & 9,674,240 & yo & 282,242,956 & 50,319,872 \\ 
        cs & 218,791,438 & 364,123,648 & kea & 0 & 0 & or & 289,074,437 & 51,471,360 & zh & 878,523,029 & 371,769,344 \\ 
        cy & 247,455,922 & 360,195,584 & kk & 299,995,454 & 359,346,688 & pa & 1,088,815,314 & 263,161,344 & zhtrad & 252,942,548 & 0 \\ 
        da & 201,340,176 & 369,430,016 & km & 1,266,785,006 & 352,097,280 & pl & 223,440,053 & 405,928,448 & zu & 189,932,839 & 240,846,336 \\ 
        de & 456,147,707 & 451,355,136 & kn & 1,198,503,370 & 364,969,984 & ps & 482,900,652 & 301,230,080 & ~ & ~ & ~ \\ 
        el & 629,383,799 & 362,368,512 & ko & 415,419,459 & 364,239,872 & pt & 189,507,878 & 407,458,816 & ~ & ~ & ~ \\ 
        en & 523,902,024 & 433,516,544 & ku & 494,346,376 & 264,826,368 & ro & 224,472,825 & 359,425,024 & ~ & ~ & ~ \\ 
        es & 193,760,704 & 361,629,184 & ky & 284,261,983 & 360,672,256 & ru & 213,581,742 & 436,944,384 & ~ & ~ & ~ \\ 
        et & 223,902,240 & 361,840,640 & lb & 58,408,912 & 293,972,992 & sd & 107,023,879 & 241,775,104 & ~ & ~ & ~ \\ 
        fa & 505,307,774 & 369,309,184 & lg & 12,860,033 & 6,671,360 & sk & 232,485,422 & 366,356,480 & ~ & ~ & ~ \\ 
        ff & 16,917,593 & 13,312 & ln & 8,942,304 & 1,702,400 & sl & 211,807,076 & 374,585,856 & ~ & ~ & ~ \\ 
        fi & 242,982,346 & 368,340,992 & lo & 932,379,487 & 183,172,096 & sn & 196,567,944 & 197,933,056 & ~ & ~ & ~ \\ 
        fr & 198,627,139 & 403,105,792 & lt & 231,673,345 & 367,811,584 & so & 255,665,827 & 357,306,368 & ~ & ~ & ~ \\ 
        ga & 190,006,560 & 355,051,520 & luo & 4,996 & 0 & sr & 217,789,020 & 364,305,920 & ~ & ~ & ~ \\ 
        gl & 145,312,272 & 371,974,144 & lv & 261,558,146 & 358,855,168 & sv & 190,891,437 & 372,169,216 & ~ & ~ & ~ \\ 
        gu & 1,157,006,948 & 360,764,928 & mi & 234,795,047 & 92,507,136 & sw & 218,972,852 & 358,616,064 & ~ & ~ & ~ \\ 
        ha & 185,399,766 & 215,533,568 & mk & 230,161,774 & 359,441,920 & ta & 805,830,274 & 364,473,856 & ~ & ~ & ~ \\ 
        he & 401,537,464 & 153,600,000 & ml & 773,141,254 & 366,278,144 & te & 1,387,859,988 & 362,731,008 & ~ & ~ & ~ \\ 
    \bottomrule
    \end{tabular}}
        \caption{The detailed information about the tokens used in the continual pre-training stage. The ``\# Para.'' shows the total tokens in the parallel dataset, and ``\# Mono.'' represents the total tokens in the monolingual dataset.}
    \label{tab:supported_lgs_token}
\end{table*}

In this section, we will introduce the sources of our training data~(Section~\ref{sec:training_set}), the evaluation benchmarks~(Section~\ref{sec:evaluation_benchmark}). For translation tasks, we apply beam search to each model with beam size=4.

\subsection{Training Dataset}
\label{sec:training_set}

The dataset was compiled from three distinct open-source datasets, with details on supported languages presented in Table~\ref{tab:supported_lgs} and continual pre-training data statistics in Table~\ref{tab:supported_lgs} and Table~\ref{tab:supported_lgs_token}.

\paragraph{MC4~\citep{xue-etal-2021-mt5}}  is a multilingual variant of the C4 dataset, comprising natural text in 101 languages sourced from the Common Crawl web scrape. It was introduced to support the training of massively multilingual pre-trained text-to-text transformers like mT5.

\paragraph{MADLAD-400~\citep{kudugunta2024madlad}} is a manually audited, general domain monolingual dataset based on CommonCrawl, encompassing 419 languages and designed for document-level analysis. It is notable for its extensive language coverage and the rigorous auditing process involved in its creation.

\paragraph{Lego-MT~\cite{yuan-etal-2023-lego}} is a benchmark for massively multilingual machine translation, featuring a detachable model built upon an efficient training recipe. It includes a comprehensive translation benchmark with data from OPUS, covering 433 languages and 1.3 billion parallel data points.

\subsection{Evaluation Benchmark}
\label{sec:evaluation_benchmark}

\paragraph{Flores-101~\citep{flores101}} is a benchmark for machine translation evaluation, comprising a multi-way dataset derived from English Wikipedia and produced by professional translators.

\paragraph{Flores-200~\citep{nllb2022}} is an extension of Flores-101 dataset and also serves as a benchmark for machine translation. This dataset contains parallel sentences for 200 languages, with each language identified by its ISO 639-3 code~( (e.g. eng)) and an additional code~(e.g., "eng\_Latn",) that describes the script.

\paragraph{WMT-23~\citep{wmt23}} is also a comprehensive translation evaluation benchmark, proposed in 2023. We incorporate this dataset into our evaluation to mitigate the risk of data leakage in LLMs.  Based on benchmark, we evaluate the English-centric translation task performance, including de$\rightarrow$en, en$\rightarrow$cs, en$\rightarrow$de, en$\rightarrow$he, en$\rightarrow$ja, en$\rightarrow$ru, en$\rightarrow$uk, en$\rightarrow$zh, he$\rightarrow$en, ja$\rightarrow$en, ru$\rightarrow$en, uk$\rightarrow$en, zh$\rightarrow$en.

\paragraph{TICO~\cite{tico}} dataset represents a joint translation effort targeting COVID-19 materials, developed in collaboration with academic, industry stakeholders, and Translators without Borders. It comprises translation memories, a glossary of translated COVID-19 terms, and functions as a benchmark for translation-related evaluations. The all evaluated translation is en$\rightarrow$\{am, bn, din, fa, fuv, hi, km, ku, ln, ms, ne, om, ps, ru, so, ta, ti\_ER, tl, zh, ar, ckb, es\_LA, fr, ha, id, kr, lg, mr, my, nus, prs, pt\_BR, rw, sw, ti, ti\_ET, ur, zu\}.

\paragraph{TED~\cite{ted}} is a massively multilingual dataset derived from TED Talk transcripts, covering 60 languages with parallel arrays of language and text. It is designed for natural language processing tasks and filters out missing or incomplete translations. We also evaluate the English-centric translation performance. The translation direction covers all 60 languages, including en$\leftrightarrow$\{af, am, ar, arq, art-x-bork, as, ast, az, be, bg, bi, bn, bo, bs, ca, ceb, cnh, cs, da, de, el, eo, es, et, eu, fa, fi, fil, fr, fr-ca, ga, gl, gu, ha, he, hi, hr, ht, hu, hup, hy, id, ig, inh, is, it, ja, ka, kk, km, kn, ko, ku, ky, la, lb, lo, lt, ltg, lv, mg, mk, ml, mn, mr, ms, mt, my, nb, ne, nl, nn, oc, pa, pl, ps, pt, pt-br, ro, ru, rup, sh, si, sk, sl, so, sq, sr, srp, sv, sw, szl, ta, te, tg, th, tl, tlh, tr, tt, ug, uk, ur, uz, vi, zh, zh-cn, zh-tw\}

\paragraph{X-CSQA~\cite{xcsqa}} is a multilingual extension of the Commonsense Question Answering (CSQA) dataset, designed for commonsense reasoning research. It facilitates the evaluation and improvement of multilingual language models in commonsense reasoning tasks.

\begin{figure*}[!t]
    \centering
   \includegraphics[width=1\textwidth]{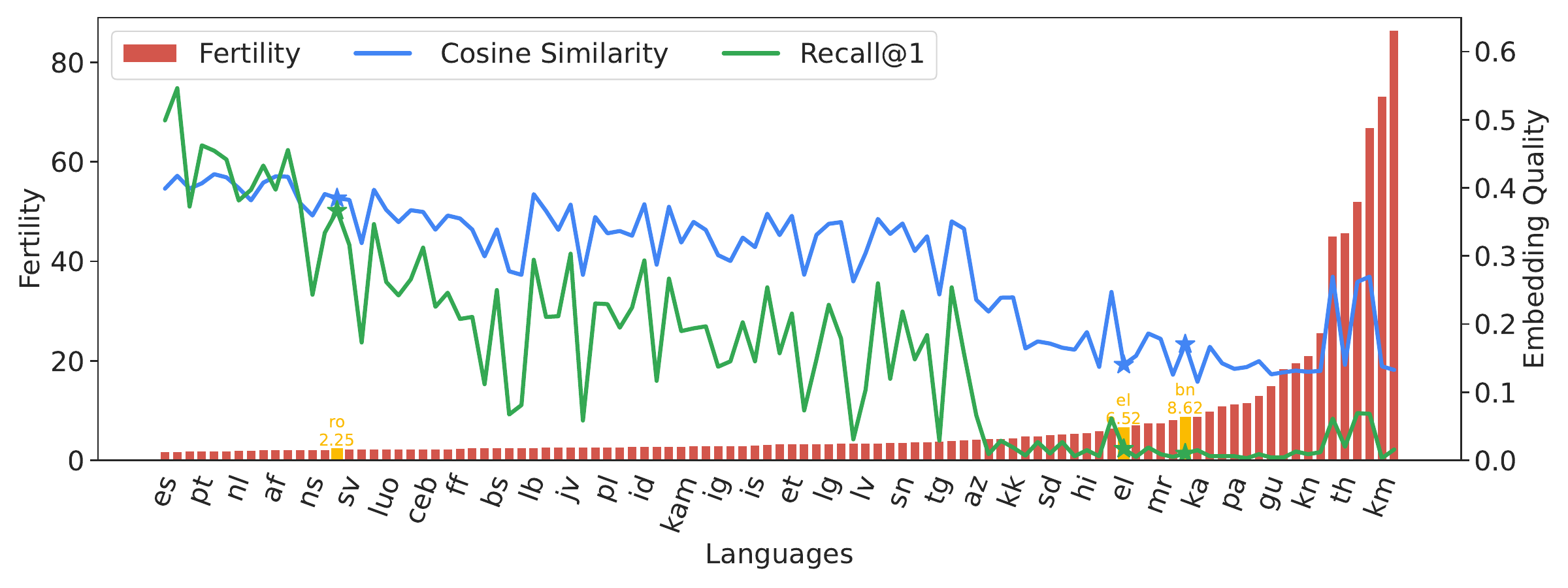}
    \caption{Correlation between embedding quality and fertility. The embedding quality of LLaMA2 is measured by cosine similarity and Recall@1 on Flores-101 test. Fertility refers to the ratio of the length of a sentence after tokenization compared to its length before tokenization. A high fertility may result in a poor quality of embedding.}
    \label{fig:correlation_fertility_with_quality}
\end{figure*}

\paragraph{XStoryCloze~\citep{XStoryCloze}} is a benchmark dataset that comprises the professionally translated English StoryCloze dataset~(Spring 2016 version) into 10 non-English languages. It is designed to evaluate the zero- and few-shot learning capabilities of multilingual language models.

\paragraph{XCOPA~\citep{XCOPA}} is a benchmark dataset that assesses machine learning models’ ability to transfer commonsense reasoning across languages. It is an extension of the English COPA dataset and includes 11 languages from diverse language families and geographical regions.

\paragraph{XWinograd~\citep{muennighoff2022crosslingual, tikhonov2021heads}} s a benchmark dataset that consists of a multilingual collection of Winograd Schemas, designed for the evaluation of cross-lingual commonsense reasoning capabilities covering six languages.

\paragraph{XNLI~\citep{xnli}} is a cross-lingual extension of the SNLI~\citep{snli}/MultiNLI~\citep{MultiNLI}, consisting of a subset of English examples translated into 14 different languages. It is used for evaluating textual entailment and classification tasks, where the goal is to determine if one sentence implies, contradicts, or is neutral to another sentence

\paragraph{MGSM~\citep{mgsm}} a dataset of grade-school math problems, each translated into 10 languages by human annotators. It is derived from the GSM8K~\citep{gsm8k} dataset and is designed to support question answering on basic mathematical problems that require multi-step reasoning.

\paragraph{MMLU~\citep{hendrycks2021ethics,hendryckstest2021}} is a benchmark for evaluating language models’ capabilities in language comprehension and reasoning across diverse domains. It consists of about 16,000 multiple-choice questions spanning 57 academic subjects, designed to measure knowledge acquired during pretraining in zero-shot and few-shot settings.

\paragraph{BBH~\citep{BBH}} is a subset of the BIG-Bench, focusing on 23 challenging tasks that current language models struggle to perform, where they do not outperform the average human-rater. It serves as a rigorous evaluation suite to test the limits of language models’ capabilities.

\paragraph{HellaSwag~\citep{zellers2019hellaswag}} s a dataset designed to evaluate advanced natural language understanding and common sense reasoning, which introduces more complexity and diversity, challenging AI models to predict the ending of incomplete narratives.

\paragraph{WinoG~\citep{winogrande}} is a large-scale dataset containing 44k problems inspired by the Winograd Schema Challenge, designed to improve the scale and hardness of coreference resolution tasks. It presents fill-in-the-blank questions with binary options, testing the model’s ability to understand nuanced human language.

\paragraph{NQ~\citep{NQ}}  is a dataset for question answering research, containing over 300,000 examples each consisting of a real user query and a corresponding Wikipedia page. It is designed to train and evaluate automatic question answering systems by emulating how people search for information.

\paragraph{HumanEval~\citep{humaneval}}  is designed to evaluate the code generation capabilities of large language models, featuring 164 hand-crafted programming challenges that include function signatures, docstrings, bodies, and unit tests. On average, each problem is accompanied by 7.7 tests to assess functional correctness.

\paragraph{MBPP~\citep{mbpp}} comprises approximately 1,000 crowd-sourced Python programming problems, aimed at entry-level programmers and covering programming fundamentals and standard library functionality. Each problem includes a task description, code solution, and three automated test cases.

\paragraph{GSM8K~\citep{gsm8k}} consists of 8.5K high-quality, linguistically diverse grade school math word problems created by human problem writers. It is designed to support question answering on basic mathematical problems that require multi-step reasoning.

\begin{table*}[!htb]
    \centering
    \footnotesize
    \begin{tabular}{l|c|cc|cc|cc|cc}
    \toprule
         \multirow{2}{*}{\textbf{Setting}} & \multirow{2}{*}{\textbf{Aug}} & \multicolumn{2}{c|}{\textbf{en-centric}} & \multicolumn{2}{c|}{\textbf{ta-centric}} & \multicolumn{2}{c|}{\textbf{th-centric}} & \multicolumn{2}{c}{\textbf{zh-centric}} \\
         & & \textbf{en$\rightarrow$X} & \textbf{X$\rightarrow$en} & \textbf{ta$\rightarrow$X} & \textbf{X$\rightarrow$ta} & \textbf{th$\rightarrow$X} & \textbf{X$\rightarrow$th} & \textbf{zh$\rightarrow$X} & \textbf{X$\rightarrow$zh} \\
    \midrule
    LLaMA2 & \xmark &  18.31 & 23.61 &  0.99 & 0.49 & 4.83 & 1.15 & 10.02 & 7.35 \\
    \midrule
    $\mathcal{D}_\mathrm{P_1}$	& \xmark & 19.06 & 25.98 & 3.20 & 0.91 & 7.66 & 3.13 & 11.32 & 7.83 \\
$\mathcal{D}_\mathrm{P_1}$+$\mathcal{D}_\mathrm{P_2}$	& \xmark & 19.46 & 26.40 & 4.17 & 1.76 & 7.28 & 3.02 & 11.65 & 8.82 \\
$\mathcal{D}_\mathrm{P_1}$+$\mathcal{D}_\mathrm{M}$	& \xmark & 19.22 & 25.91  & 3.51 & 1.34 & 7.64 & 2.83 & 11.56 & 7.99 \\
$\mathcal{D}_\mathrm{P_1}$+$\mathcal{D}_\mathrm{P_2}$+$\mathcal{D}_\mathrm{M}$	& \xmark & 19.36 & 26.47 & 4.35 & 1.82 & 7.78 & 3.49 & 11.44 & 9.14 \\
\midrule
$\mathcal{D}_\mathrm{P_1}$+$\mathcal{D}_\mathrm{P_2}'$ &	\cmark & 19.47 & 26.65 & 4.54 & 1.83 & 7.66 & 3.13 & 11.89 & 9.17 \\
$\mathcal{D}_\mathrm{P_1}$+$\mathcal{D}_\mathrm{M}'$ & 	\cmark & 18.59 & 25.98 & 3.61 & 1.36 & 6.72 & 2.35 & 10.81 & 6.45 \\
$\mathcal{D}_\mathrm{P_1}$+$\mathcal{D}_\mathrm{P_2}'$+$\mathcal{D}_\mathrm{M}$ & \cmark & \textbf{19.70} & \textbf{26.71} & 4.68 & 1.82 & \textbf{8.21} & \textbf{3.65} & \textbf{12.05} & \textbf{9.28} \\
$\mathcal{D}_\mathrm{P_1}$+$\mathcal{D}_\mathrm{P_2}$+$\mathcal{D}_\mathrm{M}'$ & \cmark & 19.17 & 26.58 & 4.57 & \textbf{1.95} & 7.12 & 3.12 & 11.52 & 7.73\\
$\mathcal{D}_\mathrm{P_1}$+$\mathcal{D}_\mathrm{P_2}'$+$\mathcal{D}_\mathrm{M}'$	& \cmark & 18.80 & 26.56 & \textbf{4.78} & 1.79 & 7.31 & 3.18 & 11.35 & 7.28 \\
\midrule
\midrule
\multirow{2}{*}{\textbf{Setting}} & \multirow{2}{*}{\textbf{Dictionary}} & \multicolumn{2}{|c}{\textbf{en-centric}} & \multicolumn{2}{|c}{\textbf{ta-centric}} & \multicolumn{2}{|c}{\textbf{th-centric}} & \multicolumn{2}{|c}{\textbf{zh-centric}} \\
         &  & \textbf{en$\rightarrow$x} & \textbf{x$\rightarrow$en} & \multicolumn{1}{|c}{\textbf{ta$\rightarrow$x}} & \textbf{x$\rightarrow$ta} & \multicolumn{1}{|c}{\textbf{th$\rightarrow$x}} & \textbf{x$\rightarrow$th} & \multicolumn{1}{|c}{\textbf{zh$\rightarrow$x}} & \textbf{x$\rightarrow$zh} \\
    \midrule
    $\mathcal{D}_\mathrm{P_1}$+$\mathcal{D}_\mathrm{P_2}'$+$\mathcal{D}_\mathrm{M}'$ &  MUSE: 1-hop &  \multicolumn{1}{|c}{18.80} & 26.56 & \multicolumn{1}{|c}{4.78} & 1.79 & \multicolumn{1}{|c}{7.31} & 3.18 & \multicolumn{1}{|c}{11.35} & 7.28 \\
    $\mathcal{D}_\mathrm{P_1}$+$\mathcal{D}_\mathrm{P_2}'$+$\mathcal{D}_\mathrm{M}'$ &  MUSE: 2-hop & \multicolumn{1}{|c}{18.70} & 26.50 & \multicolumn{1}{|c}{4.47} & 1.83 & \multicolumn{1}{|c}{7.08} & 3.26 & \multicolumn{1}{|c}{10.74} & 6.68  \\
    $\mathcal{D}_\mathrm{P_1}$+$\mathcal{D}_\mathrm{P_2}'$+$\mathcal{D}_\mathrm{M}'$ &  PanLex: 1-hop & \multicolumn{1}{|c}{19.33} & 26.54 & \multicolumn{1}{|c}{4.40} & 1.83 & \multicolumn{1}{|c}{7.57} & 3.31 & \multicolumn{1}{|c}{10.86} & 8.08 \\
    \bottomrule
    \end{tabular}
    \caption{A comprehensive analysis of data augmentation sources reveals that using a dictionary to augment parallel data alone improves translation performance. Each cell in the table refers to the average spBLEU score. ``Aug'' is a boolean representing whether a dictionary is used for augmentation. Meanwhile, we select a specific data augmentation technique and evaluate various dictionary configurations, including 1-hop and 2-hop, as well as different dictionaries.}
    \label{tab:select_hop}
\end{table*}

\begin{table}[!ht]
    \centering
     \footnotesize
    \resizebox{\linewidth}{!}{
    \begin{tabular}{cc|cc}
        \toprule 
        \multicolumn{2}{c|}{\textbf{1-hop translation}} & \multicolumn{2}{c}{\textbf{2-hop translation}} \\
         \textbf{Direction} & \textbf{Example} & \textbf{Direction} & \textbf{Example} \\
         \midrule
         en$\rightarrow$fr & dog $\rightarrow$ chien & \multirow{2}{*}{en$\rightarrow$fr$\rightarrow$de} &\multirow{2}{*}{dog $\rightarrow$ chien $\rightarrow$ Hund} \\
         fr$\rightarrow$de & chien $\rightarrow$ Hund &  \\
         \bottomrule
    \end{tabular}}
    \caption{Case of 1-hop and 2-hop translations.}
    \label{tab:hop_example}
\end{table}

\paragraph{Math~\citep{hendrycksmath2021}} is a collection of 12,500 intricate problems derived from competition mathematics. Every problem within the Math dataset includes a comprehensive solution with step-by-step guidance, which serves as a resource for training models to produce detailed answer justifications and explanations.

\section{Model Information}
\label{sec:model_info}

\begin{table*}[!htb]
    \centering
    \footnotesize
    \resizebox{\linewidth}{!}{
    \begin{tabular}{r|cccccccc}
    \toprule
         \multirow{2}{*}{\textbf{Setting}} &  \multicolumn{2}{c|}{\textbf{Translation Tasks}} & \multicolumn{3}{c|}{\textbf{General Tasks}} & \multicolumn{3}{c}{\textbf{Multilingual Tasks}}  \\
         & \textbf{ceb$\rightarrow$x} & \multicolumn{1}{c|}{\textbf{x$\rightarrow$ceb}} & \textbf{QNLI} & \textbf{QQP} & \multicolumn{1}{c|}{\textbf{MRPC}} & \textbf{XStoryCloze} & \multicolumn{1}{c}{\textbf{XCOPA}} &  \multicolumn{1}{c}{\textbf{XWinograd}} \\
    \midrule
         splited-parallel + mono & 3.36 & \multicolumn{1}{c|}{2.74} & 49.46 & 36.82 & \multicolumn{1}{c|}{68.38} & 59.20 & 56.82 & 73.72 \\
         connected-parallel + mono & 4.45 & \multicolumn{1}{c|}{3.68} & 49.46 & 36.82 & \multicolumn{1}{c|}{68.38}  & 59.10 & \multicolumn{1}{c|}{56.80} & 74.07 \\
    \midrule
    \midrule
    \textbf{Setting} &	\textbf{ceb$\rightarrow$ca}	 &	\textbf{ceb$\rightarrow$de}	 &	\textbf{ceb$\rightarrow$en}	 &	\textbf{ceb$\rightarrow$es} & \textbf{ceb$\rightarrow$fr}	 &	\textbf{ceb$\rightarrow$it} &	\textbf{ceb$\rightarrow$pt}	 &	\textbf{ceb$\rightarrow$ru} \\
    \midrule
	splited-parallel + mono & 10.32 &	8.94	 &	23.19	 &	13.30	 &	15.96	 &	10.01	 &	12.66	 &	8.05 \\
	connected-parallel + mono & 10.97	 &	11.37	 &	27.06	 &	14.91	 &	18.04	 &	12.03	 &	15.55	 &	10.26 \\
	\midrule				
    \textbf{Setting} &  \textbf{ca$\rightarrow$ceb}	 &	\textbf{de$\rightarrow$ceb}	 &	\textbf{en$\rightarrow$ceb}	 &	\textbf{es$\rightarrow$ceb}	 &	\textbf{fr$\rightarrow$ceb}	 &	\textbf{it$\rightarrow$ceb}	 &	\textbf{pt$\rightarrow$ceb}	 &	\textbf{ru$\rightarrow$ceb} \\
    \midrule
	splited-parallel + mono & 5.90	 &	4.91	 &	7.44	 &	5.14	 &	6.02	 &	5.54	 &	6.12	 &	4.24 \\
	connected-parallel + mono & 7.62	 &	6.92	 &	9.88	 &	6.41	 &	7.39	 &	6.91	 &	7.62	 &	6.54 \\
    \bottomrule
    \end{tabular}}
    \caption{Design for the utilization of parallel data, we take ceb-centric data as an example, apply two distict approaches, and discover that treating parallel data as two independent monolingual datasets harms to translation performance.}
    \label{tab:usage_para}
\end{table*}

Model details about the baseline models for comparison, including decode-only large language models~(LLMs) in Section~\ref{sec:llms} as well as translation models in Section~\ref{sec:translation_models} with an encoder-decoder structure. 

\subsection{Large Language Models}
\label{sec:llms}

\paragraph{LLaMA2~\citep{LLaMA2}} is a decoder-only language model that predicts the next token based on the input sequence of ordered tokens, with a collection of pre-trained and fine-tuned models ranging from 7 billion to 70 billion parameters. The LLaMA2 7B model serves as our foundational model. Unless otherwise specified, any reference to LLaMA or LLaMA2 is the LLaMA2 7B model. The model leverages a Byte-level Byte Pair Encoding~(BBPE;~\citealp{bbpe}) tokenizer, an efficient subword tokenizer that tokenizes at the byte level, allowing it to handle any language and be robust to noise in the data. The BBPE tokenizer is particularly useful for languages with large vocabularies and many rare words. 

\paragraph{\name} follows the model architecture of LLaMA2 without vocabulary extension. We utilize 24 A100 80GB GPUs and extended the pre-training on the amassed data for over 60 days. We set per device training batch size to 32, the learning rate to 2e-5, and the epoch number to 1.0. 

\paragraph{PolyLM~\citep{wei2023polylm}} is an open-source multilingual Large Language Model (LLM) trained on 640 billion tokens, available in two model sizes: 1.7B and 13B. It boasts proficiency in 15 major non-English languages, employing advanced training techniques to enhance its language processing capabilities. 

\paragraph{Yayi2~\citep{luo2023yayi}} is a multilingual open-source Large Language Model pre-trained from scratch on a corpus containing 2.65 trillion tokens. It is aligned with human values through supervised fine-tuning and reinforce
ment learning from human feedback.

\paragraph{TowerInstruct~\citep{alves2024tower}} is a 7B parameter language model fine-tuned on translation-related tasks, supporting multiple languages including English, Portuguese, Spanish, French, and others. It is designed for tasks such as machine translation, automatic post-editing, and paraphrase generation. In our paper, we evaluate the instruction-tuned model TowerInstruct-7B-v0.2.

\paragraph{Aya-23~\citep{aryabumi2024aya}}  is an open weights research release of an instruction fine-tuned decoder-only model with advanced multilingual capabilities, serving 23 languages. It pairs a performant pre-trained Command family of models with the Aya Collection for robust language processing tasks.

\paragraph{ChineseLLaMA2-Alpaca~\cite{cui2024efficient}} is founded on LLaMA2 and enhanced with an extensive Chinese vocabulary that concentrates on Chinese languages. This is a fine-tuned version of ChineseLLaMA2 using Alpaca~\citep{alpaca} data.

\paragraph{LLaMA2-SFT~\citep{alpaca}} is a fine-tuned version of LLaMA2 model, leveraging a set of 52,000 diverse English instructions in Alpaca~\citep{alpaca} to enhance the instruction-following capabilities of the model.

\begin{figure}[!t]
    \centering
    \includegraphics[width=1\linewidth]{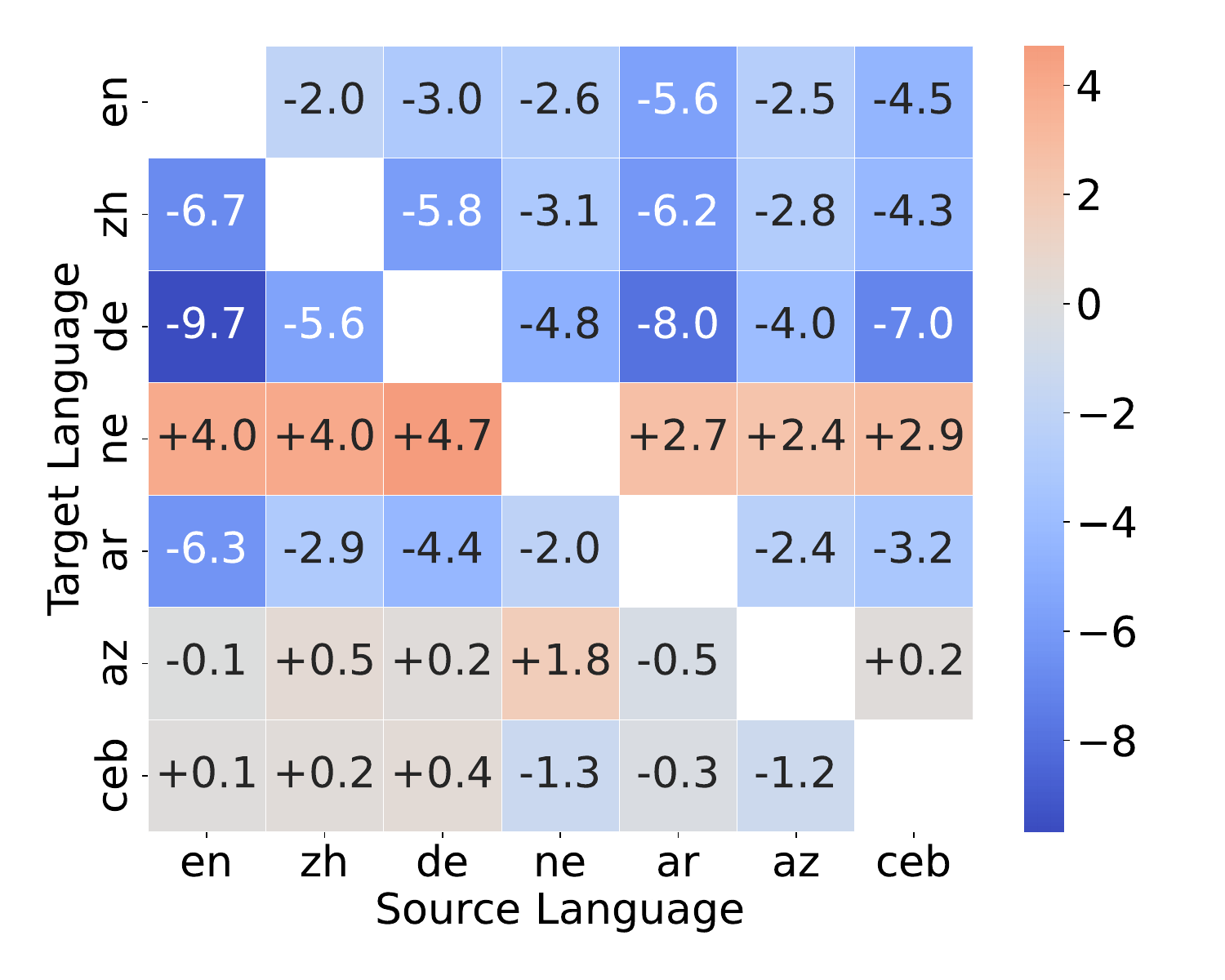}
    \caption{The spBLEU gap between \name and GPT-4. Positive scores mean the result of \name is better than GPT-4. Empirical evidence demonstrates that while \name trails GPT-4 in high-resource translation scenarios, it outperforms in low-resource translation contexts.}
    \label{fig:comparison_gpt-4}
\end{figure}

\paragraph{Qwen2-7B-Instruct~\citep{qwen2}} is part of the Qwen2 series, which is a instruction-tuned language models. It demonstrates competitiveness against proprietary models across multilingual benchmarks.

\paragraph{Swallow~\citep{fujii2024continual}} is a large language model which enhances Japanese capability based on LLaMA2. It achieves this by extending the vocabulary with Japanese characters and conducting continued pre-training on a Japanese corpus, resulting in superior performance compared to other LLMs in both English and Japanese tasks. In our paper, we evaluate the instruction-tuned model Swallow-7B-Instruct-v0.1.

\subsection{Translation Models}
\label{sec:translation_models}

\begin{table*}[!ht]
    \centering
    \footnotesize
    \resizebox{\linewidth}{!}{
    \begin{tabular}{c|cc|cc|cc||c|cc|cc|cc}
\toprule
 \multirow{2}{*}{\textbf{X}} & \multicolumn{2}{c|}{\textbf{LLaMA2-Alpaca}} & \multicolumn{2}{c|}{\textbf{ChineseLLaMA2-Alpaca}} & \multicolumn{2}{c||}{\textbf{\name-Alpaca}}  &  \multirow{2}{*}{\textbf{X}}
 & \multicolumn{2}{c|}{\textbf{LLaMA2-Alpaca}} & \multicolumn{2}{c|}{\textbf{ChineseLLaMA2-Alpaca}} & \multicolumn{2}{c}{\textbf{\name-Alpaca}} \\
 & \textbf{$R_\mathrm{zh}$} & \textbf{$R_\mathrm{X}$} & \textbf{$R_\mathrm{zh}$} & \textbf{$R_\mathrm{X}$} & \textbf{$R_\mathrm{zh}$} & \textbf{$R_\mathrm{X}$} & & \textbf{$R_\mathrm{zh}$} & \textbf{$R_\mathrm{X}$}  & \textbf{$R_\mathrm{zh}$} & \textbf{$R_\mathrm{X}$} & \textbf{$R_\mathrm{zh}$} & \textbf{$R_\mathrm{X}$} \\
 \midrule
af	&	\cellcolor{mycolor!0.2}0.20	&	\cellcolor{mycolor!28.36}28.36	&	\cellcolor{mycolor!31.32}31.32	&	\cellcolor{mycolor!0.1}0.10	&	\cellcolor{mycolor!0.3}0.30	&	\cellcolor{mycolor!79.84}79.84	&		ln	&	\cellcolor{mycolor!0.3}0.30	&	\cellcolor{mycolor!0}0.00	&	\cellcolor{mycolor!66.4}66.40	&	\cellcolor{mycolor!0}0.00	&	\cellcolor{mycolor!0}0.00	&	\cellcolor{mycolor!0}0.00 \\

am	&	\cellcolor{mycolor!1.09}1.09	&	\cellcolor{mycolor!40.12}40.12	&	\cellcolor{mycolor!67.29}67.29	&	\cellcolor{mycolor!21.15}21.15	&	\cellcolor{mycolor!0}0.00	&	\cellcolor{mycolor!89.23}89.23	&		lo	&	\cellcolor{mycolor!1.38}1.38	&	\cellcolor{mycolor!32.71}32.71	&	\cellcolor{mycolor!89.03}89.03	&	\cellcolor{mycolor!0.1}0.10	&	\cellcolor{mycolor!0}0.00	&	\cellcolor{mycolor!58.3}58.30 \\

ar	&	\cellcolor{mycolor!2.17}2.17	&	\cellcolor{mycolor!81.23}81.23	&	\cellcolor{mycolor!72.92}72.92	&	\cellcolor{mycolor!24.7}24.70	&	\cellcolor{mycolor!0}0.00	&	\cellcolor{mycolor!99.8}99.80	&		lt	&	\cellcolor{mycolor!1.09}1.09	&	\cellcolor{mycolor!14.13}14.13	&	\cellcolor{mycolor!50.69}50.69	&	\cellcolor{mycolor!24.31}24.31	&	\cellcolor{mycolor!0.2}0.20	&	\cellcolor{mycolor!96.34}96.34 \\

as	&	\cellcolor{mycolor!8.4}8.40	&	\cellcolor{mycolor!0.59}0.59	&	\cellcolor{mycolor!84.39}84.39	&	\cellcolor{mycolor!0.3}0.30	&	\cellcolor{mycolor!0}0.00	&	\cellcolor{mycolor!76.78}76.78	&			luo	&	\cellcolor{mycolor!5.83}5.83	&	\cellcolor{mycolor!0}0.00	&	\cellcolor{mycolor!87.65}87.65	&	\cellcolor{mycolor!0}0.00	&	\cellcolor{mycolor!1.38}1.38	&	\cellcolor{mycolor!0}0.00 \\

ast	&	\cellcolor{mycolor!0.3}0.30	&	\cellcolor{mycolor!0.2}0.20	&	\cellcolor{mycolor!18.77}18.77	&	\cellcolor{mycolor!0.1}0.10	&	\cellcolor{mycolor!0}0.00	&	\cellcolor{mycolor!33.2}33.20	&		lv	&	\cellcolor{mycolor!0.3}0.30	&	\cellcolor{mycolor!15.51}15.51	&	\cellcolor{mycolor!52.67}52.67	&	\cellcolor{mycolor!15.42}15.42	&	\cellcolor{mycolor!0.2}0.20	&	\cellcolor{mycolor!97.73}97.73\\

az	&	\cellcolor{mycolor!0.2}0.20	&	\cellcolor{mycolor!18.87}18.87	&	\cellcolor{mycolor!39.23}39.23	&	\cellcolor{mycolor!4.25}4.25	&	\cellcolor{mycolor!0}0.00	&	\cellcolor{mycolor!96.44}96.44	&				mi	&	\cellcolor{mycolor!0.49}0.49	&	\cellcolor{mycolor!0}0.00	&	\cellcolor{mycolor!59.58}59.58	&	\cellcolor{mycolor!0}0.00	&	\cellcolor{mycolor!0}0.00	&	\cellcolor{mycolor!0}0.00 \\

be	&	\cellcolor{mycolor!0.1}0.10	&	\cellcolor{mycolor!49.11}49.11	&	\cellcolor{mycolor!2.96}2.96	&	\cellcolor{mycolor!2.87}2.87	&	\cellcolor{mycolor!0}0.00	&	\cellcolor{mycolor!99.7}99.70	&			mk	&	\cellcolor{mycolor!0.4}0.40	&	\cellcolor{mycolor!17.19}17.19	&	\cellcolor{mycolor!7.31}7.31	&	\cellcolor{mycolor!21.94}21.94	&	\cellcolor{mycolor!0}0.00	&	\cellcolor{mycolor!99.31}99.31\\

bg	&	\cellcolor{mycolor!2.37}2.37	&	\cellcolor{mycolor!44.66}44.66	&	\cellcolor{mycolor!29.74}29.74	&	\cellcolor{mycolor!30.24}30.24	&	\cellcolor{mycolor!0.3}0.30	&	\cellcolor{mycolor!98.62}98.62	&			ml	&	\cellcolor{mycolor!8.2}8.20	&	\cellcolor{mycolor!12.15}12.15	&	\cellcolor{mycolor!79.55}79.55	&	\cellcolor{mycolor!7.51}7.51	&	\cellcolor{mycolor!0.49}0.49	&	\cellcolor{mycolor!51.88}51.88\\

bn	&	\cellcolor{mycolor!3.95}3.95	&	\cellcolor{mycolor!44.96}44.96	&	\cellcolor{mycolor!78.75}78.75	&	\cellcolor{mycolor!17.79}17.79	&	\cellcolor{mycolor!0.1}0.10	&	\cellcolor{mycolor!99.6}99.60	&		mn	&	\cellcolor{mycolor!1.58}1.58	&	\cellcolor{mycolor!17.49}17.49	&	\cellcolor{mycolor!85.67}85.67	&	\cellcolor{mycolor!1.48}1.48	&	\cellcolor{mycolor!0}0.00	&	\cellcolor{mycolor!99.51}99.51\\

bs	&	\cellcolor{mycolor!0.4}0.40	&	\cellcolor{mycolor!2.17}2.17	&	\cellcolor{mycolor!8.1}8.10	&	\cellcolor{mycolor!1.98}1.98	&	\cellcolor{mycolor!0.1}0.10	&	\cellcolor{mycolor!4.25}4.25	&		mr	&	\cellcolor{mycolor!0.4}0.40	&	\cellcolor{mycolor!19.86}19.86	&	\cellcolor{mycolor!31.42}31.42	&	\cellcolor{mycolor!1.58}1.58	&	\cellcolor{mycolor!0}0.00	&	\cellcolor{mycolor!99.01}99.01\\

ca	&	\cellcolor{mycolor!0.3}0.30	&	\cellcolor{mycolor!90.12}90.12	&	\cellcolor{mycolor!5.14}5.14	&	\cellcolor{mycolor!79.84}79.84	&	\cellcolor{mycolor!0}0.00	&	\cellcolor{mycolor!98.91}98.91	&			ms	&	\cellcolor{mycolor!0.59}0.59	&	\cellcolor{mycolor!5.93}5.93	&	\cellcolor{mycolor!20.36}20.36	&	\cellcolor{mycolor!3.95}3.95	&	\cellcolor{mycolor!0}0.00	&	\cellcolor{mycolor!43.18}43.18\\

ceb	&	\cellcolor{mycolor!0.2}0.20	&	\cellcolor{mycolor!21.94}21.94	&	\cellcolor{mycolor!6.72}6.72	&	\cellcolor{mycolor!16.01}16.01	&	\cellcolor{mycolor!0}0.00	&	\cellcolor{mycolor!95.55}95.55	&		mt	&	\cellcolor{mycolor!0.2}0.20	&	\cellcolor{mycolor!63.44}63.44	&	\cellcolor{mycolor!29.15}29.15	&	\cellcolor{mycolor!25}25.00	&	\cellcolor{mycolor!0}0.00	&	\cellcolor{mycolor!97.13}97.13\\

cs	&	\cellcolor{mycolor!0.2}0.20	&	\cellcolor{mycolor!54.55}54.55	&	\cellcolor{mycolor!24.9}24.90	&	\cellcolor{mycolor!38.14}38.14	&	\cellcolor{mycolor!0.3}0.30	&	\cellcolor{mycolor!94.76}94.76	&		my	&	\cellcolor{mycolor!1.78}1.78	&	\cellcolor{mycolor!47.33}47.33	&	\cellcolor{mycolor!38.74}38.74	&	\cellcolor{mycolor!29.74}29.74	&	\cellcolor{mycolor!0}0.00	&	\cellcolor{mycolor!99.9}99.90\\

cy	&	\cellcolor{mycolor!0.3}0.30	&	\cellcolor{mycolor!19.66}19.66	&	\cellcolor{mycolor!20.55}20.55	&	\cellcolor{mycolor!44.66}44.66	&	\cellcolor{mycolor!0}0.00	&	\cellcolor{mycolor!98.81}98.81	&		ne	&	\cellcolor{mycolor!0.49}0.49	&	\cellcolor{mycolor!35.77}35.77	&	\cellcolor{mycolor!71.64}71.64	&	\cellcolor{mycolor!3.06}3.06	&	\cellcolor{mycolor!0}0.00	&	\cellcolor{mycolor!98.72}98.72\\

da	&	\cellcolor{mycolor!0.3}0.30	&	\cellcolor{mycolor!49.01}49.01	&	\cellcolor{mycolor!22.73}22.73	&	\cellcolor{mycolor!39.72}39.72	&	\cellcolor{mycolor!0.49}0.49	&	\cellcolor{mycolor!91.8}91.80	&		nl	&	\cellcolor{mycolor!0.3}0.30	&	\cellcolor{mycolor!65.81}65.81	&	\cellcolor{mycolor!4.55}4.55	&	\cellcolor{mycolor!65.22}65.22	&	\cellcolor{mycolor!0.1}0.10	&	\cellcolor{mycolor!94.76}94.76\\

de	&	\cellcolor{mycolor!0.79}0.79	&	\cellcolor{mycolor!70.55}70.55	&	\cellcolor{mycolor!10.97}10.97	&	\cellcolor{mycolor!75.69}75.69	&	\cellcolor{mycolor!0.3}0.30	&	\cellcolor{mycolor!96.94}96.94	&		no	&	\cellcolor{mycolor!0.99}0.99	&	\cellcolor{mycolor!32.21}32.21	&	\cellcolor{mycolor!22.53}22.53	&	\cellcolor{mycolor!28.06}28.06	&	\cellcolor{mycolor!0.2}0.20	&	\cellcolor{mycolor!88.74}88.74\\

el	&	\cellcolor{mycolor!0.69}0.69	&	\cellcolor{mycolor!21.25}21.25	&	\cellcolor{mycolor!52.67}52.67	&	\cellcolor{mycolor!28.26}28.26	&	\cellcolor{mycolor!0}0.00	&	\cellcolor{mycolor!100}100.00	&		ns	&	\cellcolor{mycolor!0.2}0.20	&	\cellcolor{mycolor!0}0.00	&	\cellcolor{mycolor!38.74}38.74	&	\cellcolor{mycolor!0}0.00	&	\cellcolor{mycolor!0.1}0.10	&	\cellcolor{mycolor!0}0.00\\

en	&	\cellcolor{mycolor!0}0.00	&	\cellcolor{mycolor!100}100.00	&	\cellcolor{mycolor!0.3}0.30	&	\cellcolor{mycolor!99.7}99.70	&	\cellcolor{mycolor!0}0.00	&	\cellcolor{mycolor!100}100.00	&		ny	&	\cellcolor{mycolor!0.59}0.59	&	\cellcolor{mycolor!0}0.00	&	\cellcolor{mycolor!60.08}60.08	&	\cellcolor{mycolor!0}0.00	&	\cellcolor{mycolor!0.2}0.20	&	\cellcolor{mycolor!0}0.00\\

es	&	\cellcolor{mycolor!0.1}0.10	&	\cellcolor{mycolor!96.94}96.94	&	\cellcolor{mycolor!4.74}4.74	&	\cellcolor{mycolor!93.08}93.08	&	\cellcolor{mycolor!0}0.00	&	\cellcolor{mycolor!99.51}99.51	&		oc	&	\cellcolor{mycolor!0.1}0.10	&	\cellcolor{mycolor!0.79}0.79	&	\cellcolor{mycolor!20.55}20.55	&	\cellcolor{mycolor!0.3}0.30	&	\cellcolor{mycolor!0.4}0.40	&	\cellcolor{mycolor!59.39}59.39\\

et	&	\cellcolor{mycolor!2.27}2.27	&	\cellcolor{mycolor!8.5}8.50	&	\cellcolor{mycolor!75.49}75.49	&	\cellcolor{mycolor!2.96}2.96	&	\cellcolor{mycolor!0.1}0.10	&	\cellcolor{mycolor!96.34}96.34	&		om	&	\cellcolor{mycolor!0.2}0.20	&	\cellcolor{mycolor!0}0.00	&	\cellcolor{mycolor!38.04}38.04	&	\cellcolor{mycolor!0}0.00	&	\cellcolor{mycolor!0.2}0.20	&	\cellcolor{mycolor!0}0.00\\

fa	&	\cellcolor{mycolor!0.4}0.40	&	\cellcolor{mycolor!45.95}45.95	&	\cellcolor{mycolor!34.49}34.49	&	\cellcolor{mycolor!57.61}57.61	&	\cellcolor{mycolor!0}0.00	&	\cellcolor{mycolor!98.12}98.12	&		or	&	\cellcolor{mycolor!1.28}1.28	&	\cellcolor{mycolor!37.35}37.35	&	\cellcolor{mycolor!62.65}62.65	&	\cellcolor{mycolor!1.78}1.78	&	\cellcolor{mycolor!0}0.00	&	\cellcolor{mycolor!99.8}99.80\\

ff	&	\cellcolor{mycolor!0.49}0.49	&	\cellcolor{mycolor!0}0.00	&	\cellcolor{mycolor!73.81}73.81	&	\cellcolor{mycolor!0}0.00	&	\cellcolor{mycolor!0.59}0.59	&	\cellcolor{mycolor!0}0.00	&		pa	&	\cellcolor{mycolor!1.28}1.28	&	\cellcolor{mycolor!49.41}49.41	&	\cellcolor{mycolor!39.62}39.62	&	\cellcolor{mycolor!5.43}5.43	&	\cellcolor{mycolor!0}0.00	&	\cellcolor{mycolor!100}100.00\\

fi	&	\cellcolor{mycolor!3.95}3.95	&	\cellcolor{mycolor!55.43}55.43	&	\cellcolor{mycolor!65.22}65.22	&	\cellcolor{mycolor!17.59}17.59	&	\cellcolor{mycolor!0.3}0.30	&	\cellcolor{mycolor!97.13}97.13	&	pl	&	\cellcolor{mycolor!0.2}0.20	&	\cellcolor{mycolor!64.33}64.33	&	\cellcolor{mycolor!12.55}12.55	&	\cellcolor{mycolor!58.5}58.50	&	\cellcolor{mycolor!0}0.00	&	\cellcolor{mycolor!98.42}98.42\\

fr	&	\cellcolor{mycolor!0.1}0.10	&	\cellcolor{mycolor!94.17}94.17	&	\cellcolor{mycolor!3.46}3.46	&	\cellcolor{mycolor!92.98}92.98	&	\cellcolor{mycolor!0}0.00	&	\cellcolor{mycolor!98.72}98.72	&		ps	&	\cellcolor{mycolor!0.99}0.99	&	\cellcolor{mycolor!20.16}20.16	&	\cellcolor{mycolor!39.03}39.03	&	\cellcolor{mycolor!0.49}0.49	&	\cellcolor{mycolor!0}0.00	&	\cellcolor{mycolor!97.83}97.83\\

ga	&	\cellcolor{mycolor!0.2}0.20	&	\cellcolor{mycolor!19.37}19.37	&	\cellcolor{mycolor!8.7}8.70	&	\cellcolor{mycolor!6.82}6.82	&	\cellcolor{mycolor!0}0.00	&	\cellcolor{mycolor!93.08}93.08	&		pt	&	\cellcolor{mycolor!0.3}0.30	&	\cellcolor{mycolor!84.39}84.39	&	\cellcolor{mycolor!5.34}5.34	&	\cellcolor{mycolor!79.84}79.84	&	\cellcolor{mycolor!0.1}0.10	&	\cellcolor{mycolor!98.42}98.42\\

gl	&	\cellcolor{mycolor!0.2}0.20	&	\cellcolor{mycolor!0.89}0.89	&	\cellcolor{mycolor!26.19}26.19	&	\cellcolor{mycolor!0.1}0.10	&	\cellcolor{mycolor!0.2}0.20	&	\cellcolor{mycolor!83.99}83.99	&		ro	&	\cellcolor{mycolor!0.1}0.10	&	\cellcolor{mycolor!19.57}19.57	&	\cellcolor{mycolor!26.98}26.98	&	\cellcolor{mycolor!42.39}42.39	&	\cellcolor{mycolor!0.2}0.20	&	\cellcolor{mycolor!87.15}87.15\\

gu	&	\cellcolor{mycolor!0.59}0.59	&	\cellcolor{mycolor!36.96}36.96	&	\cellcolor{mycolor!45.65}45.65	&	\cellcolor{mycolor!29.74}29.74	&	\cellcolor{mycolor!0}0.00	&	\cellcolor{mycolor!99.6}99.60	&		ru	&	\cellcolor{mycolor!0.69}0.69	&	\cellcolor{mycolor!79.74}79.74	&	\cellcolor{mycolor!46.64}46.64	&	\cellcolor{mycolor!37.06}37.06	&	\cellcolor{mycolor!0.1}0.10	&	\cellcolor{mycolor!99.01}99.01\\

ha	&	\cellcolor{mycolor!0.79}0.79	&	\cellcolor{mycolor!0}0.00	&	\cellcolor{mycolor!67.98}67.98	&	\cellcolor{mycolor!0}0.00	&	\cellcolor{mycolor!0.1}0.10	&	\cellcolor{mycolor!0}0.00	&		sd	&	\cellcolor{mycolor!0.89}0.89	&	\cellcolor{mycolor!7.41}7.41	&	\cellcolor{mycolor!41.7}41.70	&	\cellcolor{mycolor!0.2}0.20	&	\cellcolor{mycolor!0}0.00	&	\cellcolor{mycolor!95.16}95.16\\

he	&	\cellcolor{mycolor!1.68}1.68	&	\cellcolor{mycolor!58.7}58.70	&	\cellcolor{mycolor!65.51}65.51	&	\cellcolor{mycolor!31.03}31.03	&	\cellcolor{mycolor!0}0.00	&	\cellcolor{mycolor!100}100	&		sk	&	\cellcolor{mycolor!0.4}0.40	&	\cellcolor{mycolor!20.26}20.26	&	\cellcolor{mycolor!25.4}25.40	&	\cellcolor{mycolor!3.56}3.56	&	\cellcolor{mycolor!0.1}0.10	&	\cellcolor{mycolor!97.23}97.23\\

hi	&	\cellcolor{mycolor!0.79}0.79	&	\cellcolor{mycolor!50.79}50.79	&	\cellcolor{mycolor!55.83}55.83	&	\cellcolor{mycolor!23.81}23.81	&	\cellcolor{mycolor!0}0.00	&	\cellcolor{mycolor!98.91}98.91	&		sl	&	\cellcolor{mycolor!1.19}1.19	&	\cellcolor{mycolor!37.25}37.25	&	\cellcolor{mycolor!49.6}49.60	&	\cellcolor{mycolor!16.21}16.21	&	\cellcolor{mycolor!0.69}0.69	&	\cellcolor{mycolor!91.9}91.90\\

hr	&	\cellcolor{mycolor!0.49}0.49	&	\cellcolor{mycolor!41.6}41.60	&	\cellcolor{mycolor!20.95}20.95	&	\cellcolor{mycolor!20.36}20.36	&	\cellcolor{mycolor!0.1}0.10	&	\cellcolor{mycolor!69.66}69.66	&		sn	&	\cellcolor{mycolor!0.49}0.49	&	\cellcolor{mycolor!0}0.00	&	\cellcolor{mycolor!34.58}34.58	&	\cellcolor{mycolor!0}0.00	&	\cellcolor{mycolor!0.1}0.10	&	\cellcolor{mycolor!0}0.00 \\

hu	&	\cellcolor{mycolor!0.4}0.40	&	\cellcolor{mycolor!64.33}64.33	&	\cellcolor{mycolor!27.47}27.47	&	\cellcolor{mycolor!38.74}38.74	&	\cellcolor{mycolor!0.1}0.10	&	\cellcolor{mycolor!97.13}97.13	&		so	&	\cellcolor{mycolor!0.3}0.30	&	\cellcolor{mycolor!8.7}8.70	&	\cellcolor{mycolor!58.7}58.70	&	\cellcolor{mycolor!0.2}0.20	&	\cellcolor{mycolor!0.1}0.10	&	\cellcolor{mycolor!57.71}57.71 \\

hy	&	\cellcolor{mycolor!4.74}4.74	&	\cellcolor{mycolor!47.13}47.13	&	\cellcolor{mycolor!79.15}79.15	&	\cellcolor{mycolor!12.15}12.15	&	\cellcolor{mycolor!0}0.00	&	\cellcolor{mycolor!99.6}99.60	&		sr	&	\cellcolor{mycolor!0.59}0.59	&	\cellcolor{mycolor!12.45}12.45	&	\cellcolor{mycolor!17.89}17.89	&	\cellcolor{mycolor!18.87}18.87	&	\cellcolor{mycolor!0.2}0.20	&	\cellcolor{mycolor!48.02}48.02 \\

id	&	\cellcolor{mycolor!0.49}0.49	&	\cellcolor{mycolor!81.92}81.92	&	\cellcolor{mycolor!16.21}16.21	&	\cellcolor{mycolor!60.38}60.38	&	\cellcolor{mycolor!0}0.00	&	\cellcolor{mycolor!95.85}95.85	&		sv	&	\cellcolor{mycolor!0.1}0.10	&	\cellcolor{mycolor!47.33}47.33	&	\cellcolor{mycolor!46.94}46.94	&	\cellcolor{mycolor!25}25.00	&	\cellcolor{mycolor!0.1}0.10	&	\cellcolor{mycolor!96.94}96.94 \\

ig	&	\cellcolor{mycolor!0.2}0.20	&	\cellcolor{mycolor!0}0.00	&	\cellcolor{mycolor!51.48}51.48	&	\cellcolor{mycolor!0}0.00	&	\cellcolor{mycolor!0.1}0.10	&	\cellcolor{mycolor!0}0.00	&		sw	&	\cellcolor{mycolor!0.2}0.20	&	\cellcolor{mycolor!39.23}39.23	&	\cellcolor{mycolor!36.86}36.86	&	\cellcolor{mycolor!22.73}22.73	&	\cellcolor{mycolor!0}0.00	&	\cellcolor{mycolor!94.66}94.66 \\

is	&	\cellcolor{mycolor!0.4}0.40	&	\cellcolor{mycolor!35.08}35.08	&	\cellcolor{mycolor!40.02}40.02	&	\cellcolor{mycolor!28.46}28.46	&	\cellcolor{mycolor!0.2}0.20	&	\cellcolor{mycolor!92.98}92.98	&		ta	&	\cellcolor{mycolor!1.48}1.48	&	\cellcolor{mycolor!24.41}24.41	&	\cellcolor{mycolor!55.24}55.24	&	\cellcolor{mycolor!34.09}34.09	&	\cellcolor{mycolor!0}0.00	&	\cellcolor{mycolor!98.62}98.62 \\

it	&	\cellcolor{mycolor!0.49}0.49	&	\cellcolor{mycolor!79.55}79.55	&	\cellcolor{mycolor!3.36}3.36	&	\cellcolor{mycolor!77.57}77.57	&	\cellcolor{mycolor!0.1}0.10	&	\cellcolor{mycolor!98.42}98.42	&		te	&	\cellcolor{mycolor!1.38}1.38	&	\cellcolor{mycolor!38.93}38.93	&	\cellcolor{mycolor!69.47}69.47	&	\cellcolor{mycolor!28.56}28.56	&	\cellcolor{mycolor!0}0.00	&	\cellcolor{mycolor!99.6}99.60 \\

ja	&	\cellcolor{mycolor!48.02}48.02	&	\cellcolor{mycolor!16.7}16.70	&	\cellcolor{mycolor!28.36}28.36	&	\cellcolor{mycolor!70.95}70.95	&	\cellcolor{mycolor!6.62}6.62	&	\cellcolor{mycolor!92}92.00	&		tg	&	\cellcolor{mycolor!1.28}1.28	&	\cellcolor{mycolor!2.77}2.77	&	\cellcolor{mycolor!44.86}44.86	&	\cellcolor{mycolor!7.61}7.61	&	\cellcolor{mycolor!0.2}0.20	&	\cellcolor{mycolor!97.04}97.04 \\

jv	&	\cellcolor{mycolor!0.2}0.20	&	\cellcolor{mycolor!0}0.00	&	\cellcolor{mycolor!13.83}13.83	&	\cellcolor{mycolor!0}0.00	&	\cellcolor{mycolor!0}0.00	&	\cellcolor{mycolor!64.62}64.62	&		th	&	\cellcolor{mycolor!1.28}1.28	&	\cellcolor{mycolor!58.6}58.60	&	\cellcolor{mycolor!71.25}71.25	&	\cellcolor{mycolor!28.56}28.56	&	\cellcolor{mycolor!0}0.00	&	\cellcolor{mycolor!100}100.00 \\

ka	&	\cellcolor{mycolor!3.56}3.56	&	\cellcolor{mycolor!31.72}31.72	&	\cellcolor{mycolor!70.06}70.06	&	\cellcolor{mycolor!4.74}4.74	&	\cellcolor{mycolor!0}0.00	&	\cellcolor{mycolor!99.8}99.80	&	tl	&	\cellcolor{mycolor!0.2}0.20	&	\cellcolor{mycolor!66.7}66.7	&	\cellcolor{mycolor!32.91}32.91	&	\cellcolor{mycolor!45.75}45.75	&	\cellcolor{mycolor!0}0.00	&	\cellcolor{mycolor!98.91}98.91 \\

kam	&	\cellcolor{mycolor!0.99}0.99	&	\cellcolor{mycolor!0}0.00	&	\cellcolor{mycolor!65.51}65.51	&	\cellcolor{mycolor!0}0.00	&	\cellcolor{mycolor!1.58}1.58	&	\cellcolor{mycolor!0}0.00	&		tr	&	\cellcolor{mycolor!0.89}0.89	&	\cellcolor{mycolor!37.94}37.94	&	\cellcolor{mycolor!48.02}48.02	&	\cellcolor{mycolor!31.42}31.42	&	\cellcolor{mycolor!0}0.00	&	\cellcolor{mycolor!95.65}95.65 \\

kea	&	\cellcolor{mycolor!0.59}0.59	&	\cellcolor{mycolor!0}0.00	&	\cellcolor{mycolor!35.47}35.47	&	\cellcolor{mycolor!0}0.00	&	\cellcolor{mycolor!0.4}0.40	&	\cellcolor{mycolor!0}0.00	&		uk	&	\cellcolor{mycolor!0.49}0.49	&	\cellcolor{mycolor!71.54}71.54	&	\cellcolor{mycolor!10.38}10.38	&	\cellcolor{mycolor!28.06}28.06	&	\cellcolor{mycolor!0.49}0.49	&	\cellcolor{mycolor!98.62}98.62 \\

kk	&	\cellcolor{mycolor!0.99}0.99	&	\cellcolor{mycolor!45.95}45.95	&	\cellcolor{mycolor!37.06}37.06	&	\cellcolor{mycolor!29.45}29.45	&	\cellcolor{mycolor!0}0.00	&	\cellcolor{mycolor!98.32}98.32	&		umb	&	\cellcolor{mycolor!0.59}0.59	&	\cellcolor{mycolor!0}0.00	&	\cellcolor{mycolor!54.94}54.94	&	\cellcolor{mycolor!0}0.00	&	\cellcolor{mycolor!0.3}0.30	&	\cellcolor{mycolor!0}0.00 \\

km	&	\cellcolor{mycolor!1.58}1.58	&	\cellcolor{mycolor!29.25}29.25	&	\cellcolor{mycolor!58.89}58.89	&	\cellcolor{mycolor!28.26}28.26	&	\cellcolor{mycolor!0}0.00	&	\cellcolor{mycolor!100}100.00	&		ur	&	\cellcolor{mycolor!1.68}1.68	&	\cellcolor{mycolor!19.86}19.86	&	\cellcolor{mycolor!75.49}75.49	&	\cellcolor{mycolor!14.82}14.82	&	\cellcolor{mycolor!0.1}0.10	&	\cellcolor{mycolor!96.54}96.54 \\

kn	&	\cellcolor{mycolor!3.16}3.16	&	\cellcolor{mycolor!38.24}38.24	&	\cellcolor{mycolor!75.59}75.59	&	\cellcolor{mycolor!14.72}14.72	&	\cellcolor{mycolor!0}0.00	&	\cellcolor{mycolor!100}100.00	&		uz	&	\cellcolor{mycolor!0.2}0.20	&	\cellcolor{mycolor!30.24}30.24	&	\cellcolor{mycolor!58.99}58.99	&	\cellcolor{mycolor!2.77}2.77	&	\cellcolor{mycolor!0.1}0.10	&	\cellcolor{mycolor!89.92}89.92 \\

ko	&	\cellcolor{mycolor!3.85}3.85	&	\cellcolor{mycolor!71.94}71.94	&	\cellcolor{mycolor!75.69}75.69	&	\cellcolor{mycolor!23.52}23.52	&	\cellcolor{mycolor!0}0.00	&	\cellcolor{mycolor!98.02}98.02	&		vi	&	\cellcolor{mycolor!0.1}0.10	&	\cellcolor{mycolor!92.69}92.69	&	\cellcolor{mycolor!13.44}13.44	&	\cellcolor{mycolor!81.13}81.13	&	\cellcolor{mycolor!0}0.00	&	\cellcolor{mycolor!99.7}99.70 \\

ku	&	\cellcolor{mycolor!0.1}0.10	&	\cellcolor{mycolor!14.13}14.13	&	\cellcolor{mycolor!31.72}31.72	&	\cellcolor{mycolor!0}0.00	&	\cellcolor{mycolor!0.4}0.40	&	\cellcolor{mycolor!75.2}75.20	&		wo	&	\cellcolor{mycolor!0.3}0.30	&	\cellcolor{mycolor!0}0.00	&	\cellcolor{mycolor!56.62}56.62	&	\cellcolor{mycolor!0}0.00	&	\cellcolor{mycolor!0.49}0.49	&	\cellcolor{mycolor!0}0.00 \\

ky	&	\cellcolor{mycolor!1.19}1.19	&	\cellcolor{mycolor!25.99}25.99	&	\cellcolor{mycolor!48.62}48.62	&	\cellcolor{mycolor!4.35}4.35	&	\cellcolor{mycolor!0}0.00	&	\cellcolor{mycolor!99.11}99.11	&			xh	&	\cellcolor{mycolor!0.2}0.20	&	\cellcolor{mycolor!0}0.00	&	\cellcolor{mycolor!40.51}40.51	&	\cellcolor{mycolor!0}0.00	&	\cellcolor{mycolor!0.1}0.10	&	\cellcolor{mycolor!0}0.00 \\

lb	&	\cellcolor{mycolor!0.1}0.10	&	\cellcolor{mycolor!24.21}24.21	&	\cellcolor{mycolor!30.73}30.73	&	\cellcolor{mycolor!0.4}0.40	&	\cellcolor{mycolor!0.59}0.59	&	\cellcolor{mycolor!89.53}89.53	&	yo	&	\cellcolor{mycolor!0.1}0.10	&	\cellcolor{mycolor!3.56}3.56	&	\cellcolor{mycolor!57.91}57.91	&	\cellcolor{mycolor!0.4}0.40	&	\cellcolor{mycolor!0.1}0.10	&	\cellcolor{mycolor!15.81}15.81 \\

lg	&	\cellcolor{mycolor!10.57}10.57	&	\cellcolor{mycolor!0}0.00	&	\cellcolor{mycolor!79.35}79.35	&	\cellcolor{mycolor!0}0.00	&	\cellcolor{mycolor!6.13}6.13	&	\cellcolor{mycolor!0}0.00	&		zhtrad	&	\cellcolor{mycolor!98.12}98.12	&	\cellcolor{mycolor!0}0.00	&	\cellcolor{mycolor!98.42}98.42	&	\cellcolor{mycolor!0}0.00	&	\cellcolor{mycolor!99.51}99.51	&	\cellcolor{mycolor!0}0.00 \\

	&		&		&		&		&		&		&			zu	&	\cellcolor{mycolor!0.2}0.20	&	\cellcolor{mycolor!0}0.00	&	\cellcolor{mycolor!45.55}45.55	&	\cellcolor{mycolor!0}0.00	&	\cellcolor{mycolor!0.1}0.10	&	\cellcolor{mycolor!0}0.00 \\

\midrule
    \end{tabular}}
    \caption{Using langdetect~\citep{joulin2016fasttext}, we individually identify the language of the translation output in zh$\rightarrow$X (where X represents any of the 101 languages included in Flores-101) for the LLaMA2-Alpaca, ChineseLLaMA2-Alpaca, and \name-Alpaca models on the Flores-101 devtest. $R_\mathrm{zh}$ refers to the proportion of sentences in the zh$\rightarrow$X translation output where the top predicted language is Chinese. $R_\mathrm{X}$, on the other hand, denotes the proportion where the top prediction corresponds to the target translated language.}
    \label{tab:compare_zh}
\end{table*}

\begin{table*}[!ht]
    \centering
    \footnotesize
    \resizebox{1\linewidth}{!}{
    \begin{tabular}{c|cc|cc|cc||c|cc|cc|cc}
\toprule
 \multirow{2}{*}{\textbf{X}} & \multicolumn{2}{c|}{\textbf{LLaMA2-Alpaca}} & \multicolumn{2}{c|}{\textbf{Swallow}} & \multicolumn{2}{c||}{\textbf{\name-Alpaca}}  &  \multirow{2}{*}{\textbf{X}}
 & \multicolumn{2}{c|}{\textbf{LLaMA2-Alpaca}} & \multicolumn{2}{c|}{\textbf{Swallow}} & \multicolumn{2}{c}{\textbf{\name-Alpaca}} \\
 & \textbf{$R_\mathrm{ja}$} & \textbf{$R_\mathrm{X}$} & \textbf{$R_\mathrm{ja}$} & \textbf{$R_\mathrm{X}$} & \textbf{$R_\mathrm{ja}$} & \textbf{$R_\mathrm{X}$} & & \textbf{$R_\mathrm{ja}$} & \textbf{$R_\mathrm{X}$}  & \textbf{$R_\mathrm{ja}$} & \textbf{$R_\mathrm{X}$} & \textbf{$R_\mathrm{ja}$} & \textbf{$R_\mathrm{X}$} \\
 \midrule
af & \cellcolor{mycolor!0.2}0.20 & \cellcolor{mycolor!35.28}35.28 & \cellcolor{mycolor!72.23}72.23 & \cellcolor{mycolor!0}0.00 & \cellcolor{mycolor!0.59}0.59 & \cellcolor{mycolor!75.69}75.69 & lo & \cellcolor{mycolor!0.3}0.30 & \cellcolor{mycolor!37.85}37.85 & \cellcolor{mycolor!75.89}75.89 & \cellcolor{mycolor!0.1}0.10 & \cellcolor{mycolor!0}0.00 & \cellcolor{mycolor!54.55}54.55 \\
am & \cellcolor{mycolor!0.2}0.20 & \cellcolor{mycolor!61.96}61.96 & \cellcolor{mycolor!77.67}77.67 & \cellcolor{mycolor!0.1}0.10 & \cellcolor{mycolor!0.69}0.69 & \cellcolor{mycolor!90.91}90.91 & lt & \cellcolor{mycolor!4.74}4.74 & \cellcolor{mycolor!32.41}32.41 & \cellcolor{mycolor!70.85}70.85 & \cellcolor{mycolor!4.55}4.55 & \cellcolor{mycolor!3.66}3.66 & \cellcolor{mycolor!94.76}94.76 \\
ar & \cellcolor{mycolor!0.69}0.69 & \cellcolor{mycolor!93.97}93.97 & \cellcolor{mycolor!64.72}64.72 & \cellcolor{mycolor!13.93}13.93 & \cellcolor{mycolor!0}0.00 & \cellcolor{mycolor!99.9}99.90 & luo & \cellcolor{mycolor!0.49}0.49 & \cellcolor{mycolor!0}0.00 & \cellcolor{mycolor!71.25}71.25 & \cellcolor{mycolor!0}0.00 & \cellcolor{mycolor!0.89}0.89 & \cellcolor{mycolor!0}0.00 \\
as & \cellcolor{mycolor!3.66}3.66 & \cellcolor{mycolor!1.38}1.38 & \cellcolor{mycolor!74.01}74.01 & \cellcolor{mycolor!0}0.00 & \cellcolor{mycolor!0.1}0.10 & \cellcolor{mycolor!73.22}73.22 & lv & \cellcolor{mycolor!1.09}1.09 & \cellcolor{mycolor!39.92}39.92 & \cellcolor{mycolor!66.8}66.80 & \cellcolor{mycolor!5.53}5.53 & \cellcolor{mycolor!1.68}1.68 & \cellcolor{mycolor!95.36}95.36 \\
ast & \cellcolor{mycolor!0.2}0.20 & \cellcolor{mycolor!1.48}1.48 & \cellcolor{mycolor!71.44}71.44 & \cellcolor{mycolor!0}0.00 & \cellcolor{mycolor!0.2}0.20 & \cellcolor{mycolor!34.19}34.19 & mi & \cellcolor{mycolor!0.2}0.20 & \cellcolor{mycolor!0}0.00 & \cellcolor{mycolor!61.46}61.46 & \cellcolor{mycolor!0}0.00 & \cellcolor{mycolor!0.2}0.20 & \cellcolor{mycolor!0}0.00 \\
az & \cellcolor{mycolor!0.2}0.20 & \cellcolor{mycolor!26.58}26.58 & \cellcolor{mycolor!69.57}69.57 & \cellcolor{mycolor!5.53}5.53 & \cellcolor{mycolor!0.3}0.30 & \cellcolor{mycolor!97.43}97.43 & mk & \cellcolor{mycolor!0.3}0.30 & \cellcolor{mycolor!17.98}17.98 & \cellcolor{mycolor!78.46}78.46 & \cellcolor{mycolor!0}0.00 & \cellcolor{mycolor!0.49}0.49 & \cellcolor{mycolor!98.81}98.81\\
be & \cellcolor{mycolor!0.4}0.40 & \cellcolor{mycolor!60.18}60.18 & \cellcolor{mycolor!72.92}72.92 & \cellcolor{mycolor!0}0.00 & \cellcolor{mycolor!0.2}0.20 & \cellcolor{mycolor!99.11}99.11 & ml & \cellcolor{mycolor!1.28}1.28 & \cellcolor{mycolor!36.17}36.17 & \cellcolor{mycolor!74.41}74.41 & \cellcolor{mycolor!1.68}1.68 & \cellcolor{mycolor!0.49}0.49 & \cellcolor{mycolor!70.75}70.75\\
bg & \cellcolor{mycolor!1.09}1.09 & \cellcolor{mycolor!60.28}60.28 & \cellcolor{mycolor!77.67}77.67 & \cellcolor{mycolor!0.3}0.30 & \cellcolor{mycolor!0.89}0.89 & \cellcolor{mycolor!98.02}98.02 & mn & \cellcolor{mycolor!0.59}0.59 & \cellcolor{mycolor!35.18}35.18 & \cellcolor{mycolor!75.59}75.59 & \cellcolor{mycolor!1.48}1.48 & \cellcolor{mycolor!0}0.00 & \cellcolor{mycolor!99.31}99.31\\
bn & \cellcolor{mycolor!1.78}1.78 & \cellcolor{mycolor!64.62}64.62 & \cellcolor{mycolor!75.69}75.69 & \cellcolor{mycolor!1.78}1.78 & \cellcolor{mycolor!0}0.00 & \cellcolor{mycolor!99.9}99.90 & mr & \cellcolor{mycolor!0.59}0.59 & \cellcolor{mycolor!35.87}35.87 & \cellcolor{mycolor!76.88}76.88 & \cellcolor{mycolor!0}0.00 & \cellcolor{mycolor!0.1}0.10 & \cellcolor{mycolor!99.01}99.01\\
bs & \cellcolor{mycolor!0.69}0.69 & \cellcolor{mycolor!1.38}1.38 & \cellcolor{mycolor!73.52}73.52 & \cellcolor{mycolor!0}0.00 & \cellcolor{mycolor!1.98}1.98 & \cellcolor{mycolor!3.16}3.16 & ms & \cellcolor{mycolor!0.1}0.10 & \cellcolor{mycolor!5.53}5.53 & \cellcolor{mycolor!61.86}61.86 & \cellcolor{mycolor!0.2}0.20 & \cellcolor{mycolor!0}0.00 & \cellcolor{mycolor!39.92}39.92\\
ca & \cellcolor{mycolor!0.4}0.40 & \cellcolor{mycolor!89.92}89.92 & \cellcolor{mycolor!65.02}65.02 & \cellcolor{mycolor!11.07}11.07 & \cellcolor{mycolor!0.49}0.49 & \cellcolor{mycolor!98.12}98.12 & mt & \cellcolor{mycolor!0.4}0.40 & \cellcolor{mycolor!60.08}60.08 & \cellcolor{mycolor!68.38}68.38 & \cellcolor{mycolor!3.16}3.16 & \cellcolor{mycolor!0.69}0.69 & \cellcolor{mycolor!94.07}94.07\\
ceb & \cellcolor{mycolor!0.1}0.10 & \cellcolor{mycolor!33.3}33.30 & \cellcolor{mycolor!44.57}44.57 & \cellcolor{mycolor!3.56}3.56 & \cellcolor{mycolor!0}0.00 & \cellcolor{mycolor!95.06}95.06 & my & \cellcolor{mycolor!1.68}1.68 & \cellcolor{mycolor!56.03}56.03 & \cellcolor{mycolor!78.85}78.85 & \cellcolor{mycolor!1.48}1.48 & \cellcolor{mycolor!0.1}0.10 & \cellcolor{mycolor!99.9}99.90\\
cs & \cellcolor{mycolor!1.19}1.19 & \cellcolor{mycolor!61.46}61.46 & \cellcolor{mycolor!72.13}72.13 & \cellcolor{mycolor!5.24}5.24 & \cellcolor{mycolor!1.68}1.68 & \cellcolor{mycolor!93.38}93.38 & ne & \cellcolor{mycolor!0.2}0.20 & \cellcolor{mycolor!50}50.00 & \cellcolor{mycolor!70.45}70.45 & \cellcolor{mycolor!0}0.00 & \cellcolor{mycolor!0}0.00 & \cellcolor{mycolor!99.01}99.01\\
cy & \cellcolor{mycolor!0.2}0.20 & \cellcolor{mycolor!30.83}30.83 & \cellcolor{mycolor!66.9}66.90 & \cellcolor{mycolor!2.47}2.47 & \cellcolor{mycolor!0.2}0.20 & \cellcolor{mycolor!98.52}98.52 & nl & \cellcolor{mycolor!0.4}0.40 & \cellcolor{mycolor!76.78}76.78 & \cellcolor{mycolor!61.36}61.36 & \cellcolor{mycolor!22.33}22.33 & \cellcolor{mycolor!0.2}0.20 & \cellcolor{mycolor!92.09}92.09\\
da & \cellcolor{mycolor!0.79}0.79 & \cellcolor{mycolor!57.51}57.51 & \cellcolor{mycolor!70.06}70.06 & \cellcolor{mycolor!4.64}4.64 & \cellcolor{mycolor!0.59}0.59 & \cellcolor{mycolor!91.8}91.80 & no & \cellcolor{mycolor!1.38}1.38 & \cellcolor{mycolor!44.47}44.47 & \cellcolor{mycolor!69.57}69.57 & \cellcolor{mycolor!3.16}3.16 & \cellcolor{mycolor!0.69}0.69 & \cellcolor{mycolor!86.66}86.66\\
de & \cellcolor{mycolor!1.28}1.28 & \cellcolor{mycolor!83.4}83.40 & \cellcolor{mycolor!57.41}57.41 & \cellcolor{mycolor!29.25}29.25 & \cellcolor{mycolor!1.28}1.28 & \cellcolor{mycolor!94.17}94.17 & ns & \cellcolor{mycolor!1.58}1.58 & \cellcolor{mycolor!0}0.00 & \cellcolor{mycolor!62.55}62.55 & \cellcolor{mycolor!0}0.00 & \cellcolor{mycolor!1.38}1.38 & \cellcolor{mycolor!0}0.00\\
el & \cellcolor{mycolor!1.09}1.09 & \cellcolor{mycolor!42}42.00 & \cellcolor{mycolor!75.2}75.20 & \cellcolor{mycolor!7.41}7.41 & \cellcolor{mycolor!0}0.00 & \cellcolor{mycolor!100}100.00 & ny & \cellcolor{mycolor!0.49}0.49 & \cellcolor{mycolor!0}0.00 & \cellcolor{mycolor!72.53}72.53 & \cellcolor{mycolor!0}0.00 & \cellcolor{mycolor!0.79}0.79 & \cellcolor{mycolor!0}0.00\\
en & \cellcolor{mycolor!0}0.00 & \cellcolor{mycolor!100}100.00 & \cellcolor{mycolor!67.29}67.29 & \cellcolor{mycolor!32.41}32.41 & \cellcolor{mycolor!0}0.00 & \cellcolor{mycolor!100}100.00 & oc & \cellcolor{mycolor!0.2}0.20 & \cellcolor{mycolor!1.09}1.09 & \cellcolor{mycolor!68.97}68.97 & \cellcolor{mycolor!0}0.00 & \cellcolor{mycolor!0.59}0.59 & \cellcolor{mycolor!58.1}58.10\\
es & \cellcolor{mycolor!0.4}0.40 & \cellcolor{mycolor!97.04}97.04 & \cellcolor{mycolor!57.81}57.81 & \cellcolor{mycolor!20.26}20.26 & \cellcolor{mycolor!0.1}0.10 & \cellcolor{mycolor!99.21}99.21 & om & \cellcolor{mycolor!0.3}0.30 & \cellcolor{mycolor!0}0.00 & \cellcolor{mycolor!72.53}72.53 & \cellcolor{mycolor!0}0.00 & \cellcolor{mycolor!2.57}2.57 & \cellcolor{mycolor!0}0.00\\
et & \cellcolor{mycolor!0.69}0.69 & \cellcolor{mycolor!14.03}14.03 & \cellcolor{mycolor!68.48}68.48 & \cellcolor{mycolor!8.7}8.70 & \cellcolor{mycolor!4.35}4.35 & \cellcolor{mycolor!89.13}89.13 & or & \cellcolor{mycolor!0.69}0.69 & \cellcolor{mycolor!61.86}61.86 & \cellcolor{mycolor!79.45}79.45 & \cellcolor{mycolor!0}0.00 & \cellcolor{mycolor!1.09}1.09 & \cellcolor{mycolor!98.52}98.52\\
fa & \cellcolor{mycolor!0.3}0.30 & \cellcolor{mycolor!83.89}83.89 & \cellcolor{mycolor!75.79}75.79 & \cellcolor{mycolor!4.35}4.35 & \cellcolor{mycolor!0}0.00 & \cellcolor{mycolor!98.42}98.42 & pa & \cellcolor{mycolor!0.4}0.40 & \cellcolor{mycolor!77.67}77.67 & \cellcolor{mycolor!72.04}72.04 & \cellcolor{mycolor!1.78}1.78 & \cellcolor{mycolor!0.79}0.79 & \cellcolor{mycolor!98.91}98.91\\
ff & \cellcolor{mycolor!0.69}0.69 & \cellcolor{mycolor!0}0.00 & \cellcolor{mycolor!73.12}73.12 & \cellcolor{mycolor!0}0.00 & \cellcolor{mycolor!11.96}11.96 & \cellcolor{mycolor!0}0.00 & pl & \cellcolor{mycolor!0.79}0.79 & \cellcolor{mycolor!73.32}73.32 & \cellcolor{mycolor!71.54}71.54 & \cellcolor{mycolor!8.4}8.40 & \cellcolor{mycolor!0.49}0.49 & \cellcolor{mycolor!98.02}98.02\\
fi & \cellcolor{mycolor!3.36}3.36 & \cellcolor{mycolor!74.11}74.11 & \cellcolor{mycolor!66.01}66.01 & \cellcolor{mycolor!17.39}17.39 & \cellcolor{mycolor!2.37}2.37 & \cellcolor{mycolor!96.25}96.25 & ps & \cellcolor{mycolor!0.2}0.20 & \cellcolor{mycolor!43.28}43.28 & \cellcolor{mycolor!75.4}75.40 & \cellcolor{mycolor!0}0.00 & \cellcolor{mycolor!0}0.00 & \cellcolor{mycolor!98.22}98.22\\
fr & \cellcolor{mycolor!0.49}0.49 & \cellcolor{mycolor!97.04}97.04 & \cellcolor{mycolor!52.47}52.47 & \cellcolor{mycolor!34.29}34.29 & \cellcolor{mycolor!0}0.00 & \cellcolor{mycolor!99.7}99.70 & pt & \cellcolor{mycolor!1.09}1.09 & \cellcolor{mycolor!90.71}90.71 & \cellcolor{mycolor!63.14}63.14 & \cellcolor{mycolor!8.2}8.20 & \cellcolor{mycolor!0.2}0.20 & \cellcolor{mycolor!98.22}98.22\\
ga & \cellcolor{mycolor!0.2}0.20 & \cellcolor{mycolor!26.98}26.98 & \cellcolor{mycolor!64.23}64.23 & \cellcolor{mycolor!2.96}2.96 & \cellcolor{mycolor!0}0.00 & \cellcolor{mycolor!94.07}94.07 & ro & \cellcolor{mycolor!0.3}0.30 & \cellcolor{mycolor!45.95}45.95 & \cellcolor{mycolor!68.97}68.97 & \cellcolor{mycolor!4.25}4.25 & \cellcolor{mycolor!0.3}0.30 & \cellcolor{mycolor!89.53}89.53\\
gl & \cellcolor{mycolor!0.1}0.10 & \cellcolor{mycolor!1.58}1.58 & \cellcolor{mycolor!63.34}63.34 & \cellcolor{mycolor!3.56}3.56 & \cellcolor{mycolor!0.2}0.20 & \cellcolor{mycolor!83.3}83.30 & ru & \cellcolor{mycolor!0.3}0.30 & \cellcolor{mycolor!83.1}83.10 & \cellcolor{mycolor!71.44}71.44 & \cellcolor{mycolor!12.45}12.45 & \cellcolor{mycolor!0.2}0.20 & \cellcolor{mycolor!99.41}99.41\\
gu & \cellcolor{mycolor!0.3}0.30 & \cellcolor{mycolor!67.59}67.59 & \cellcolor{mycolor!77.47}77.47 & \cellcolor{mycolor!0.99}0.99 & \cellcolor{mycolor!1.48}1.48 & \cellcolor{mycolor!96.64}96.64 & sd & \cellcolor{mycolor!0.89}0.89 & \cellcolor{mycolor!2.47}2.47 & \cellcolor{mycolor!74.31}74.31 & \cellcolor{mycolor!0}0.00 & \cellcolor{mycolor!0}0.00 & \cellcolor{mycolor!92.59}92.59\\
ha & \cellcolor{mycolor!0.59}0.59 & \cellcolor{mycolor!0}0.00 & \cellcolor{mycolor!70.06}70.06 & \cellcolor{mycolor!0}0.00 & \cellcolor{mycolor!0.99}0.99 & \cellcolor{mycolor!0}0.00 & sk & \cellcolor{mycolor!0.49}0.49 & \cellcolor{mycolor!27.27}27.27 & \cellcolor{mycolor!65.42}65.42 & \cellcolor{mycolor!7.81}7.81 & \cellcolor{mycolor!0.59}0.59 & \cellcolor{mycolor!94.57}94.57\\
he & \cellcolor{mycolor!1.78}1.78 & \cellcolor{mycolor!76.19}76.19 & \cellcolor{mycolor!63.34}63.34 & \cellcolor{mycolor!16.6}16.60 & \cellcolor{mycolor!0}0.00 & \cellcolor{mycolor!100}100.00 & sl & \cellcolor{mycolor!0.79}0.79 & \cellcolor{mycolor!58.79}58.79 & \cellcolor{mycolor!61.66}61.66 & \cellcolor{mycolor!3.56}3.56 & \cellcolor{mycolor!1.38}1.38 & \cellcolor{mycolor!91.11}91.11\\
hi & \cellcolor{mycolor!0.69}0.69 & \cellcolor{mycolor!70.75}70.75 & \cellcolor{mycolor!67.98}67.98 & \cellcolor{mycolor!7.91}7.91 & \cellcolor{mycolor!0}0.00 & \cellcolor{mycolor!99.9}99.90 & sn & \cellcolor{mycolor!0.4}0.40 & \cellcolor{mycolor!0}0.00 & \cellcolor{mycolor!68.18}68.18 & \cellcolor{mycolor!0}0.00 & \cellcolor{mycolor!1.58}1.58 & \cellcolor{mycolor!0}0.00\\
hr & \cellcolor{mycolor!0.89}0.89 & \cellcolor{mycolor!54.55}54.55 & \cellcolor{mycolor!69.37}69.37 & \cellcolor{mycolor!1.28}1.28 & \cellcolor{mycolor!1.19}1.19 & \cellcolor{mycolor!66.6}66.60 & so & \cellcolor{mycolor!0.1}0.10 & \cellcolor{mycolor!7.71}7.71 & \cellcolor{mycolor!74.31}74.31 & \cellcolor{mycolor!0.2}0.20 & \cellcolor{mycolor!0.99}0.99 & \cellcolor{mycolor!59.19}59.19\\
hu & \cellcolor{mycolor!0.4}0.40 & \cellcolor{mycolor!69.96}69.96 & \cellcolor{mycolor!71.44}71.44 & \cellcolor{mycolor!10.67}10.67 & \cellcolor{mycolor!0.3}0.30 & \cellcolor{mycolor!93.87}93.87 & sr & \cellcolor{mycolor!1.48}1.48 & \cellcolor{mycolor!15.22}15.22 & \cellcolor{mycolor!75.49}75.49 & \cellcolor{mycolor!1.48}1.48 & \cellcolor{mycolor!1.98}1.98 & \cellcolor{mycolor!44.07}44.07\\
hy & \cellcolor{mycolor!0.69}0.69 & \cellcolor{mycolor!77.08}77.08 & \cellcolor{mycolor!79.55}79.55 & \cellcolor{mycolor!1.09}1.09 & \cellcolor{mycolor!0}0.00 & \cellcolor{mycolor!99.9}99.90 & sv & \cellcolor{mycolor!2.57}2.57 & \cellcolor{mycolor!49.9}49.90 & \cellcolor{mycolor!66.01}66.01 & \cellcolor{mycolor!13.34}13.34 & \cellcolor{mycolor!1.68}1.68 & \cellcolor{mycolor!95.16}95.16\\
id & \cellcolor{mycolor!0.2}0.20 & \cellcolor{mycolor!84.98}84.98 & \cellcolor{mycolor!70.65}70.65 & \cellcolor{mycolor!7.61}7.61 & \cellcolor{mycolor!0}0.00 & \cellcolor{mycolor!97.04}97.04 & sw & \cellcolor{mycolor!0.2}0.20 & \cellcolor{mycolor!48.32}48.32 & \cellcolor{mycolor!67.49}67.49 & \cellcolor{mycolor!0.99}0.99 & \cellcolor{mycolor!0.59}0.59 & \cellcolor{mycolor!94.76}94.76\\
ig & \cellcolor{mycolor!0.1}0.10 & \cellcolor{mycolor!0}0.00 & \cellcolor{mycolor!74.8}74.80 & \cellcolor{mycolor!0}0.00 & \cellcolor{mycolor!0.2}0.20 & \cellcolor{mycolor!0}0.00 & ta & \cellcolor{mycolor!0.3}0.30 & \cellcolor{mycolor!53.46}53.46 & \cellcolor{mycolor!74.31}74.31 & \cellcolor{mycolor!1.98}1.98 & \cellcolor{mycolor!0}0.00 & \cellcolor{mycolor!99.8}99.80\\
is & \cellcolor{mycolor!0.3}0.30 & \cellcolor{mycolor!55.34}55.34 & \cellcolor{mycolor!58.2}58.20 & \cellcolor{mycolor!19.76}19.76 & \cellcolor{mycolor!0.2}0.20 & \cellcolor{mycolor!95.06}95.06 & te & \cellcolor{mycolor!0.2}0.20 & \cellcolor{mycolor!73.12}73.12 & \cellcolor{mycolor!75.79}75.79 & \cellcolor{mycolor!2.47}2.47 & \cellcolor{mycolor!0}0.00 & \cellcolor{mycolor!99.8}99.80\\
it & \cellcolor{mycolor!0.59}0.59 & \cellcolor{mycolor!85.47}85.47 & \cellcolor{mycolor!55.24}55.24 & \cellcolor{mycolor!24.11}24.11 & \cellcolor{mycolor!0}0.00 & \cellcolor{mycolor!97.63}97.63 & tg & \cellcolor{mycolor!0.69}0.69 & \cellcolor{mycolor!6.23}6.23 & \cellcolor{mycolor!74.01}74.01 & \cellcolor{mycolor!0}0.00 & \cellcolor{mycolor!0.4}0.40 & \cellcolor{mycolor!97.33}97.33\\
jv & \cellcolor{mycolor!1.38}1.38 & \cellcolor{mycolor!0.1}0.10 & \cellcolor{mycolor!66.9}66.90 & \cellcolor{mycolor!0}0.00 & \cellcolor{mycolor!0.89}0.89 & \cellcolor{mycolor!67.79}67.79 & th & \cellcolor{mycolor!0}0.00 & \cellcolor{mycolor!84.39}84.39 & \cellcolor{mycolor!70.75}70.75 & \cellcolor{mycolor!12.15}12.15 & \cellcolor{mycolor!0}0.00 & \cellcolor{mycolor!100}100.00\\
ka & \cellcolor{mycolor!1.28}1.28 & \cellcolor{mycolor!63.14}63.14 & \cellcolor{mycolor!65.91}65.91 & \cellcolor{mycolor!16.01}16.01 & \cellcolor{mycolor!0}0.00 & \cellcolor{mycolor!100}100.00 & tl & \cellcolor{mycolor!0.2}0.20 & \cellcolor{mycolor!73.62}73.62 & \cellcolor{mycolor!62.94}62.94 & \cellcolor{mycolor!6.72}6.72 & \cellcolor{mycolor!0.1}0.10 & \cellcolor{mycolor!99.31}99.31\\
kam & \cellcolor{mycolor!0.3}0.30 & \cellcolor{mycolor!0}0.00 & \cellcolor{mycolor!73.22}73.22 & \cellcolor{mycolor!0}0.00 & \cellcolor{mycolor!3.56}3.56 & \cellcolor{mycolor!0}0.00 & tr & \cellcolor{mycolor!0.79}0.79 & \cellcolor{mycolor!42.39}42.39 & \cellcolor{mycolor!67.69}67.69 & \cellcolor{mycolor!11.86}11.86 & \cellcolor{mycolor!0.4}0.40 & \cellcolor{mycolor!95.26}95.26\\
kea & \cellcolor{mycolor!0.2}0.20 & \cellcolor{mycolor!0}0.00 & \cellcolor{mycolor!71.25}71.25 & \cellcolor{mycolor!0}0.00 & \cellcolor{mycolor!0.99}0.99 & \cellcolor{mycolor!0}0.00 & uk & \cellcolor{mycolor!0.59}0.59 & \cellcolor{mycolor!89.53}89.53 & \cellcolor{mycolor!74.31}74.31 & \cellcolor{mycolor!3.36}3.36 & \cellcolor{mycolor!0.49}0.49 & \cellcolor{mycolor!98.12}98.12\\
kk & \cellcolor{mycolor!0.1}0.10 & \cellcolor{mycolor!55.93}55.93 & \cellcolor{mycolor!76.48}76.48 & \cellcolor{mycolor!0.49}0.49 & \cellcolor{mycolor!0.1}0.10 & \cellcolor{mycolor!99.21}99.21 & umb & \cellcolor{mycolor!0.69}0.69 & \cellcolor{mycolor!0}0.00 & \cellcolor{mycolor!68.68}68.68 & \cellcolor{mycolor!0}0.00 & \cellcolor{mycolor!1.38}1.38 & \cellcolor{mycolor!0}0.00\\
km & \cellcolor{mycolor!0.4}0.40 & \cellcolor{mycolor!53.66}53.66 & \cellcolor{mycolor!80.34}80.34 & \cellcolor{mycolor!0.69}0.69 & \cellcolor{mycolor!0}0.00 & \cellcolor{mycolor!99.9}99.90 & ur & \cellcolor{mycolor!1.19}1.19 & \cellcolor{mycolor!25.49}25.49 & \cellcolor{mycolor!76.19}76.19 & \cellcolor{mycolor!2.77}2.77 & \cellcolor{mycolor!0.3}0.30 & \cellcolor{mycolor!97.92}97.92\\
kn & \cellcolor{mycolor!3.06}3.06 & \cellcolor{mycolor!49.6}49.60 & \cellcolor{mycolor!78.56}78.56 & \cellcolor{mycolor!1.09}1.09 & \cellcolor{mycolor!0.1}0.10 & \cellcolor{mycolor!99.9}99.90 & uz & \cellcolor{mycolor!0.4}0.40 & \cellcolor{mycolor!32.71}32.71 & \cellcolor{mycolor!74.51}74.51 & \cellcolor{mycolor!0.2}0.20 & \cellcolor{mycolor!1.78}1.78 & \cellcolor{mycolor!86.36}86.36\\
ko & \cellcolor{mycolor!1.58}1.58 & \cellcolor{mycolor!94.17}94.17 & \cellcolor{mycolor!60.57}60.57 & \cellcolor{mycolor!21.84}21.84 & \cellcolor{mycolor!0.1}0.10 & \cellcolor{mycolor!99.51}99.51 & vi & \cellcolor{mycolor!0}0.00 & \cellcolor{mycolor!95.85}95.85 & \cellcolor{mycolor!56.42}56.42 & \cellcolor{mycolor!13.24}13.24 & \cellcolor{mycolor!0.1}0.10 & \cellcolor{mycolor!99.7}99.70\\
ku & \cellcolor{mycolor!0.2}0.20 & \cellcolor{mycolor!28.06}28.06 & \cellcolor{mycolor!60.28}60.28 & \cellcolor{mycolor!0.49}0.49 & \cellcolor{mycolor!2.77}2.77 & \cellcolor{mycolor!72.73}72.73 & wo & \cellcolor{mycolor!1.09}1.09 & \cellcolor{mycolor!0}0.00 & \cellcolor{mycolor!73.32}73.32 & \cellcolor{mycolor!0}0.00 & \cellcolor{mycolor!2.96}2.96 & \cellcolor{mycolor!0}0.00\\
ky & \cellcolor{mycolor!0.4}0.40 & \cellcolor{mycolor!40.71}40.71 & \cellcolor{mycolor!75.79}75.79 & \cellcolor{mycolor!0}0.00 & \cellcolor{mycolor!0.1}0.10 & \cellcolor{mycolor!99.41}99.41 & xh & \cellcolor{mycolor!0.2}0.20 & \cellcolor{mycolor!0}0.00 & \cellcolor{mycolor!70.55}70.55 & \cellcolor{mycolor!0}0.00 & \cellcolor{mycolor!0.59}0.59 & \cellcolor{mycolor!0}0.00\\
lb & \cellcolor{mycolor!0.69}0.69 & \cellcolor{mycolor!31.23}31.23 & \cellcolor{mycolor!66.11}66.11 & \cellcolor{mycolor!0}0.00 & \cellcolor{mycolor!2.27}2.27 & \cellcolor{mycolor!87.75}87.75 & yo & \cellcolor{mycolor!0.1}0.10 & \cellcolor{mycolor!3.95}3.95 & \cellcolor{mycolor!67}67.00 & \cellcolor{mycolor!0}0.00 & \cellcolor{mycolor!0.1}0.10 & \cellcolor{mycolor!13.93}13.93\\
lg & \cellcolor{mycolor!1.38}1.38 & \cellcolor{mycolor!0}0.00 & \cellcolor{mycolor!74.11}74.11 & \cellcolor{mycolor!0}0.00 & \cellcolor{mycolor!12.65}12.65 & \cellcolor{mycolor!0}0.00 & zh & \cellcolor{mycolor!23.22}23.22 & \cellcolor{mycolor!70.16}70.16 & \cellcolor{mycolor!37.15}37.15 & \cellcolor{mycolor!35.67}35.67 & \cellcolor{mycolor!5.93}5.93 & \cellcolor{mycolor!93.08}93.08\\
ln & \cellcolor{mycolor!0.3}0.30 & \cellcolor{mycolor!0}0.00 & \cellcolor{mycolor!71.84}71.84 & \cellcolor{mycolor!0}0.00 & \cellcolor{mycolor!0.79}0.79 & \cellcolor{mycolor!0}0.00 & zhtrad & \cellcolor{mycolor!32.41}32.41 & \cellcolor{mycolor!0}0.00 & \cellcolor{mycolor!43.87}43.87 & \cellcolor{mycolor!0}0.00 & \cellcolor{mycolor!7.31}7.31 & \cellcolor{mycolor!0}0.00\\
 &  &  &  &  &  &  & zu & \cellcolor{mycolor!0.1}0.10 & \cellcolor{mycolor!0}0.00 & \cellcolor{mycolor!67.39}67.39 & \cellcolor{mycolor!0}0.00 & \cellcolor{mycolor!1.38}1.38 & \cellcolor{mycolor!0}0.00\\
 \bottomrule
    \end{tabular}}
    \caption{We utilize langdetect to identify the translation outputs from ja$\rightarrow$X of LLaMA2-Alpaca, Swallow and \name-Alpaca models on Flores-101 benchmark. $R_\mathrm{ja}$ represents the ratio of sentence in the translation predicted result where the top predicted language is Japanese. Conversely, $R_\mathrm{X}$ refers to the proportion where the top predicted language aligns with the target translated language. }
    \label{tab:compare_ja}
\end{table*}

\paragraph{M2M-100~\citep{fan2021beyond}} encompasses multilingual machine translation models designed to translate between any pair of 100 languages directly, without the need for English as an intermediary. The M2M-100 series includes models of varying sizes, specifically 418M, 1.2B, and 12B parameters. These models are part of a groundbreaking approach in the field of machine translation, aiming to enhance direct translation efficiency across a wide array of languages.

\paragraph{Lego-MT~\citep{yuan-etal-2023-lego}} is a novel approach to massively multilingual machine translation, featuring detachable models with individual branches for each language or group of languages. This design supports plug-and-play training and inference, enhancing flexibility and efficiency in language processing tasks.

\paragraph{MADLAD-400~\citep{kudugunta2024madlad}} is a multilingual machine translation model that leverages the T5 architecture and has been trained on a vast corpus of 250 billion tokens, covering over 450 languages.

\paragraph{Aya-101~\citep{aryabumi2024aya}}is an open-source, massively multilingual generative language model that operates on the mT5~\citep{xue-etal-2021-mt5} architecture, covering 101 languages and designed to bridge the performance gap in non-dominant languages. It incorporates a 13B parameter base and has undergone instruction-finetuning to achieve high performance across its extensive language range.

\section{The correlation between fertility and representation quality.}
\label{sec:corr_fert_and_quality}

We conduct experiments on Flores-101. Fertility is defined as the ratio of the $L_s$ to the $L_T$, where $L_s$ is the number of words for space-separated languages and characters for others and $L_T$ is the number of tokens after applying LLaMA2 tokenizer. The quality estimation of LLaMA on Flores-101 test. Cosine similarity focuses on the similarity in the expressions of LLaMA across sentence representation of the same sentence in English and other languages. Recall@1 is often used in the context of information retrieval, which measures the quality of representation. The experimental results, as shown in Figure~\ref{fig:correlation_fertility_with_quality}, indicate fertility has a high correlation with the representation quality.

\begin{table*}[!ht]
    \centering
    \resizebox{1\linewidth}{!}{
    \begin{tabular}{l|p{24.9cm}}
    \toprule
    \textbf{Model} & \textbf{Templates} \\
    \midrule

    \makecell[l]{\method-\\Alpaca} & \makecell[l]{Below is an instruction that describes a task, paired with an input that provides further context. Write a response that appropriately completes the request. \\
\#\#\# Instruction: \\
Translate the following sentences from English to Chinese Simpl \\
\#\#\# Input: \\
"We now have 4-month-old mice that are non-diabetic that used to be diabetic," he added. \\
\#\#\# Response:\begin{CJK}{UTF8}{gbsn}{他补充道：“我们现在有 4 个月大没有糖尿病的老鼠，但它们曾经得过该病。”}\end{CJK}} \\
    
    \midrule
    
     \makecell[l]{LLaMA \\Series Models}    & \makecell[l]{Below is an instruction that describes a task, paired with an input that provides further context. Write a response that appropriately completes the request. \\
\#\#\# Instruction: \\
Translate the following sentences from English to Chinese Simpl \\
\#\#\# Input: \\
"We now have 4-month-old mice that are non-diabetic that used to be diabetic," he added. \\
\#\#\# Response:\begin{CJK}{UTF8}{gbsn}{他补充道：“我们现在有 4 个月大没有糖尿病的老鼠，但它们曾经得过该病。”}\end{CJK}} \\

\midrule

yayi2 & \makecell[l]{Below is an instruction that describes a task, paired with an input that provides further context. Write a response that appropriately completes the request. \\
\#\#\# Instruction: \\
Translate the following sentences from English to Chinese Simpl \\
\#\#\# Input: \\
"We now have 4-month-old mice that are non-diabetic that used to be diabetic," he added. \\
\#\#\# Response:\begin{CJK}{UTF8}{gbsn}{他补充道：“我们现在有 4 个月大没有糖尿病的老鼠，但它们曾经得过该病。”}\end{CJK}} \\

\midrule

polylm & \makecell[l]{"We now have 4-month-old mice that are non-diabetic that used to be diabetic," he added. \\
Translate this sentence English to Chinese Simpl. \begin{CJK}{UTF8}{gbsn}{他补充道：“我们现在有 4 个月大没有糖尿病的老鼠，但它们曾经得过该病。”}\end{CJK}} \\

\midrule

TowerInstruct & \makecell[l]{
<|im\_start|>user \\
Translate the following text from English into Chinese. \\
English: "We now have 4-month-old mice that are non-diabetic that used to be diabetic," he added. \\
Chinese:<|im\_end|> \\
<|im\_start|>assistant \\
\begin{CJK}{UTF8}{gbsn}{他补充道：“我们现在有 4 个月大没有糖尿病的老鼠，但它们曾经得过该病。”}\end{CJK}} \\

\midrule

aya23 & <BOS\_TOKEN><|START\_OF\_TURN\_TOKEN|><|USER\_TOKEN|>Translate the following sentences from English to Chinese: "We now have 4-month-old mice that are non-diabetic that used to be diabetic," he added.<|END\_OF\_TURN\_TOKEN|><|START\_OF\_TURN\_TOKEN|><|CHATBOT\_TOKEN|>\begin{CJK}{UTF8}{gbsn}{他补充道：“我们现在有 4 个月大没有糖尿病的老鼠，但它们曾经得过该病。”}\end{CJK}<|END\_OF\_TURN\_TOKEN|>\\

\midrule

Qwen2 instruct & \makecell[l]{
system \\
You are a helpful assistant. \\
user \\
Translate the following sentences from English to Chinese Simpl: "We now have 4-month-old mice that are non-diabetic that used to be diabetic," he added. \\
assistant \\
\begin{CJK}{UTF8}{gbsn}{他补充道：“我们现在有 4 个月大没有糖尿病的老鼠，但它们曾经得过该病。”}\end{CJK}} \\

\midrule

ChineseAlpaca-2 & \makecell[l]{
[INST] <<SYS>> \\
You are a helpful assistant. \begin{CJK}{UTF8}{gbsn}{你是一个乐于助人的助手。}\end{CJK} \\
<</SYS>> \\
Translate the following sentences from English to Chinese Simpl: "We now have 4-month-old mice that are non-diabetic that used to be diabetic," he added. [/INS\\ T] \begin{CJK}{UTF8}{gbsn}{他补充道：“我们现在有 4 个月大没有糖尿病的老鼠，但它们曾经得过该病。”}\end{CJK} 
} \\

\midrule

Swallow & \makecell[l]{
[INST] <<SYS>> \\
\begin{CJK}{UTF8}{gbsn}{あなたは誠実で優秀な日本人のアシスタントです。}\end{CJK}
<</SYS>> \\
Translate the following sentences from Japanese to Chinese Simpl: \begin{CJK}{UTF8}{gbsn}{「我々が飼っている生後4か月のマウスはかつて糖尿病でしたが現在は糖尿病ではない、」}\end{CJK}\\ \begin{CJK}{UTF8}{gbsn}{と彼は付け加えました。 [/INST] 「他补充道：“我们现在有 4 个月大没有糖尿病的老鼠，但它们曾经得过该病。”」}\end{CJK}
}\\

\midrule
Madlad & '<2zh> "We now have 4-month-old mice that are non-diabetic that used to be diabetic," he added.' \begin{CJK}{UTF8}{gbsn}{他补充道：“我们现在有 4 个月大没有糖尿病的老鼠，但它们曾经得过该病。”}\end{CJK} \\

\bottomrule
\end{tabular}}
\caption{Examples of instruction templates utilized for all evaluated LLMs, with the translation result, \begin{CJK}{UTF8}{gbsn}{他补充道：“我们现在有 4 个月大没有糖尿病的老鼠，但它们曾经得过该病。”}\end{CJK}, using the reference instead of the model's output.}
\label{tab:instruction_example}
\end{table*}

\section{Introduction to KS-Lottery.}
\label{sec:ks_lottery}

KS-Lottery is a technique designed to identify a small, highly effective subset of parameters within LLMs for multilingual capability transfer. The core concept of this method involves utilizing the Kolmogorov-Smirnov Test to examine the distribution shift of parameters before and after fine-tuning. This approach helps in pinpointing the ``winning tickets'' or the most impactful parameters that contribute significantly to the model’s performance in multilingual tasks.

\section{1-hop translation in data augmentation is enough.}
\label{sec:select_hop_translation}

Given a parallel dataset subset~($\mathcal{D}_\mathrm{P}$) from~$\mathcal{D}_\mathrm{para}^A$ that contains translations in all directions for 6 languages~(en,fr,es,zh,ta,th) and a monolingual subset~($\mathcal{D}_\mathrm{M}$) from ~$\mathcal{D}_\mathrm{mono}^A$ for the same 6 languages. We then perform non-repetitive sampling 12,500 sentence pairs from $\mathcal{D}_\mathrm{P}$ in each direction to generate two subsets of parallel corpus data $\mathcal{D}_\mathrm{P_1}$ and $\mathcal{D}_\mathrm{P_2}$, respectively. Consequently, we preserve $\mathcal{D}_\mathrm{P_1}$ and evaluate the effect of augmentation on parallel data $\mathcal{D}_\mathrm{P_2}$ or monolingual data $\mathcal{D}_\mathrm{M}$, resulting in two new dataset, $\mathcal{D}_\mathrm{P_2}'$ and $\mathcal{D}_\mathrm{M}'$, post-augmentation. To assess both the in-domain and out-of-domain capabilities of the model, we perform inference on it using 10 languages (en, fr, es, pt, de, zh, ta,  th, is, zu), utilizing the Flores-101.

We use two different multilingual dictionaries MUSE provided by~\citet{lample2018unsupervised}~\footnote{https://github.com/facebookresearch/MUSE.}, and PanLex~\cite{wang-etal-2022-expanding}. In the context of a multilingual dictionary, we can use ``1-hop''  and ``2-hop'' to characterize the translation relationship among different languages, an example shown in Table~\ref{tab:select_hop}.

We use the MUSE dictionary to perform data augmentation on both parallel $\mathcal{D}_\mathrm{P_2}$ and monolingual $\mathcal{D}_\mathrm{M}$ data, utilizing 1-hop and 2-hop translations. As shown in Table~\ref{tab:select_hop}, using different hop translation for augmentation does not significantly impact the final translation performance. Multi-hop translation sometimes can even result in poorer performance.

\section{Design of parallel format}
\label{sec:parallel_format}
\paragraph{The Usage of  Parallel Data.} Parallel data can be utilized in two distinct ways: split-parallel or connected-parallel. \textbf{Split-Parallel}: Consider the source language data and target language data involved in parallel data as two distinct monolingual datasets, which are randomly shuffled throughout the entire training set. \textbf{Connected-Parallel}: In the training process, we treat each pair of source and target language sentences from the parallel dataset as a single data point by concatenating them.

Based on different forms of parallel data, supervised fine-tuning~(SFT) is conducted separately on ceb-centric using both parallel and monolingual datasets. As indicated in Table~\ref{tab:usage_para}, we observed that the form of parallel data primarily impacts translation performance, with no significant difference in general tasks and cross-lingual general tasks; however, the disparity in translation is pronounced. We specifically highlighted some high-resource translation directions and found that such gaps are quite significant.

\section{Comparison Results Between Our Model and GPT-4}
\label{sec:gpt4}
In Figure~\ref{fig:comparison_gpt-4}, we compare the performance gap between our model and GPT-4. 
Considering the API cost of evaluating GPT-4, we only evaluate the mutual translation performance among seven languages~(en, zh, de, ne, ar, az, ceb).  
Experiment results show that while our model lags behind in high-resource translation directions, it achieves on-par or even superior performance in low-resource translation.

\section{Comparsion between \name-Alpaca and language-specific LLMs.} 
\label{sec:compare_lg_specific}

\begin{figure}
    \centering
    \includegraphics[width=1\linewidth]{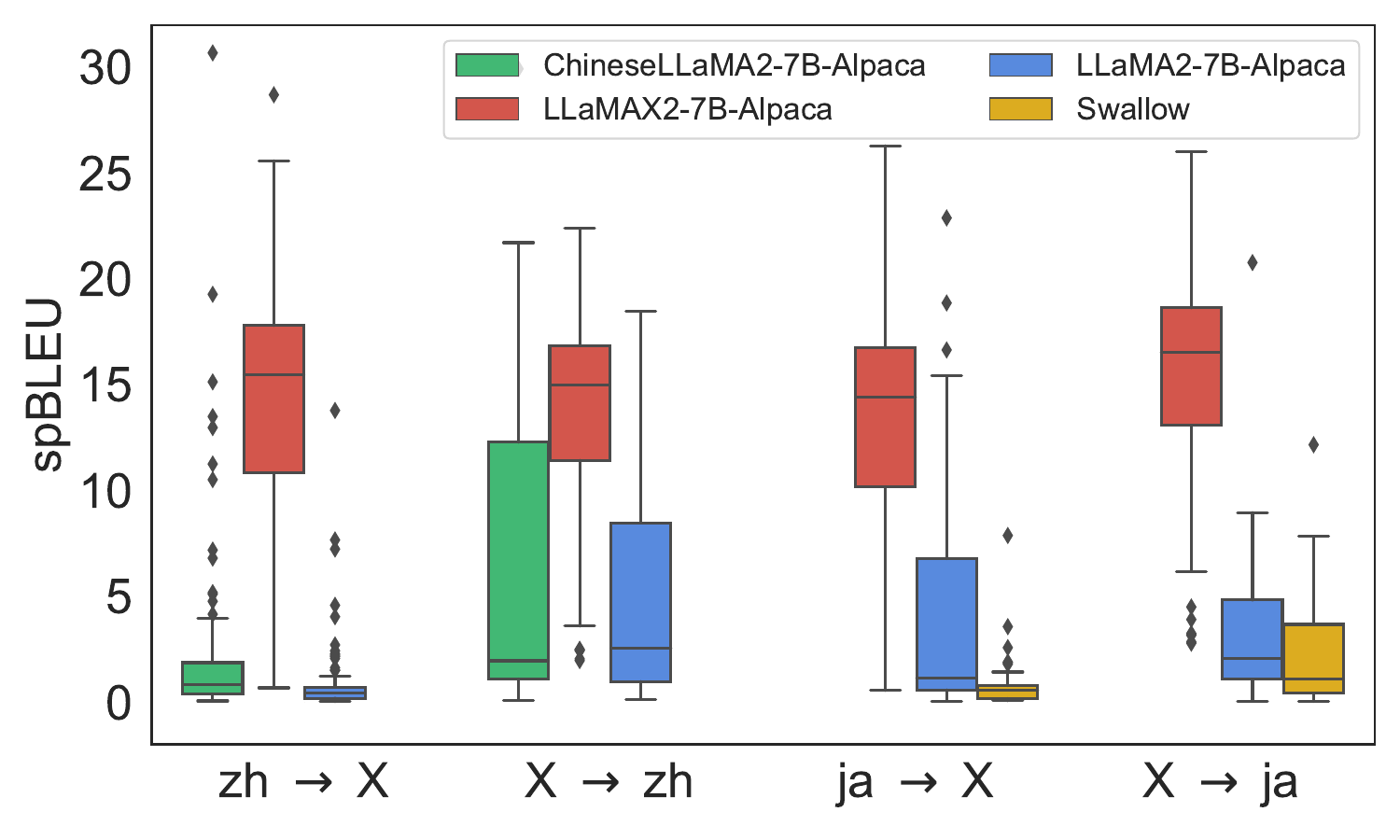}
    \caption{Significant improvements in language-specific-centric translation are observed with \name-Alpaca compared to LLaMA2-7B-Alpaca, ChineseLLaMA2-7B-Alpaca, and Swallow, as demonstrated in the translation performance analysis on all translation directions in the Flores-101 dataset.}
    \label{fig:compare_llms}
\end{figure}

The comparison between \name-Alpaca, ChineseLLaMA2-Alpaca, and Swallow~(a Japanese-specific LLM) explores the difference between the traditional pipeline for enhancing specific language capabilities based on existing pre-trained models and our proposed recipe. As shown in Figure~\ref{fig:compare_llms}, we evaluate language-specific LLMs to translate from the enhanced language to any of the 101 languages on Flores-101 and find that their performance is not significantly different from the original LLaMA2 model, but there exists a notable performance gap compared to \name-Alpaca. As we described in Section~\ref{sec:vocab_analysis}, excessively adding new language-specific tokens can shift the focus of training the LLM. 

In addition, we conduct a deeper analysis of translation output to identify the factors contributing to the limited improvement in translation performance. The experimental results in Table~\ref{tab:compare_zh} indicate that the language-specific LLM obtained through the traditional pipeline tends to output specific languages, while \name can accurately produce the answer with the corresponding language. 

We perform further comparisons between \name-Alpaca and Japanese-specific LLMs-Swallow. After using \name-Alpaca and Swallow to generate translations from Japanese~(ja) to any language in Flores-101, we apply langdetect to determine the language of each translation result and calculate the proportion of Japanese and target translated language respectively. The experimental result, as shown in Table~\ref{tab:compare_ja}, indicates that the Japanese-specific LLM tends to output Japanese, whereas \name-Alpaca performs more accurately in producing the target language.

\section{Prompt Templates}
\label{sec:instruction_prompt}

We offer a comprehensive collection of prompt instruction templates, as illustrated in Table~\ref{tab:instruction_example}, which are utilized for all evaluated LLMs. These templates are meticulously designed based on existing LLMs, playing a crucial role in obtaining accurate model results and ensuring fairness in comparisons. Our goal in providing these templates is to promote transparency and make it easier to reproduce our findings.